\documentclass[twocolumn,letterpaper]{IEEEAerospaceCLS}  

\usepackage[]{graphicx, amsmath, subfig}    
\newcommand{\ignore}[1]{}  

\usepackage{tikz}

\newcommand\centerofmass{%
    \tikz[radius=0.4em] {%
        \fill (0,0) -- ++(0.4em,0) arc [start angle=0,end angle=90] -- ++(0,-0.8em) arc [start angle=270, end angle=180];%
        \draw (0,0) circle;%
    }%
}

\def \deg{\ensuremath{^\circ}}
\def \norm#1{\vert\vert#1\vert\vert}

\begin{document}
\title{Mid-Air Helicopter Delivery at Mars Using a Jetpack}

\author{%
Jeff Delaune$^1$, Jacob Izraelevitz$^1$, Samuel Sirlin$^1$, David Sternberg$^1$, Louis Giersch$^1$, L. Phillipe Tosi$^1$, \\
Evgeniy Skliyanskiy$^1$, Larry Young$^2$, Michael Mischna$^1$, Shannah Withrow-Maser$^2$, Juergen Mueller$^1$, \\
Joshua Bowman$^2$, Mark S Wallace$^1$,  H\r{a}vard F. Grip$^1$, Larry Matthies$^1$, Wayne Johnson$^2$, Matthew Keennon$^3$, \\
Benjamin Pipenberg$^3$, Harsh Patel$^1$, Christopher Lim$^1$, Aaron Schutte$^1$, Marcel Veismann$^4$, \\
Haley Cummings$^2$, Sarah Conley$^2$, Jonathan Bapst$^1$, Theodore Tzanetos$^1$, Roland Brockers$^1$, \\
Abhinandan Jain$^1$, David Bayard$^1$, Art Chmielewski$^1$, Olivier Toupet$^1$, Joel Burdick$^4$, Morteza Gharib$^4$ \\
and J. (Bob) Balaram$^1$\\
\and
$^1$Jet Propulsion Laboratory, California Institute of Technology \\
Pasadena, CA\\
\and
$^2$NASA Ames Research Center \\
Moffett Field, CA\\
\and
$^3$Aerovironment Inc.\\
Simi Valley, CA \\
\and
\hspace{2cm}
\and
$^4$California Institute of Technology\\
Pasadena, CA 91109\\
\and
Email to: delaune@jpl.nasa.gov
\thanks{\footnotesize 978-1-6654-3760-8/22/$\$31.00$ \copyright2021 IEEE} 
}

\maketitle

\thispagestyle{plain}
\pagestyle{plain}

\begin{abstract}
Mid-Air Helicopter Delivery (MAHD) is a new Entry, Descent and Landing (EDL) architecture to enable in situ mobility for Mars science at lower cost than previous missions. It uses a jetpack to slow down a Mars Science Helicopter (MSH) after separation from the backshell, and reach aerodynamic conditions suitable for helicopter take-off in mid air.
For given aeroshell dimensions, only MAHD's lander-free approach leaves enough room in the aeroshell to accommodate the largest rotor option for MSH. This drastically improves flight performance, notably allowing +150\% increased science payload mass. Compared to heritage EDL approaches, the simpler MAHD architecture is also likely to reduce cost, and enables access to more hazardous and higher-elevation terrains on Mars.
This paper introduces a design for the MAHD system architecture and operations. We present a mechanical configuration that fits both MSH and the jetpack within the 2.65-m Mars heritage aeroshell, and a jetpack control architecture which fully leverages the available helicopter avionics. We discuss preliminary numerical models of the flow dynamics resulting from the interaction between the jets, the rotors and the side winds. We define a force-torque sensing architecture capable of handling the wind and trimming the rotors to prepare for safe take-off. Finally, we analyze the dynamic environment and closed-loop control simulation results to demonstrate the preliminary feasibility of MAHD.
\end{abstract}

\tableofcontents

\section{Introduction}

Planetary rotorcraft became a reality on April 19, 2021, when the \emph{Ingenuity} Mars Helicopter demonstrated the first powered airborne flight on another world. While initially designed for up to five flights as a technology demonstration~\cite{balaram2018scitech}, the aircraft has exceeded all expectations and has been assigned to an operation demonstration mission. In this new role, Ingenuity now acts as an aerial scout to help increase the science return to the \emph{Perseverance} rover. As of October 2021, it has successfully flown thirteen times on Mars.

\begin{figure}
\centering
\includegraphics[width=3.25in]{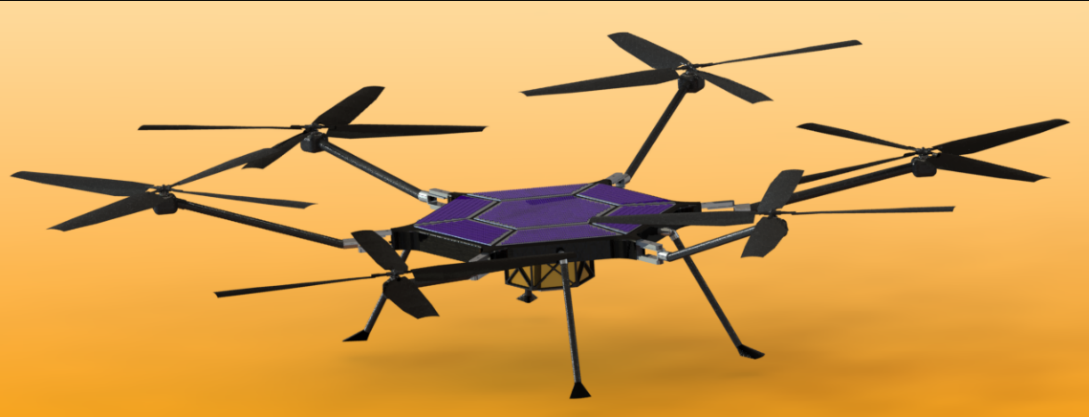}\\
\caption{\textbf{3D rendering of the Mars Science Helicopter (MSH) concept in flight configuration, with rotor arms and landing legs deployed.}}
\label{fig:msh-deployed}
\end{figure}

Ingenuity's demonstrations have paved the way for a new era of planetary exploration. Aerial robots equipped with science instruments could fill in the gap in resolution and range between the data returned by orbital probes and the data returned by surface assets~\cite{Rapin2020}. They also enable science access to more hazardous terrains than rovers and to the atmospheric boundary layer~\cite{Bapst2020}. The \emph{Mars Science Helicopter} (MSH), illustrated in Figure~\ref{fig:msh-deployed}, is a concept for the next generation of Mars rotorcraft~\cite{tzanetos2021aeroconf}. In flight conditions representative of Ingenuity's early flights\footnote{Approximately local noon flights in the spring at Jezero crater.}, the 4-m-wide MSH rotorcraft illustrated in Figure~\ref{fig:msh-deployed} is expected to carry 5 kg of science payload up to 13 km in range.

MSH is designed as a helicopter-only mission to enable in situ mobility for Mars science at a lower cost than past rovers. Unlike Ingenuity, MSH will not be able to get assistance from a large rover coming with its own Entry, Descent and Landing (EDL) system to be deployed on the surface of Mars. Current EDL architectures are also limited in both the hazardousness and elevation of the terrains they can reach. Using MSH as a baseline, our research addresses the problem of EDL for a Mars helicopter-only mission. Specifically, we seek an architecture which maximizes the helicopter flight performance and terrain access, while minimizing cost and risk to the mission.

\begin{figure}
\centering
\includegraphics[width=3.25in]{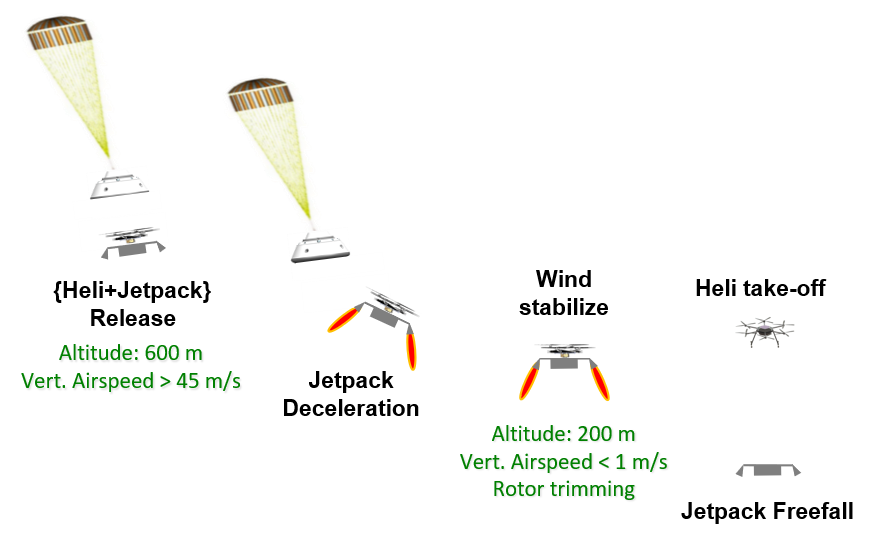}\\
\caption{\textbf{Concept of operations for the Mid-Air Helicopter Delivery (MAHD) system using a jetpack.}}
\label{fig:conops-simple}
\end{figure}

Our solution is a new EDL architecture called the \emph{Mid-Air Helicopter Delivery} system (MAHD), and is illustrated in Figure~\ref{fig:conops-simple}. MAHD does not rely on a traditional lander. The helicopter takes off from a jetpack in mid-air after separation from the backshell, and lands like any other flight. The jetpack is only needed to decelerate and reach aerodynamic conditions suitable for helicopter take-off. It is entirely controlled with the helicopter avionics. Compared to alternative EDL approaches, there are critical advantages to using MAHD for a MSH mission:
\begin{enumerate}
    \item Improved rotorcraft performance,
    \item Simpler architecture, likely to reduce cost,
    \item Access to more hazardous access,
    \item Access to high-elevation terrains.
\end{enumerate}

This paper covers the following research contributions:
\begin{itemize}
    \item MAHD system architecture, for EDL of a rotorcraft-only mission to Mars,
    \item mechanical design of the jetpack and its interfaces to the helicopter and backshell,
    \item analysis of the wind conditions at the key MAHD events using mesoscale modeling,
    \item preliminary modeling of the jet, rotor and wind interactions,
    \item Guidance, Navigation and Control (GNC) architecture of the jetpack,
    \item architecture to safely take off in the wind capability using force-torque sensing,
    \item closed-loop controls simulation results for the entry and descent, jetpack powered phase and helicopter take-off.
\end{itemize}
Section~\ref{sec:related-work} details the technical design of MSH and compares MAHD to alternative EDL approaches to deliver a helicopter-only mission to Mars. Section~\ref{sec:system} provides an overview of the system architecture and operations designed to meet the MAHD requirements. Section~\ref{sec:mech} covers the mechanical design of the jetpack assembly, its interfaces to the backshell and MSH, as well as a discussion of the changes imposed on MSH by that design. Section~\ref{sec:wind} introduces the modeling effort performed to obtain wind dynamics representative of a candidate first landing site for MSH. Section~\ref{sec:flow} discusses the preliminary computational fluid dynamics simulation results regarding the interaction of the jets, rotors and winds. Section~\ref{sec:GNC} introduces the GNC architecture of the jetpack, and discusses closed-loop controls performance in a Monte Carlo analysis. Finally, Section~\ref{sec:takeoff} investigates the feasibility of force-torque sensing of the wind before helicopter take-off.

\section{Related Work}
\label{sec:related-work}

\subsection{Mars Science Helicopter}

MSH is a hexacopter configuration illustrated in Figure~\ref{fig:msh-deployed}. Each of the six rotors measures 1.28 m in diameter and includes four blades. The maximum diameter of the whole vehicle is 2.38 m in stowed configuration, and 3.97 m with the rotor arms deployed. These dimensions are driven by the requirement to fit in the 2.65-m diameter, Mars-heritage aeroshell used by MAHD. This aeroshell was selected for cost saving purposes. It flew on the Pathfinder, Mars Exploration Rover, Phoenix and Insight Mars missions. For a total mass of 31.2 kg, MSH is capable of carrying 5 kg of science payload up to 13.3 km in range for any single flight, or 6.5 minutes hover time, in conditions representative of Ingenuity's early flights. For comparison, Ingenuity weighs 1.8 kg and its coaxial rotors are 1.21 m in diameter, but it carries no dedicated science payload and it was only designed to fly up to 0.18 km per flight, or 1.5 minute hover time\footnote{These are Ingenuity's design values, using similar model to MSH for comparison purposes. Ingenuity already demonstrated these models are very conservative, by flying 625 m in flight 9, and 2.8 min in flight 12.}. Table~\ref{tab:msh-vs-ing} compares the design performance of MSH and Ingenuity. For more details about the MSH design, the reader should refer to \cite{tzanetos2021aeroconf} and \cite{johnson2020}.
\begin{table}
\renewcommand{\arraystretch}{1.3}
\caption{\bf Comparison between the MSH and Ingenuity design performance.}
\label{tab:msh-vs-ing}
\centering
\begin{tabular}{|c||c|c|}
\hline
& \bfseries MSH & \bfseries Ingenuity \\
\hline
\hline
\bfseries Total mass [kg] & 31.2 & 1.8 \\
\bfseries Payload mass [kg] & 5 & 0 \\
\bfseries Max. diameter [m] & 3.97 & 1.21\\
\bfseries Range [km] & 13.3 & 0.18\\
\bfseries Hover time [min] & 6.5 & 1.5\\
\bfseries Number rotors  & 6 & 2 (coaxial) \\
\bfseries Number blades & 4 & 2 \\
\bfseries Rotor diameter [m] & 1.28 & 1.21\\
\bfseries Rotor solidity & 0.25 & 0.148\\
\hline
\end{tabular}
\end{table}

\subsection{Alternative EDL Approaches}

Past successful landings at Mars can be divided into three categories: legged landers\footnote{Viking 1 and 2 (1976), Phoenix (2008), Insight (2018) and Zhurong rover (2021).}, airbag landers\footnote{Mars Pathfinder (1997), Mars Exploration Rovers Spirit and Opportunity (2004).} and the skycrane\footnote{Curiosity (2012) and Perseverance (2021) rovers.}~\cite{braun2007}. The main advantage of re-using one of the successful Mars EDL architectures for a helicopter-only mission is the high Technology Readiness Level (TRL). However, compared to MAHD, the landing elements associated with the past options significantly limit the internal volume available for the rotors within the aeroshell. Smaller rotors lead to degraded flight performance.

It is worth emphasizing how much rotorcraft performance is impacted by the smaller rotors. In previous work, we identified that only modified MSH configurations with 50-cm or 58-cm rotor radius could be folded within the tetrahedron volume of the heritage Pathfinder airbag lander~\cite{johnson2020}. Only MAHD allows the 64-cm rotors to fit within the 2.65-m aeroshell. Table~\ref{tab:edl-type-vs-msh-perf} compares the performance of MSH between the 58 and 64-cm rotors\footnote{There is no significant change between the 50 and 58-cm rotor configuration for the parameters shown in the table.}. Compared to an airbag lander, MAHD enables +27\% range, +30\% hover time and notably +150\% payload mass. With the 2.65-m aeroshell, MAHD can be considered an enabler for a MSH-only mission.
\begin{table}
\renewcommand{\arraystretch}{1.3}
\caption{\bf Performance comparison between the MSH version optimized for an airbag lander, and the MSH version optimized for MAHD. All options assume the same 2.65-m aeroshell.}
\label{tab:edl-type-vs-msh-perf}
\centering
\begin{tabular}{|c||c|c|}
\hline
\bfseries EDL Type & \bfseries Airbag & \bfseries MAHD \\
\hline
\hline
\bfseries Rotor Radius [cm] & 58 & 64 \\
\bfseries Range [km] & 10.5 & 13.3 \\
\bfseries Hover Time [min] & 5.0 & 6.5 \\
\bfseries Payload Mass [kg] & 2 & 5 \\
\hline
\end{tabular}
\end{table}

In addition to accommodating a larger rotorcraft with improved performance, there are other advantages to using MAHD. First, the jetpack is a simpler design compared to a legged lander, an airbag lander, or the skycrane. Even when considering lowered redevelopment costs for heritage solutions, the simplicity of MAHD is expected to make it the most affordable EDL option. Second, MAHD uses the helicopter for landing. MSH's vision-based landing site detection capability will enable access to more hazardous terrains than any past mission. Last, MAHD is the lightest EDL option. Combined with the reduced weight of a helicopter compared to past rover or lander missions, this reduces the ballistic coefficient of the aeroshell by a factor of two. This means higher entry deceleration at entry, and access to the high-elevation terrains never explored on Mars~\cite{delaune2020mad}.

In previous work, we explored mid-air deployment of a smaller coaxial Mars helicopter without a jetpack~\cite{delaune2020mad}. While flying the rotorcraft directly out the backshell may be the ultimate mid-air deployment, the airspeed at separation on Mars is predicted to be in the 45-60 m/s range, mostly vertical. These conditions are currently beyond MSH's 30 m/s max design airspeed. They are also hard to test on Earth. The jetpack is a testable solution for MSH-like Mars rotorcraft. On Titan, NASA's \emph{Dragonfly} mission is planning to execute a jetpack-free mid-air deployment maneuver in 2034~\cite{cornelius2021vfs}. However, the dynamics at backshell separation ($\sim$~2 m/s) are much slower on Titan than on Mars\footnote{For reference, Dragonfly's max-range design airspeed is 10 m/s~\cite{lorenz2018}.}. Lessons learned from MAHD can potentially help reduce the risk for Dragonfly.

\section{System Overview}
\label{sec:system}

\subsection{Design Objectives}

\textbf{The primary objective of the MAHD EDL architecture is to use a jetpack to allow the MSH rotorcraft to deploy under stable flight conditions while in mid-air.} In particular, MAHD must be robust to the wind dynamics at Mars. Helicopter take-off should take place at MSH's nominal cruise height of 200 m height Above Ground Level (AGL).

A secondary set of objectives is also defined to keep the cost and complexity of MAHD to a minimum, and maximize the science return potential of the mission:
\begin{itemize}
    \item fit within the 2.65-m-diameter, 70-deg-cone-angle heritage aeroshell;
    \item re-use TRL-9 Mars EDL elements when possible;
    \item impose as few changes as possible to the MSH rotorcraft design optimized for the science mission;
    \item use MSH avionics hardware and software to control the whole EDL.
\end{itemize}

For design purposes, we defined a notional landing site in the Eridania basin (37$\deg$ S, 184$\deg$ E) based upon its high science potential and community interest. A Mars-Earth transfer was designed to reach the Eridania Basin during the late summer (Ls = 335 deg), which is a desirable season for EDL. MAHD was assumed to happen between 1000 hr and 1100 local mean solar time, based on UHF communications constraints through the TGO and MAVEN orbiters, as well as atmospheric density constraints for MSH.

\subsection{System Architecture}

\begin{figure}
\centering
\includegraphics[width=2.75in]{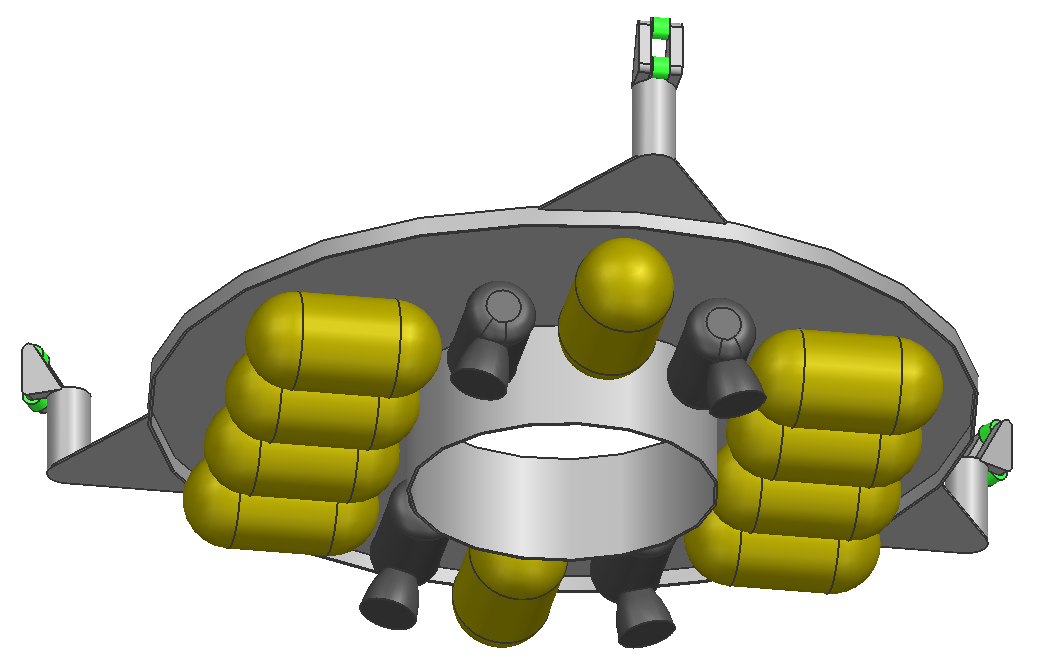}\\
\caption{\textbf{Mechanical model of the jetpack assembly. The four thrusters are colored in dark gray. The ten tanks are colored in yellow (helicopter not shown).}}
\label{fig:jetpack-assembly}
\end{figure}

The jetpack module uses four thrusters drawing propellant from ten tanks. This design is illustrated in Figure~\ref{fig:jetpack-assembly}. The thrusters are assumed to be similar to Aerojet Rocketdyne MR-107N flown on the Phoenix and Insight landers, but with a 90-deg nozzle angle to fit in the aeroshell. The thrusters are pulse-width-modulated, with a maximum steady-state thrust of 269 N and minimum ON time of 20 ms. The tanks are assumed to be similar to ATK-PSI 80588-1, providing up to 1.3 L of propellant per tank. The jetpack is 95 cm in diameter and attaches at the bottom of MSH's fuselage. It features a 35-cm-diameter central porthole for the MSH terrain navigation sensor. The wet mass of the jetpack, including structure, tanks, thrusters, but not including the mass of the helicopter, is 49.52 kg (including margins).

\begin{figure}
\centering
\includegraphics[width=3.25in]{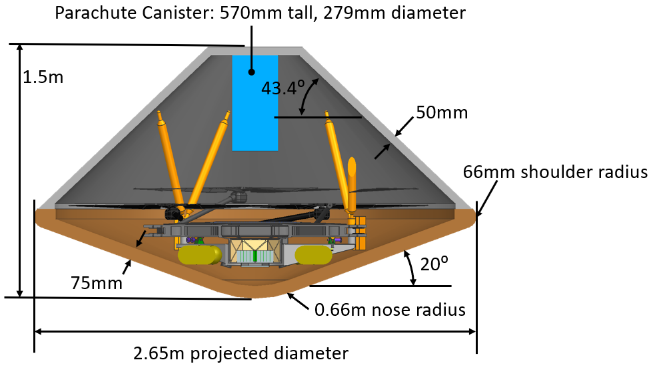}\\
\caption{\textbf{Mechanical configuration showing that both the jetpack and MSH in stowed configuration fit inside the 2.65-m heritage aeroshell.}}
\label{fig:mahd-aeroshell}
\end{figure}

MAHD assumes ballistic atmospheric entry using the 2.65-m-diameter aeroshell with the 70-deg sphere cone heatshield geometry. MSH is rigidly attached to the jetpack, which is itself mounted to the backshell through a set of bipods, as illustrated in Figure~\ref{fig:mahd-aeroshell}. The supersonic parachute is based on the 11.8-m-diameter disk gap band design used for the Insight lander. The parachute trigger logic is based on acceleration threshold limits.

Table~\ref{tab:mass-budget} summarizes the mass allocations for the system, which culminates in a 275-kg entry mass. This is less than half of the lightest entry mass for previous missions, which was Phoenix at 573 kg.
\begin{table}
\renewcommand{\arraystretch}{1.3}
\caption{\bf Mass allocations for MAHD, including margins greater than 20\% on all subsystems.}
\label{tab:mass-budget}
\centering
\begin{tabular}{|c||c|}
\hline
\bfseries Module & \bfseries Mass allocation [kg]\\
\hline
\hline
\bfseries MSH & 37.4 (31.2+20\%)\\
\bfseries Jetpack & 46.9 \\
\bfseries Aeroshell & 190.7 \\
\hline
\bfseries Total & 275 \\
\hline
\end{tabular}
\end{table}

\subsection{Concept of operation}

\begin{figure*}
\centering
\includegraphics[width=7in]{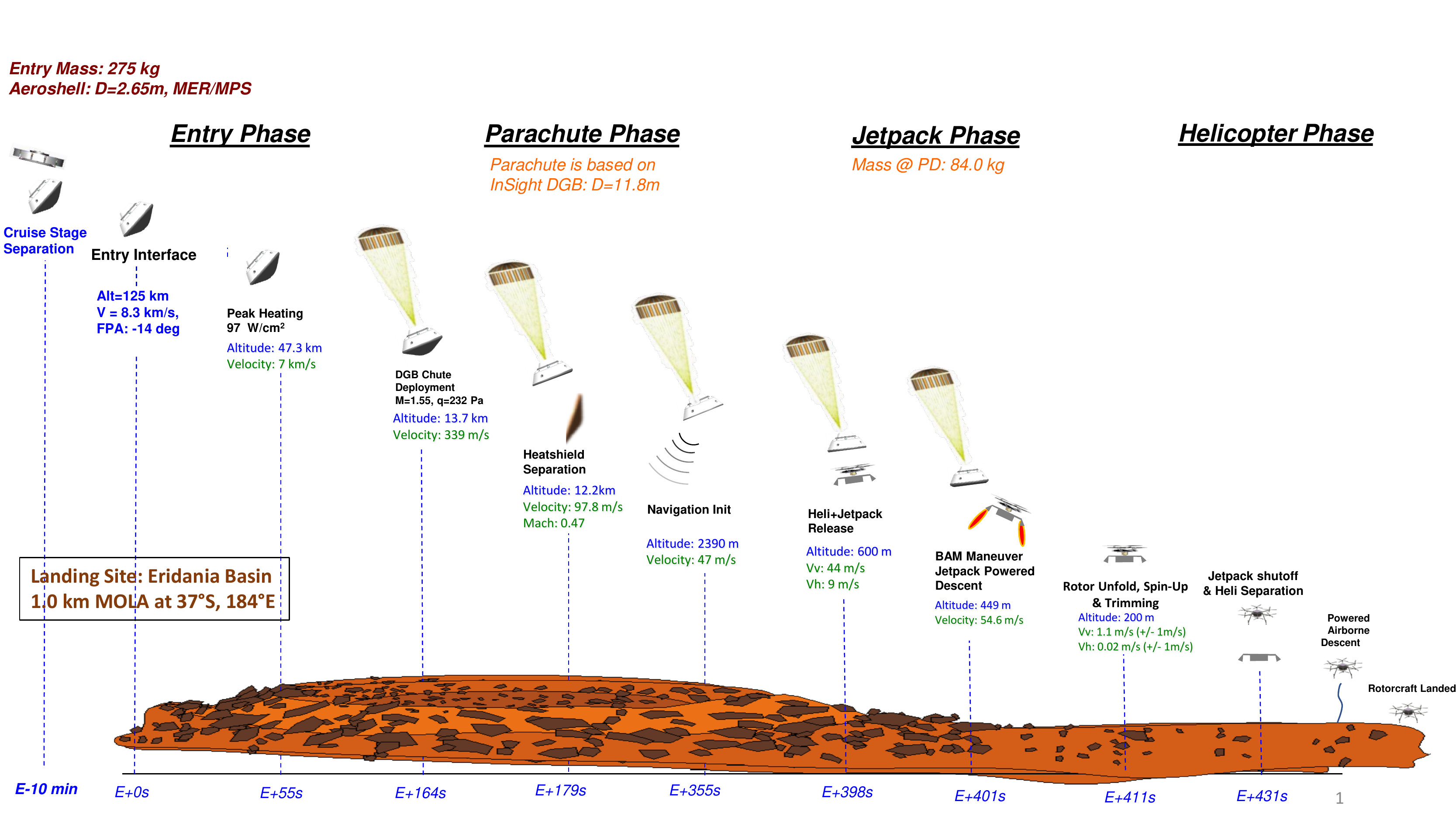}
\caption{\bf{Concept of operations for the jetpack-assisted Mid-Air Helicopter Delivery system (MAHD}. Numbers are representative of a landing site at Eridania Basin at Ls = 335 deg (late Southern summer). Velocity and altitude are expressed with respect to the local terrain.}
\label{fig:conops-detailed}
\end{figure*}

Figure~\ref{fig:conops-detailed} illustrates the complete concept of operations for Mars EDL with the MAHD system, from atmospheric entry to helicopter touchdown at the Eridania basin reference landing site. Up to heatshield separation at 12.2 km AGL, the aeroshell follows a ballistic entry trajectory and parachute deployment strategy directly inherited from the InSight landing.

The operation specific to MAHD truly starts once the heatshield is jettisoned. At 2.4 km AGL, the helicopter range-visual-inertial navigation system is initialized and will stay on until touchdown~\cite{delaune2020aero}. At 600 m AGL, the {jetpack+helicopter} assembly separates from the backshell. Following a 3-s freefall, the thrusters fire up to stabilize the jetpack and enter a polynomial guidance mode to safely divert away from the backshell and decelerate to reach terrain-relative hover. During the whole powered phase, the jetpack is entirely controlled using position and attitude feedback coming from the helicopter navigation system. The GNC software is running on the helicopter computer, using the sensors mounted on the helicopter. The thrusters are controlled using umbilical connections between the helicopter and the jetpack.

\begin{figure*}[h!]
    \centering
    \includegraphics[width=\linewidth]{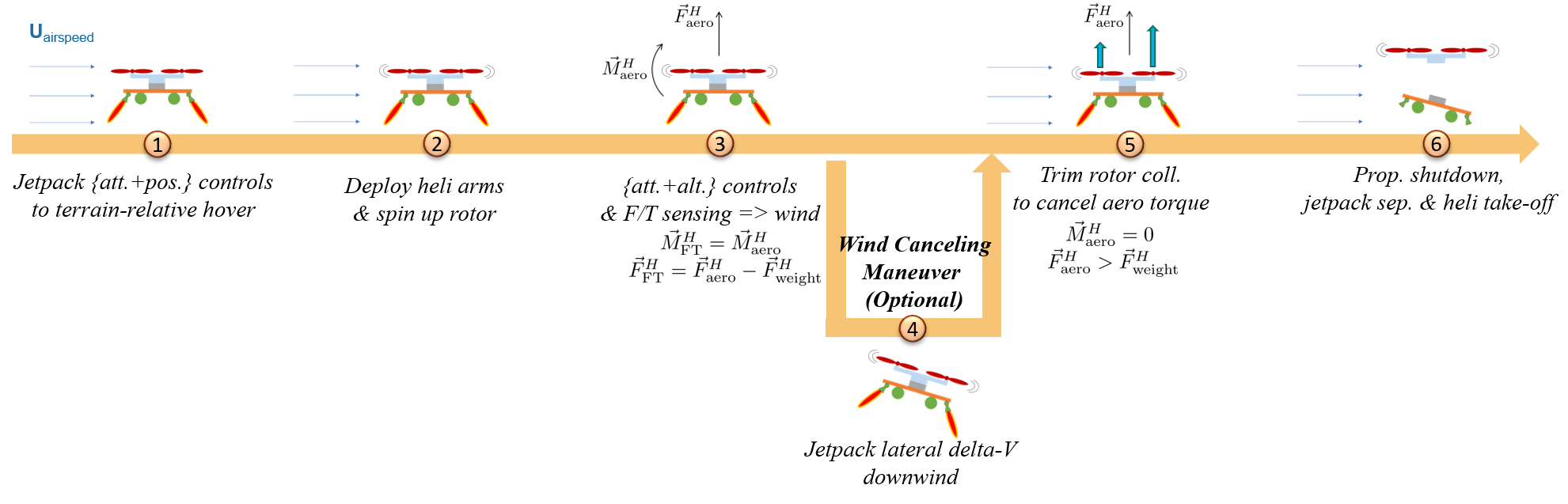}
    \caption{\bf Detailed concept of operation of MSH take-off from the jetpack.}
    \label{fig:FTconops}
\end{figure*}

Once in terrain-relative hover at 200 m AGL, the vertical airspeed is negligible but the horizontal winds can reach $\sim$30 m/s (3$\sigma$) in Eridania. To address this, the jetpack enters in a wind stabilization phase using a force-torque meter to trim the helicopter rotors and ensure a safe take-off. The operations consist in the following steps, illustrated in Figure \ref{fig:FTconops}:
\begin{enumerate}
    \item \textbf{Jetpack starts in terrain-relative hover}, closing both position (altitude \& lateral) and attitude control loops. Because rotors are not spinning yet, the wind effects are still small at this stage (few Newtons max estimate) due to the low atmospheric density on Mars. Wind will only cause a slight tilt in the jetpack.
    \item \textbf{Helicopter deploys its arms and spins up the rotors}. At this point, the rotors are commanded to a fixed RPM (Rotations Per Minute) and collective, employing a hover rotor state predicted from weather models for the  time of landing. The jetpack closed-loop controls are still active, which results in decreased jet thrust to maintain altitude and avoid climbing.
    \item \textbf{Jetpack turns on force-torque meter.}  The force-torque meter reports on the total wrench (forces and moments) at the interface between the jetpack and helicopter. This wrench is transformed into the aerodynamic wrench on the helicopter by the onboard computer. Prior to this step, the jetpack controls switch to altitude and attitude only, to minimize the aerodynamic wrench on the jetpack. This also means the assembly can drift horizontally with wind while maintaining level flight. 
    \item (Optional) \textbf{Jetpack performs a lateral delta-V maneuver.} Employing a known system model for the helicopter, the onboard planner inverts the aerodynamic wrench on the helicopter into the apparent wind direction and magnitude. If this wind exceeds control bounds for stable flight (i.e., if the helicopter rotors would saturate their controls to compensate for the aerodynamic wrench), the jetpack can re-establish position control, accelerate downwind, and retry until the airspeed reaches a comfortable value.
    \item \textbf{Trim helicopter rotors to hover}. Over the course of a few seconds, the helicopter trims the collective on each rotor to reduce the aerodynamic moment to zero and trims the measured lift to the helicopter weight plus takeoff margin; employing the force-torque meter measurement to predict these adjustments and feedforward the corrections. Simultaneously, the rotor RPM can be trimmed to the local atmospheric density. The jetpack remains in altitude \& attitude closed loop.
    \item \textbf{Propulsion shutdown and helicopter takeoff}. After trimming is complete, thrusters are shut down and the jetpack is dropped, carrying the force-torque sensor with it. At the same time instant, the helicopter trimmed to the local flow conditions is release to take off.
\end{enumerate}

\section{Mechanical Design}
\label{sec:mech}

\subsection{Jetpack Mechanical Assembly}

In order to evaluate the viability of a mid-air deployment of MSH from a jetpack, it was assumed that the {helicopter+jetpack} system would be delivered to suitable conditions from inside the 2.65-m diameter heritage aeroshell (Figure~\ref{fig:mahd-aeroshell}). The inner mold line of the heatshield was assumed to be 75 mm from the associated outer mold line geometry: this 75 mm may be sub-allocated to a combination of thermal protection material thickness, heatshield structure thickness, and allowable aeroelastic distortion. The inner mold line of the backshell was assumed to be 50 mm from the associated outer mold line: this 50 mm may be sub-allocated to a combination of thermal protection material thickness and backshell structure thickness. A 570 mm tall, 279 mm diameter cylinder was assumed for the parachute canisters at the aft side of the backshell.

Under the assumptions described above, MSH may be stowed within a limited region inside the heatshield and backshell such that the stowed helicopter blades are close to the maximum projected diameter of the heatshield and backshell: attempting to stow the helicopter either further aft or forward of this position would result in mechanical interference with the backshell or heatshield, respectively. The jetpack is assumed to stow between the stowed helicopter and the heatshield, constraining the allowable jetpack geometry.

In response to these constraints, the jetpack illustrated in Figure~\ref{fig:jetpack-assembly} was designed. It is based on a 95-cm-diameter circular platform, with a 35-cm-diameter circular porthole at the center. The porthole accommodates MSH's payload box, including the terrain sensors which are exposed in the ground direction. The jetpack is large enough to accommodate the ten fuel tanks, as well as the four thrusters. Yet, it is slim enough to fit inside the aeroshell with the helicopter. The jetpack structure is assumed to be fabricated from aluminum 6061, for which the mass was estimated to be 13.95 kg, including margins.

\subsection{Backshell Interface and Separation Mechanism}

\begin{figure}
\centering
\includegraphics[width=2.75in]{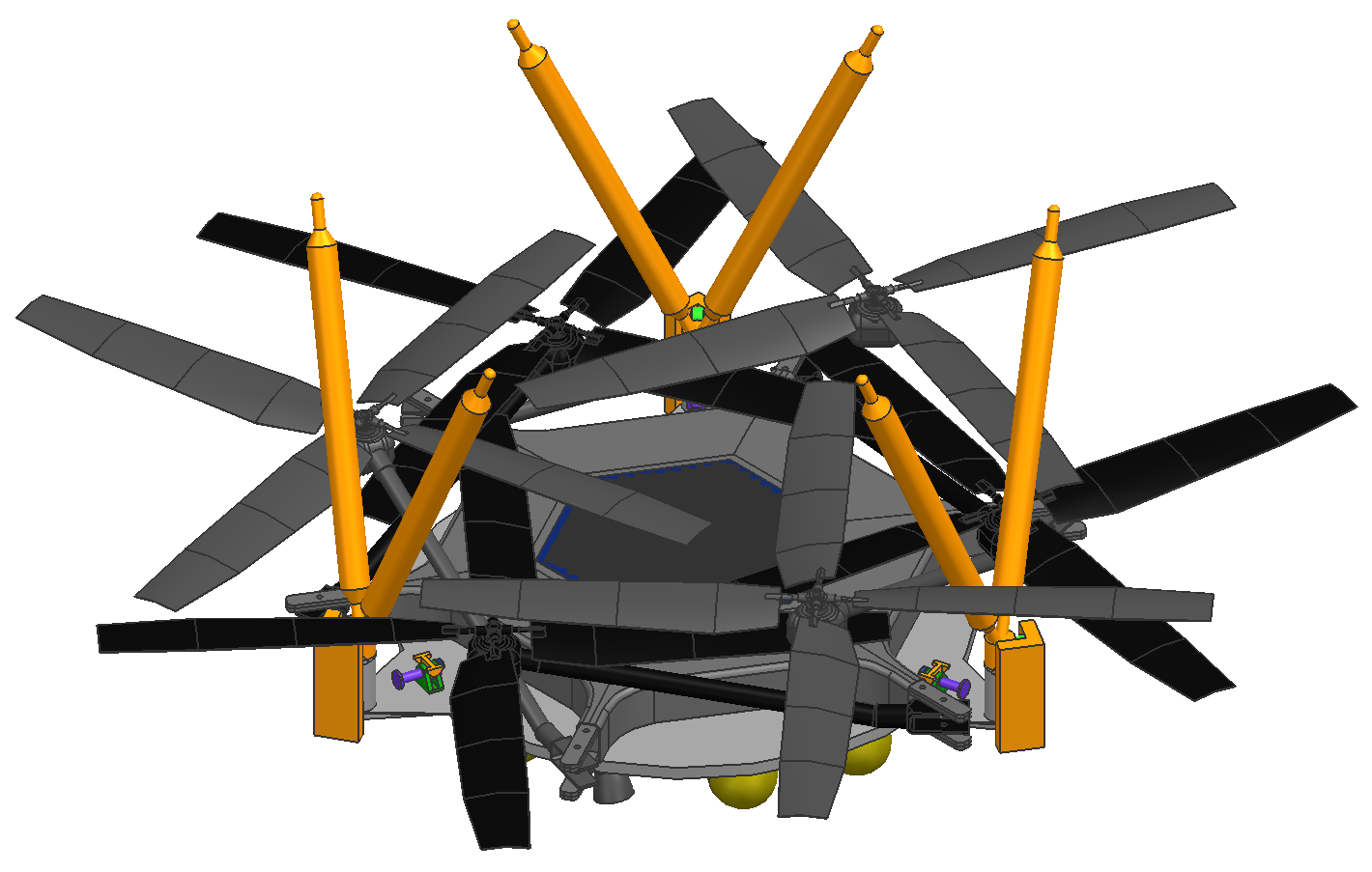}\\
\caption{\textbf{View of the stowed helicopter, jetpack, and three sets of bipods used to attach the jetpack to the backshell (backshell not shown).}}
\label{fig:stowed_msh_bipods}
\end{figure}

\begin{figure}
\centering
\includegraphics[width=2.75in]{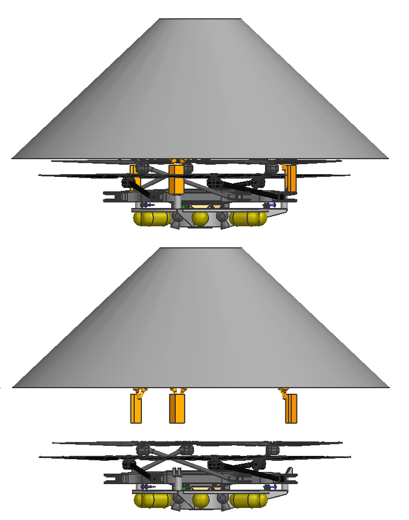}\\
\caption{\textbf{Views of the stowed jetpack + helicopter combination showing the configuration just before (top) and just after (bottom) separation from the backshell (parachute not shown).}}
\label{fig:backshell_sep}
\end{figure}

It is assumed that the jetpack would provide the mechanical connection between the backshell and the helicopter (as opposed to the helicopter directly connecting to the backshell). This is done to fulfill MAHD's design objective of minimizing potential mass impacts to the helicopter. The mechanical connection between the backshell and the jetpack is implemented using three sets of bipods (Figure~\ref{fig:stowed_msh_bipods}). Each of the three bipods attaches to the backshell at two points and attach to the jetpack at one point. The bipods include notional separation devices, push-off springs, and guide rails to ensure the jetpack + helicopter combination separates from the backshell predictably and without a significant risk of deleterious hardware contact during separation (Figure~\ref{fig:backshell_sep}).

\subsection{Helicopter Interface and Separation Mechanism}

Two methods for attaching the helicopter to the jetpack have been considered. The first method uses three mounting points between the jetpack and helicopter: at each mounting point, the helicopter presents a monoball and the jetpack presents a retractable pin. The pins pass through the holes in the monoballs, holding the jetpack and helicopter together. When the pins are retracted, the helicopter is pushed away from the jetpack by push-off springs. The pins are held in place with a frangible nut or similar device, and the pin retraction force is assumed to be provided by compressed springs. One concern with this approach is the potential for the three attachment points to separate at slightly different times, potentially introducing deleterious torques on the helicopter and/or jetpack. In order to address this concern, a single attachment method was also developed. In the single attachment method, the helicopter and jetpack present ring-like interfaces that are joined together with a circumferential restraint similar to a Marman clamp. Both the three mounting point and single attachment methods result in approximately 0.15 kg of residual mass on the helicopter associated with structure needed to implement the interface.

\subsection{Changes applied to MSH mechanical design}

MSH was initially designed to maximize flight performance and science return, while meeting the requirement to fit in an empty 2.65-m aeroshell~\cite{johnson2020}. One of the design objectives of MAHD stated in Section~\ref{sec:system} is to impose as few constraints as possible on this initial MSH design. In its latest iteration with MAHD, only minimal changes have been applied to the fuselage, rotor arms and landing legs. More details about the MSH latest design can be found in \cite{tzanetos2021aeroconf}.

Each rotor arm folds in toward the fuselage in the stowed condition illustrated in Figure~\ref{fig:msh-deployed}. The folding angle of the arms was adjusted to ensure enough clearance between the outer diameter of the rotor blades and the aeroshell, while maintaining enough clearance between each rotor blade.

MSH also includes four landing legs, which fold upward in the stowed configuration. The fastening points of the landing legs were moved from the bottom panel to the side of the fuselage, to avoid interference with the jetpack and add structural strength. Legs are 56-cm long in the current design. They could be extended up to 74 cm while still respecting the geometrical constraint of the aeroshell and jetpack. This change is currently under consideration to improve surface stability and lower the risk of rotor blades striking the ground. 

\section{Wind Dynamics}
\label{sec:wind}

We have simulated a sample Martian environment at high spatial resolution to evaluate the effect of the local wind field on MAHD. This analysis is based on the Mars Weather Research and Forecasting (MarsWRF) general circulation model (GCM)\cite{richardson2007,toigo2012} in its nested, mesoscale mode at the Eridania reference site.

MarsWRF is a global model derived from the terrestrial mesoscale WRF model~\cite{skamarock2008}, and is a Mars-specific implementation of the PlanetWRF GCM. It is a finite difference grid-point model projected onto an Arakawa-C grid with user-defined horizontal and vertical resolution. The vertical grid follows a modified-sigma (terrain-following) coordinate from 0 to $\sim$80 km with 60 layers of increasing depth with height.  

Two distinct capabilities of MarsWRF are of particular value for the present investigation as well as future studies. First, the single-framework design of MarsWRF allows the model to be run as both a global and mesoscale model within the same model architecture, and permits multiple subgrids (“nests”) at increased resolution. Because the same model is used at both the mesoscale and global scale, the physics and dynamics employed at both scales are self-consistent.  For this application, four nested model domains were incorporated, with a maximum resolution of 4.4 km per gridpoint at the highest resolution. Second, the computational framework of MarsWRF allows to easily examine regions of the surface that are traditionally difficult for GCMs to simulate. MarsWRF has the capability of orienting the geographical coordinates within the model's computational domain regions, allowing to reposition the desired surface locations at the center of the computational domain, eliminating the effects of grid stretching and distortion at the domain boundaries.

\begin{figure}
\centering
\includegraphics[width=2.75in]{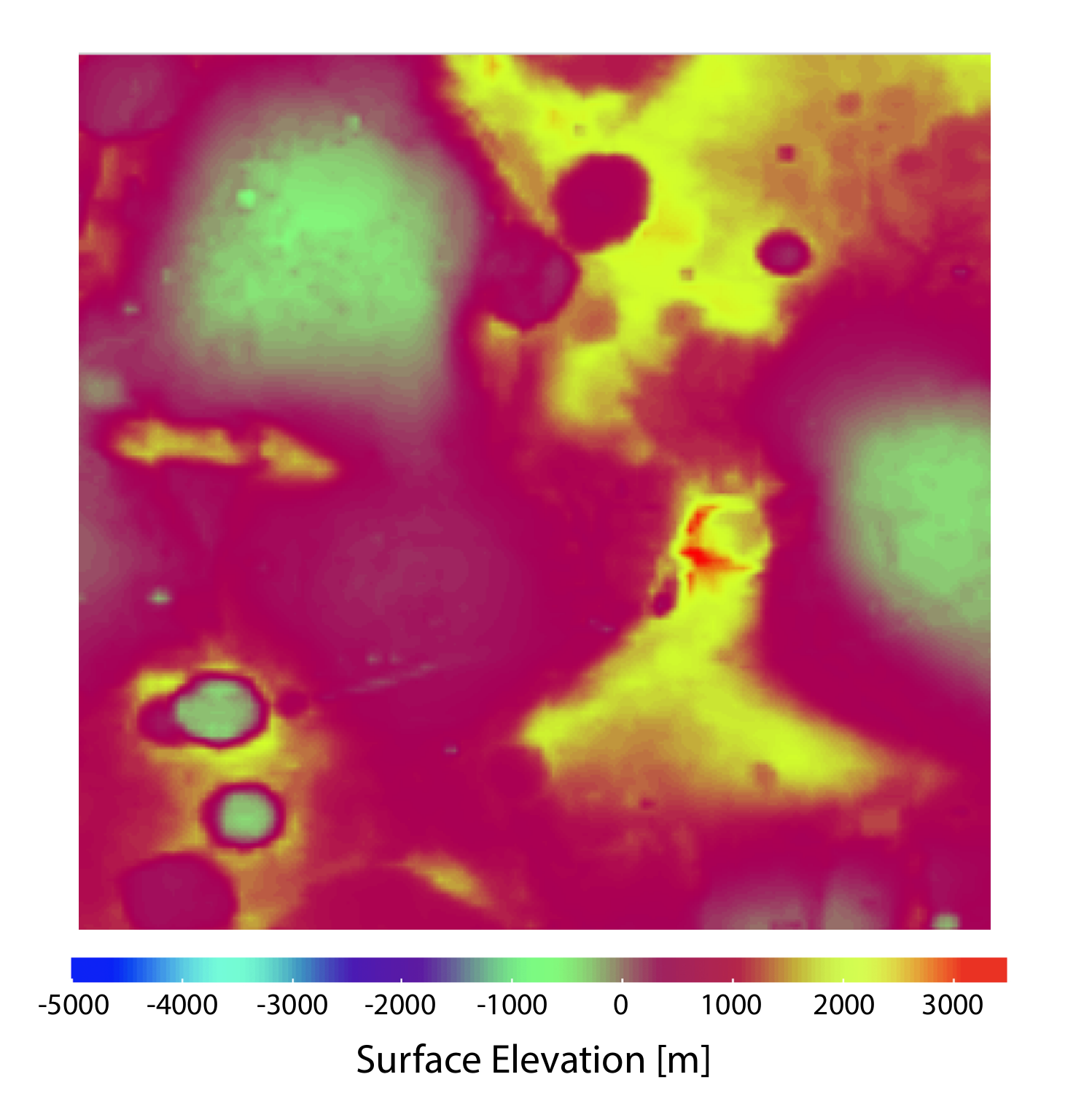}\\
\caption{\textbf{Surface elevation of the Eridania basin showing high elevation and extreme topography.}}
\label{fig:mesosocale-winds}
\end{figure}

The Eridania site, shown in Figure~\ref{fig:mesosocale-winds}, exhibits large variations in topography, and is comprised of bare regolith and bedrock with no near-surface ice. A simulation was run at MAHD's design solar longitude and time of day specified in Section~\ref{sec:system}.

MarsWRF was used to calculate the wind profile at three distinct altitudes above the local surface: 600 m AGL (backshell separation), 400 m AGL ($\sim$ jetpack powered deceleration start), and 200 m AGL (helicopter take-off). The nested model configuration is run for several Martian days to ensure steady state in the system. Data, including vertical and horizontal wind speed, is output every five minutes from MarsWRF on the model’s native vertical grid. This data is interpolated to the three desired altitudes, and various measures of the wind dynamics are produced, including peak winds, wind shear and wind gust magnitude. Of greatest interest to the MAHD project are:
\begin{itemize}
    \item 3$\sigma$ instantaneous horizontal and vertical winds,
    \item 3$\sigma$ vertical wind shear between backshell separation at 600 m AGL, and helicopter take-off at 200 m AGL,
    \item short-term temporal variations in wind gusts.
\end{itemize}
Winds were averaged over the highest resolution model domain (120x120 grid points) to provide representative winds in the EDL region.  Because of the still-coarse nature of these mesoscale runs compared to the spatial scale of interest for MAHD activities, we scaled the modeled wind speeds up by a factor of 1.5, a value derived from the statistics of mid-latitude large eddy simulations (LES) of mean-to-3-sigma wind speeds at a resolution of $\sim$10 m.  The LES results are obtained with MarsWRF as well, but are not meant to represent the physical characteristics of any specific landing site.  Simple runs of this nature were designed simply to inform the team about representative small-scale wind variability beyond what could be reproduced by the mesoscale simulation. Results for these scaled wind statistics are shown in Table~\ref{tab:winds}:
\begin{table}
\footnotesize
\renewcommand{\arraystretch}{1.3}
\caption{\bf Scaled wind statistics are the Eridania reference site.}
\label{tab:winds}
\centering
\begin{tabular}{|c||c|c|c|}
\hline
\bfseries Height AGL [m] & \bfseries 200 & \bfseries 400 & \bfseries 600 \\
\hline
\hline
\bfseries 3$\sigma$ instantaneous horizontal windspeed [m/s] & 34.1 & 33.2 & 32.6\\
\bfseries 3$\sigma$ instantaneous vertical windspeed [m/s] & 1.4 & 1.2 & 1.2 \\
\hline
\bfseries 3$\sigma$ wind diff. 600m vs. 200 m [m/s] & \multicolumn{3}{c|}{15.1} \\
\hline
\bfseries 3$\sigma$ wind diff. at 200 m over 5 s [m/s] & \multicolumn{3}{c|}{2.1} \\
\hline
\end{tabular}
\end{table}

To ensure robust helicopter take-off at 200 m AGL with MAHD, we are only interested in wind variations of the order of $\sim$10 m/s and above, which can affect rotorcraft controls. They key conclusions of this wind analysis are the wind velocity at helicopter take-off (200 m AGL) :
\begin{enumerate}
    \item cannot be predicted before reaching the take-off altitude,
    \item is predominantly horizontal,
    \item is within 34.1 m/s 3$\sigma$,
    \item can be assumed constant for $\sim$1 s for helicopter control purposes.
\end{enumerate}
\textbf{Therefore, we set the requirement that the wind measurements before take-off must be available at a frequency greater than 1 Hz.}

Future studies will provide a more rigorous assessment of specific landing sites with more stressing, high-value science locations, featuring greater topographic complexity and surface variability. Many such sites remain attractive targets for future exploration because of their unexplored environments and high science value for understanding both early and late Martian climate.

\section{Flow Dynamics}
\label{sec:flow}

\subsection{Jet Entrainment Analysis} \label{sec:JetEntrainment}
As the jetpack fires its thrusters into the Martian atmosphere, the underexpanded nature of the ensuing jets, as well as their supersonic flow speeds, present a potential challenge to the rotorcraft take-off conditions.  The flow field in the region above the jetpack take-off platform is of particular interest because it may present non-uniformity, unsteadiness, and large-enough flow velocities to affect the aerodynamics near the rotors and, hence, the stability and margins of the control system.  

\begin{figure}[hbt!]
    \centering
    \centering
    \includegraphics[width=.5\textwidth]{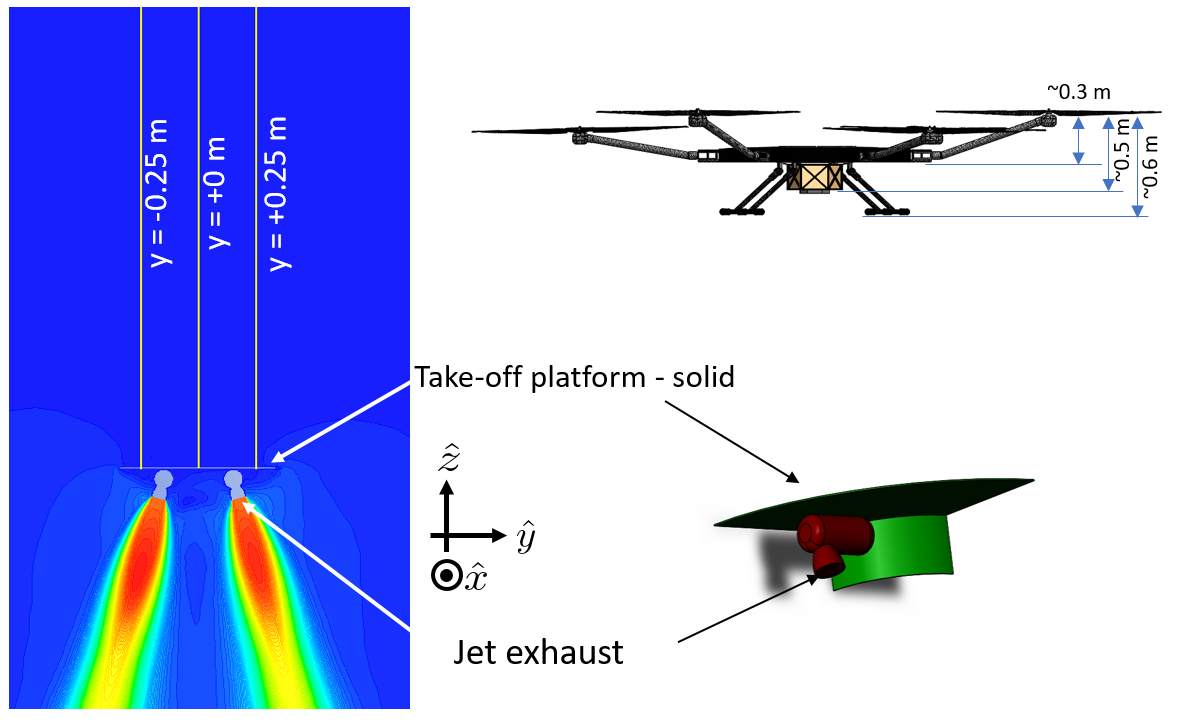}%
    \caption{\bf Example of plane contour plot at hover condition, with virtual probe line locations for velocity profiles (left).  Hexacopter illustration with rotor locations (top-right). One-quarter symmetry of geometry within simulation (bottom-right). }
    \label{fig:FlowGeometrySetup}
\end{figure}

A preliminary attempt to understand the effect the firing thrusters have on the flow field above the take-off jetpack platform consisted of numerical simulations carried out in computational fluid dynamics (CFD) software FLUENT to map the flow velocities in that region. In this first attempt, a single fluid component flow simulation consisted of carbon dioxide gas exhausted jets into a carbon dioxide Martian atmosphere considering the approximate geometry of the jetpack shown in Figure~\ref{fig:FlowGeometrySetup}. 

The thruster exhaust velocity and pressure are roughly calculated from the chamber pressure, area ratio, and maximum thrust from the Aerojet Rocketdyne MR-107N hydrazine monopropellant rocket engines, shown in Table~\ref{tab:jetpars}. They represent the higher-end of the performance in the MR-107 series and were intentionally chosen higher than the values discussed in Section \ref{sec:GNC}. The chosen pressure and velocity values are intended to provide a fair upper limiting case for the jet exhaust conditions rather than mimic the exact make and model values.    

The geometry consists of a half-symmetry of the jetpack perpendicular with the wind direction ($-\hat{y}$), yielding a total of two jets instead of four.  The mesh consists of $4\times10^6$ elements, with the boundaries spanning [150,-800] $D_{\mathrm{ex}}$ in $\hat{z}$, [500,500] $D_{\mathrm{ex}}$ in $\hat{y}$, and 400 $D_{\mathrm{ex}}$ in $\hat{x}$ ($\hat{x}$ is the symmetry direction).  The simulation solved the steady Reynolds-averaged Navier-Stokes (RANS) equations, with an ideal gas as the equation of state.  Viscosity was resolved as temperature dependent with a two-term Sutherland's law equation \cite{white2006viscous}.    

\begin{table}[hbt!] 
\centering
\caption{\bf Table parameter for jet inlet conditions.} \label{tab:jetpars}
\begin{tabular}{| c | c | c | }  \hline 
\textbf{Variable} & \textbf{Description} & \textbf{Value} \\ 
\hline \hline
 $P_0$ & Chamber Pressure [Pa] & 1.1170e+06 \\ 
 $T_0$ & Chamber Temperature [K] & 1300 \\   
 $A/A^*$ & Area Ratio [ND] & 20.7 \\  
 $D_{\mathrm{ex}}$ & Exhaust diameter [m] & 0.066\\  
 $M$ & CO$_2$ Molar Mass [kg/mol] & 0.04401\\  
 $\gamma$ & CO$_2$ Ratio of Specific Heats [ND] & $1.289$ \\  	
 $U_{\mathrm{ex}}$ & Exhaust Velocity [m/s] & 1200\\  	
 $P_{\mathrm{ex}}$ & Exhaust Pressure [Pa] & 4300 \\	
 \hline					
 \end{tabular} 
\end{table}

\begin{figure}[hbt!]
    \centering
    \centering
    \includegraphics[width=.5\textwidth]{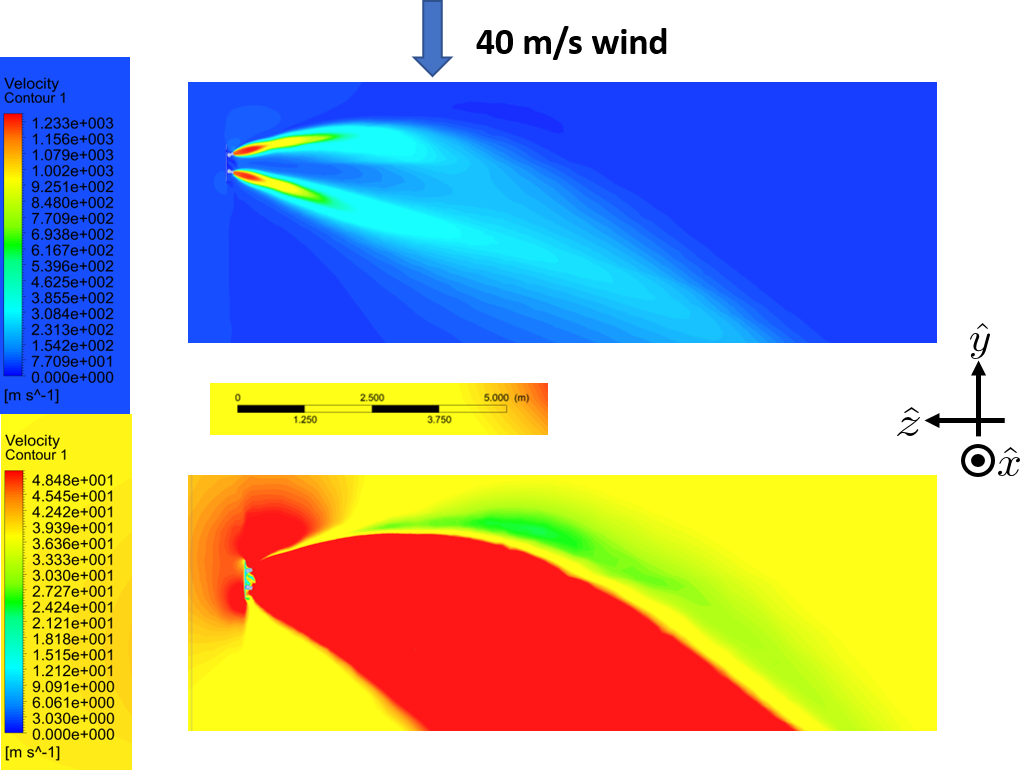}%
    \caption{\bf Contour plots for full velocity range (top) and saturated range of 0-50 [m/s] (bottom) for wind speed of 40 [m/s].}
    \label{fig:FlowContourPlots40}
\end{figure}

\begin{figure}[hbt!]
    \centering
    \centering
    \includegraphics[width=.5\textwidth]{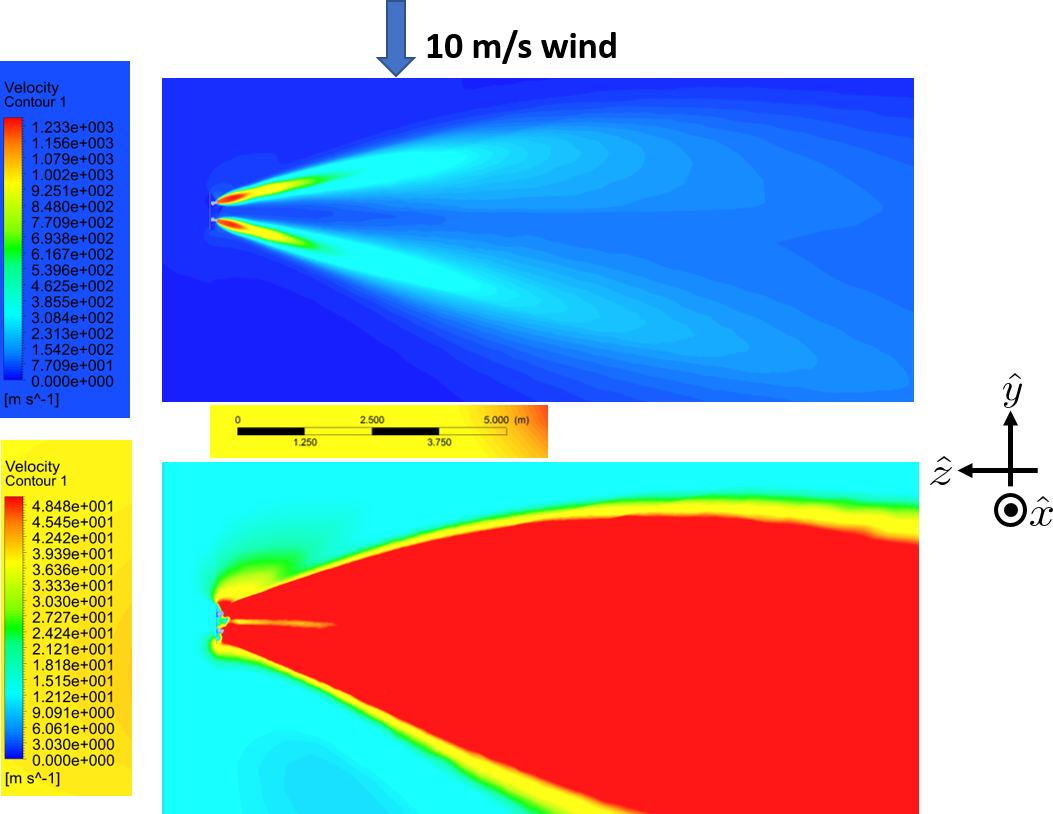}%
    \caption{\bf Contour plots for full velocity range (top) and saturated range of 0-50 [m/s] (bottom) for wind speed of 10 [m/s].}
    \label{fig:FlowContourPlots10}
\end{figure}

A number of cases were run with uniform incident wind velocity in the $-\hat{y}$ direction at values of 0, 10, 20, 30, and 40 [m/s]. Contour plots for 40 [m/s] and 10 [m/s] are shown in Figures \ref{fig:FlowContourPlots40} and \ref{fig:FlowContourPlots10}, respectively.  Contours are taken at a plane parallel to the half-symmetry plane and cut through the center point of the two jets. The full range contours show the high speed section at the core of the jets, and give a notion of the magnitude of the jet velocity as compared to the surrounding flow field induced by the wind. The saturated contours span the jets, but also show the flow field in the areas of interest for the rotorcraft take-off on the left (i.e., region above take-off platform).  
The flow asymmetry about the jetpack center axis ($\hat{z}$) becomes clear and more pronounced when wind speeds are increased.  

\begin{figure}[hbt!]
    \centering
    \centering
    \includegraphics[width=.45\textwidth]{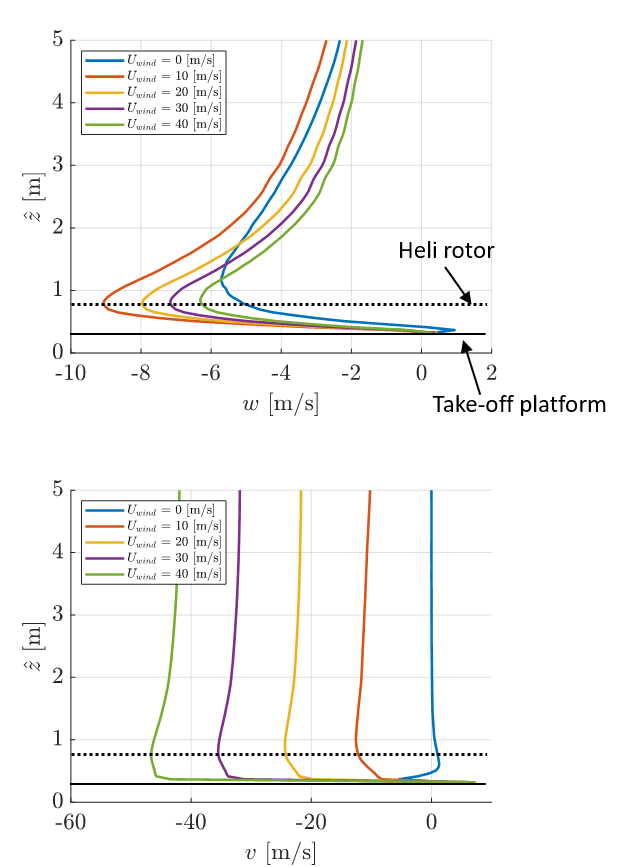}%
    \caption{\bf $\hat{y}$ and $\hat{z}$ velocity components, $v$ and $w$ respectively, as a function of vertical position in $\hat{z}$ at $\hat{y} = 0.25$ [m].}
    \label{fig:ProfilePlot0p25}
\end{figure}

\begin{figure}[hbt!]
    \centering
    \centering
    \includegraphics[width=.45\textwidth]{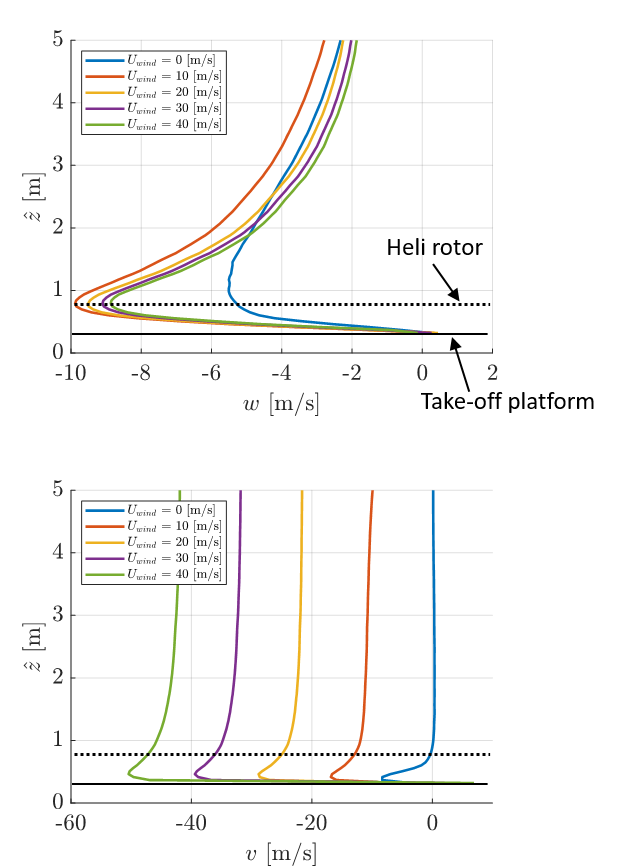}%
    \caption{\bf $\hat{y}$ and $\hat{z}$ velocity components, $v$ and $w$ respectively, as a function of vertical position in $\hat{z}$ at $\hat{y} = -0.25$ [m].}
    \label{fig:ProfilePlot-0p25}
\end{figure}

Asymmetry in the flow field can be further examined by probing the $\hat{y}$ and $\hat{z}$ velocity components, $v$ and $w$, respectively, at the two distinct $\hat{y}$ locations shown in Figure~\ref{fig:FlowGeometrySetup} ($\hat{y} = [0.25, -0.25]$ [m]), with profiles in Figures~\ref{fig:ProfilePlot0p25} and \ref{fig:ProfilePlot-0p25}.  In particular, $w$ is considerably faster towards the trailing edge of the take-off platform than the leading edge. The profile over increasing wind speeds appears to asymptote to a profile shape that is best captured by the 40 [m/s] case results. 
As wind speed increases past 10 [m/s], the peak values of $w$ decrease, while values of $v$ increase significantly, tracking the wind velocity.  For all cases, the ``peaky'' profile shape appears to hold, with a quickly accelerating region near the platform, followed by a slowly decreasing section towards the free stream. 
The peaks seem to approximately correspond to $\hat{z}$ location of the rotorcraft rotors, both in the $v$ and $w$ components.  

In summary, $w$ appears constrained to less than 10 m/s downward, which is compatible with helicopter take-off. However, $v$ is amplified relative to the free-stream wind velocity, with the peak amplification at the rotor positions (per Figures~\ref{fig:FlowGeometrySetup}, \ref{fig:ProfilePlot0p25}, and \ref{fig:ProfilePlot-0p25} ).  This feature, in addition to the asymmetry in the profile, may present unfavorable conditions for the rotorcraft take-off.  Further study is being conducted to understand the coupled flow field between jet entrainment and rotor downwash and its impact to the rotorcraft control system.

\subsection{Rotor-Jet Interaction Preliminary Studies}

\begin{figure}[hbt!]
    \centering
    \centering
    \includegraphics[width=.45\textwidth]{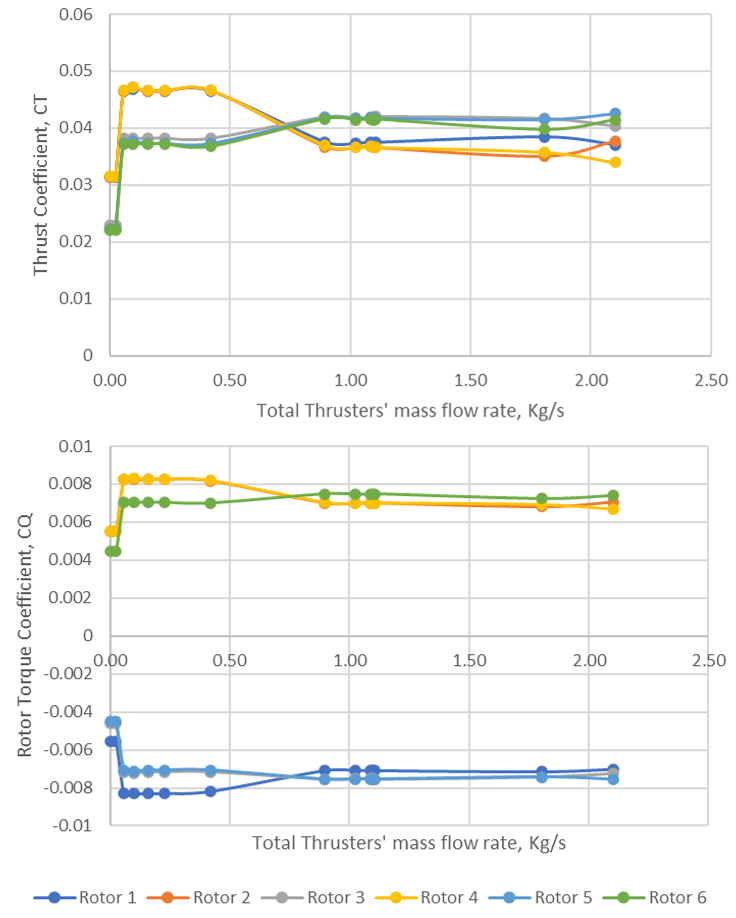}%
    \caption{\bf Preliminary results for hexacopter rotors thrust and torque (with interactional aerodynamics effects) as a function of the jetpack thruster mass flow rate.}
    \label{fig:RotCFDTorqueMoments}
\end{figure}

\begin{figure}[hbt!]
    \centering
    \centering
    \includegraphics[width=.3\textwidth]{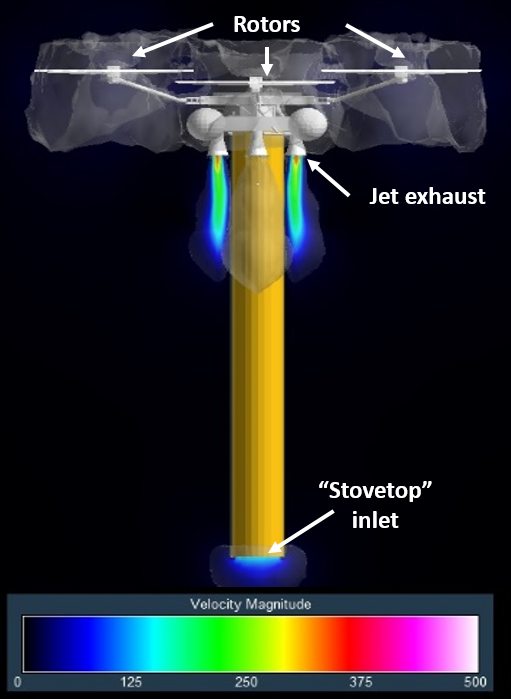}%
    \caption{\bf Rotor wake and jet flow interactional aerodynamic flow field predictions for non-canted nozzles: zoomed our view of the rotorcraft/jetpack and ``stovepipe'' duct.}
    \label{fig:StoveTopRotCFD}
\end{figure}

RotCFD is a hybrid mid-fidelity CFD tool that models rotors using actuator disk, or lifting line, and blade element theory modeling. The rest of the flow field is computed using an unsteady RANS flow solver~\cite{rajagopalan2012rotcfd,novak2015rotcfd,rajagopalan2018potential}.  
Consequently, a direct study of the novel rotor wake and jet flow interactional aerodynamics has been performed, albeit, so far, at reduced thruster jet exit velocities and mass flow rates compared to the FLUENT studies (see Figure~\ref{fig:RotCFDTorqueMoments}).   
The jet flow in RotCFD is not modeled by means of a single (per thruster jet) momentum source but, instead, relies on an ad-hoc approach to provide flow to the thruster nozzles by means of virtual mini rotors embedded in an internal duct and plenum (denoted herein as ``turbo-pump rotors'') that then feeds the thruster jets.  Three sets of RotCFD rotor wake modeling efforts are currently being performed: thruster jet nozzles not canted (Figure \ref{fig:StoveTopRotCFD}), canted nozzles (15 $^{\circ}$), and a foundational one-rotor and one-jet parametric model. Note that the jetpack geometry in the figure is not representative of the current design.  
This effort is being carried out to better understand the limitations of the tool and gather insights of the general jet-rotor interaction problem, and will serve as a starting point for future studies that will address the problem in more detailed physics, numerical, and parametric fidelity. RotCFD was also used to provide preliminary estimates of the jetpack pitching moments with regards to the helicopter take-off operation discussed in Section~\ref{sec:takeoff}. 

\section{Jetpack Guidance, Navigation and Control} \label{sec:GNC}

\subsection{Jetpack GNC Architecture}

\begin{figure}[h]
    \centering
    \includegraphics[width=0.5\textwidth]{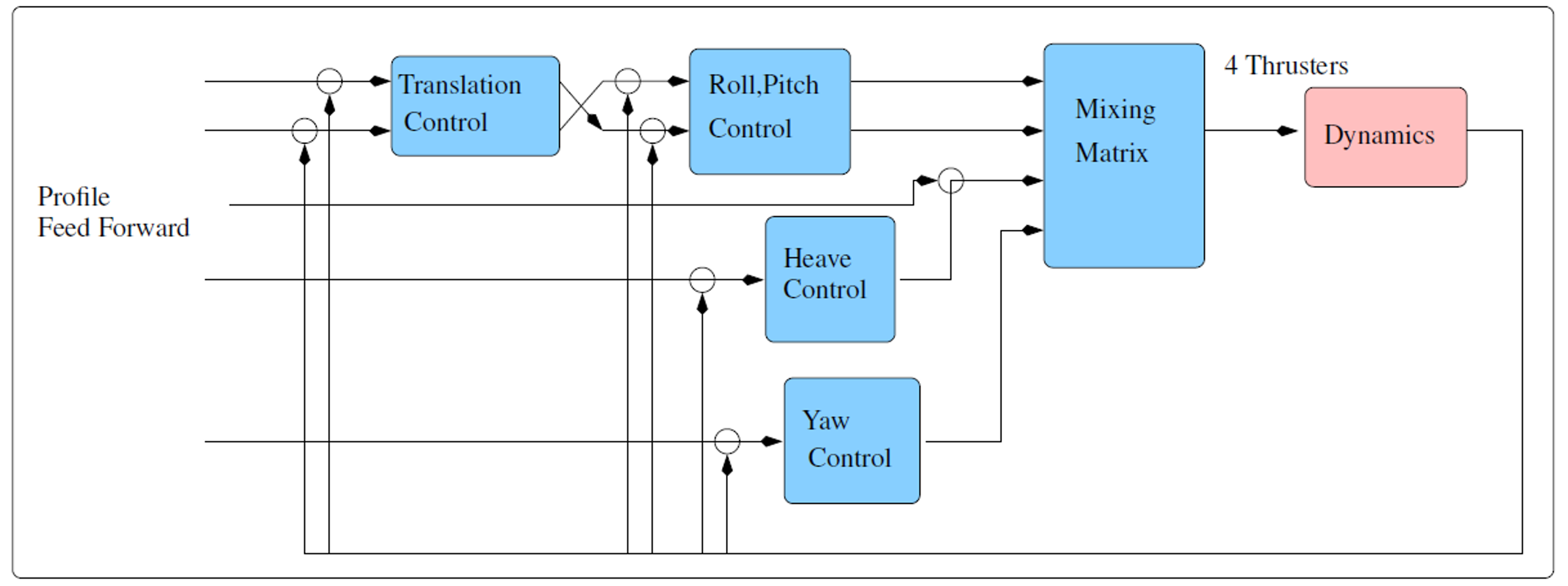}
    \caption{\bf Controller Block Diagram}
    \label{fig:controller}
\end{figure}

The control architecture has heritage from prior Mars missions. It is a nested loop of two controllers. The inner, high-bandwidth loop controls heave and roll, pitch, and yaw. The outer, low-bandwidth loop controls transverse translation. There is no gain limit, 54 deg of phase margin, a 3.7-Hz crossover frequency for the inner loop; and 8-dB gain margin, 45-deg phase margin, and a 0.27-Hz crossover frequency for the outer loop. This control architecture performs the divert and hover phases of the mission, with the reference trajectory originating from a polynomial guidance algorithm. Figure \ref{fig:controller} shows the block diagram for how these loops interact and provide thruster firing commands.
An alternate controller with inner and outer loop crossover of 0.37 Hz and 0.1 Hz,
respectively, was developed for use as the simulation transitions to lower sample rates with pulse width modulation.

The polynomial guidance algorithm has been adapted from heritage algorithms from the Curiosity and Mars 2020 missions, while the navigational uncertainties have been scaled from the MSH performance~\cite{delaune2020aero}. The guidance law specifies initial and final altitudes, a divert distance for backshell separation safety, and a thrust level allocation for a given duration of maneuver time. The implementation of the algorithm was validated against Curiosity mission results prior to inclusion in the preliminary MAHD simulation. In this environment, the guidance polynomials are generated at the start of the simulation with the corresponding reference and feedforward terms computed at each simulation time step.

\begin{figure*}
\centering
\includegraphics[width=7in]{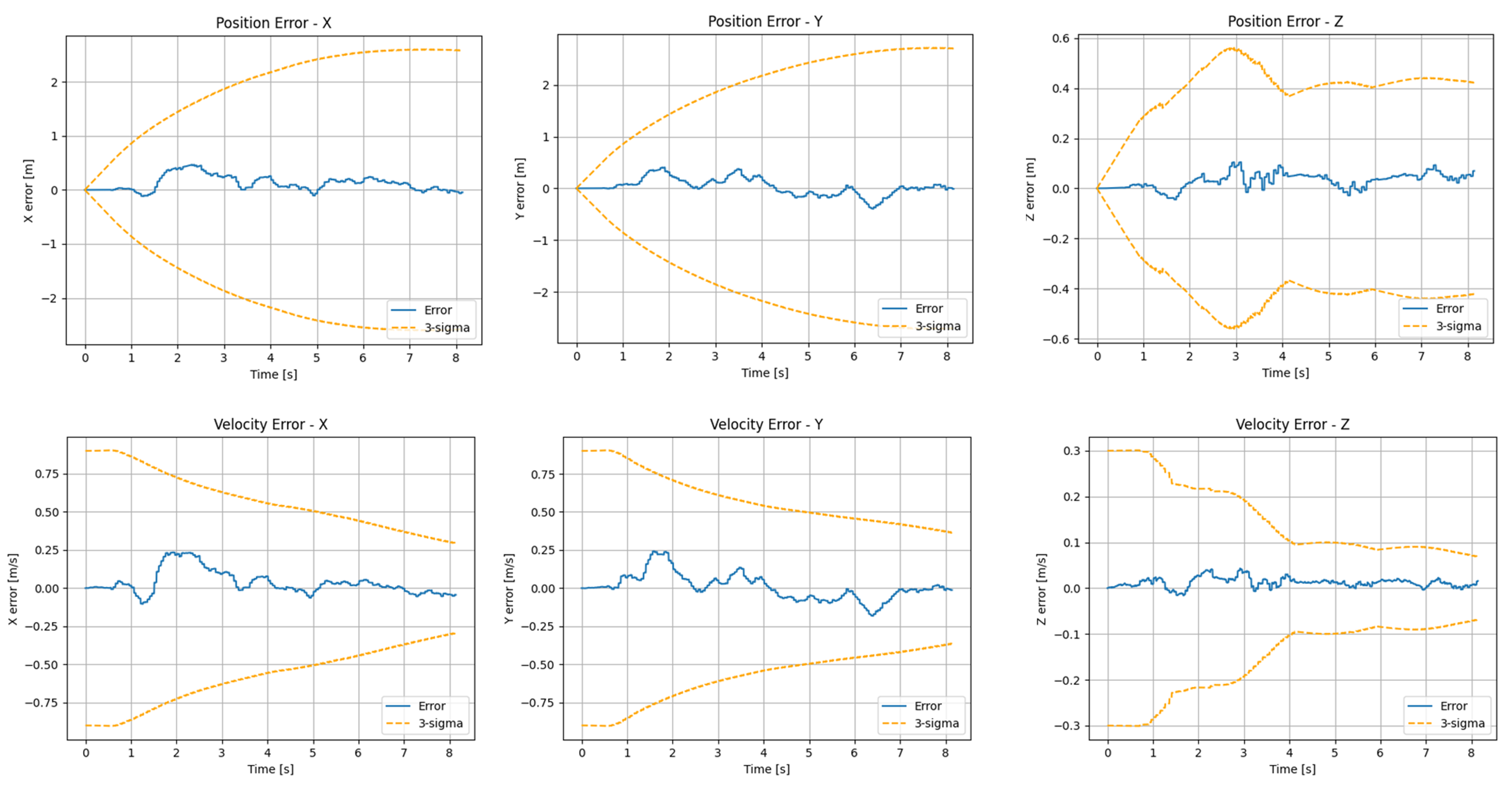}
\caption{\bf{Terrain-relative position and velocity estimation error and estimated uncertainty for the MSH vision-based navigation system during the jetpack powered phase. This was evaluated in high-fidelity point-based simulation with representative sensor noise, and initial uncertainty assumed from a batch vision-based initialization method.}}
\label{fig:msh-nav}
\end{figure*}

The jetpack uses terrain-relative position and attitude estimates coming from the range-visual-inertial navigation filter of MSH. A detailed description and performance analysis of the algorithm on real data can be found in \cite{delaune2020aero} and \cite{delaune2021}. Performance analyses to date have been limited to heights around 10 m AGL. At higher altitude, the larger pixel footprint is expected to affect the translational accuracy for position and velocity estimation. To characterize the impact of this for MAHD, a vertical deceleration trajectory similar to the jetpack powered phase we simulated: 43 m/s at 374 m AGL $\rightarrow$ 0 m/s at 200 m AGL. The simulation used representative sensor noise for the image feature tracks, altimeter and the IMU. The initial filter uncertainty was defined assuming a batch vision-based initialization method, although that may be revisited in future work. The navigation error and the 3$\sigma$ uncertainty estimated by the filter are shown in Figure~\ref{fig:msh-nav}.

In the current design, the jetpack fully relies on the MSH avionics through an umbilical connection. This satisfies the design driver to keep cost and complexity to a minimum, but this may be revisited this in future work if jetpack-specific constraints appear on the avionics. MSH features a Qualcomm-based computer architecture for computationally-intensive / low update rate operations (e.g., vision-based navigation updates), coupled with a flight controller for high-rate operations (e.g., IMU integration). This architecture is similar to Ingenuity's~\cite{balaram2018scitech}. All GNC software for MAHD runs on the MSH computers. The notional navigation sensors currently assumed for MAHD are the Ingenuity navigation camera~\cite{balaram2018scitech}, a Sensonor STIM300 IMU, and a Jenoptik DLEM~20 laser range finder.

\subsection{Preliminary Simulation of Jetpack Software}
Simulating the MAHD jetpack was first performed using the Small Satellite Dynamics Testbed (SSDT) MATLAB/Simulink simulation, which affords the ability to implement the GNC architecture in a closed-loop control simulation environment. This simulation has been used in the past for developing and testing guidance, navigation, and control software for spacecraft in low Earth orbit and in deep space, but not for planetary body surface operations ~\cite{sternberg2018ssdt} ~\cite{sternberg2021ssdt}. Despite this environmental difference, the simulation provides the necessary validated hardware and software modules to enable the rapid creation of a MAHD jetpack GNC simulation for creating the initial control architecture of the jetpack’s flight, from initial backshell deployment through final impact with the Martian surface.

The modifications to the SSDT simulation to suit the MAHD concept focused on creating a Martian surface environment. Steps included adjusting the central body to be Mars instead of Earth by adjusting the atmospheric and wind properties, environmental force and torque disturbances, gravity level, and reference frames. Additionally, the simulation was modified to run at 64 Hz to model a possible flight software architecture. These changes created the environment in which a MAHD jetpack could be simulated.

The jetpack itself was modeled in the simulation as a set of GNC hardware, a set of software elements, and a basic structural model. The simulation is comparatively of low fidelity in regard to structural modeling: the jetpack is considered to be a monolithic structure with rigid body dynamics, a rigidly attached body representing the helicopter, and an outer surface defined by a set of basic envelope faces. This specific simulation does not model the helicopter flight dynamics, which is ongoing work in extending Ingenuity's HeliCAT simulation software \cite{grip2019scitech}. The model of the jetpack’s structure does not change over the duration of the simulation, but the fuel burned does decrement the total mass and inertia of the jetpack based on the assumption that the decrease in mass occurs uniformly across the entirety of the structure.

This simulated jetpack is the representation of the physical structure that contains the planned hardware that will sense and actuate in the Martian environment. The hardware models used in the SSDT simulation have undergone prior validation and verification efforts using both analyses and comparison to flight data. The modular and parameterized nature of these hardware elements allows them to be incorporated, as necessary, into various project simulations. For MAHD, the hardware includes a set of variable-thrust thrusters, an Inertial Measurement Unit (IMU), and a position sensor akin to a helicopter vision-based navigation system. The error models for the navigation sensor were fitted onto the high-fidelity navigation simulation results for Mars Science Helicopter vision-based navigation described  in the Jetpack GNC subsection. Table \ref{table:thruster_properties} lists the key properties for the jetpack thrusters. Note that the initial SSDT simulation uses a continuous thruster model for these preliminary simulations, while the team is developing the higher-fidelity simulation using a pulse-width-modulated thruster model for future studies.

\begin{table}
\begin{center}
\begin{tabular}{|c | c|} 
 \hline
 Property & Value \\ [0.5ex] 
 \hline\hline
 Initial Tank Pressure & 2.861e+6 Pa \\ 
 Tank Volume & 10.8 L \\
 Fuel Density & 1.004 kg/L \\
 Fuel Volume & 6.74 L \\
 Thruster Operation & \begin{tabular}{@{}c@{}}Blowdown \\ Continuous Thrust\end{tabular} \\
 Thruster Cant Angle & 15 deg \\
 Thruster Resolution & 0.00001 N \\
 Maximum Thrust & 267 N \\
 Isp & 220 sec \\
 Thruster Control Frequency & 64 Hz \\
 \hline
\end{tabular}
\caption{\bf Thruster Properties in Initial Simulation}
\label{table:thruster_properties}
\end{center}
\end{table}

\subsection{Preliminary Monte Carlo Results}

The MAHD entry and descent analysis is conducted with the DARTS/DSENDS simulation software (Dynamics Simulator for Entry, Descent and Surface landing)~\cite{Jain2019DARTSM}. The EDL simulation is developed from atmospheric entry interface point at R=3522.2 km, followed by hypersonic deceleration phase. An 11.8-m supersonic disk-gap-band parachute is nominally deployed at 400 m/s (Mach 1.55).  Once the lander reaches subsonic conditions, a heatshield is jettisoned at 12 km AGL (see Figure~\ref{fig:conops-detailed}). Based on the outlined concept of operations, a DSENDS Monte Carlo simulation was completed in order to analyze the stressing initial conditions at jetpack jettison event.

The jetpack closed-loop control simulation environment described in the previous subsection was used to run a series of Monte Carlo simulations as a way to assess the feasibility for the control architecture to cope with varying operational conditions and to provide a baseline performance to be used as a comparison dataset during the development of the subsequent, higher-fidelity simulation. The Monte Carlo simulation data produced key measures of jetpack performance across the varying operating conditions, including fuel use, rotational and lateral position histories, and controller metrics. The Monte Carlo data shown in this paper draws random initial rotation rates from a uniform distribution between -2 deg/s to +2 deg/s about each axis, the initial altitude from a normal distribution with mean 650 m and standard deviation 1 m, the initial vertical velocity from a normal distribution with mean -41.7174 m/s and standard deviation 3.164 m/s, and the initial lateral velocity from a normal distribution with mean 7.6735 m/s and 7.6499 m/s standard deviation, though capped to 24.3 m/s (95th-percentile) to avoid unrealistically large lateral velocities. 500 runs are shown in the following figures.

\begin{figure}[h]
    \centering
    \includegraphics[width=0.5\textwidth]{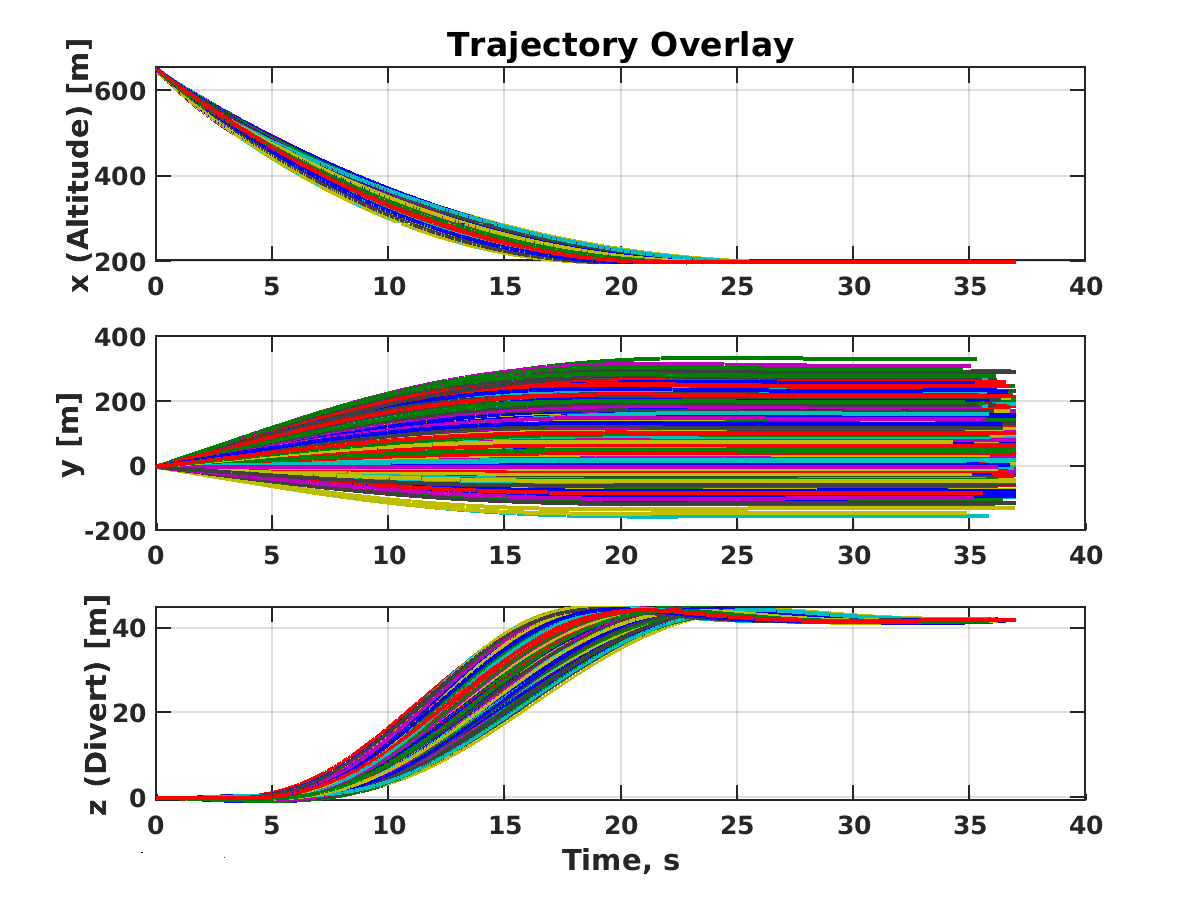}
    \caption{\bf Trajectory Overlay for All Monte Carlo Simulations}
    \label{fig:furball}
\end{figure}

\begin{figure}[h]
    \centering
    \includegraphics[width=0.5\textwidth]{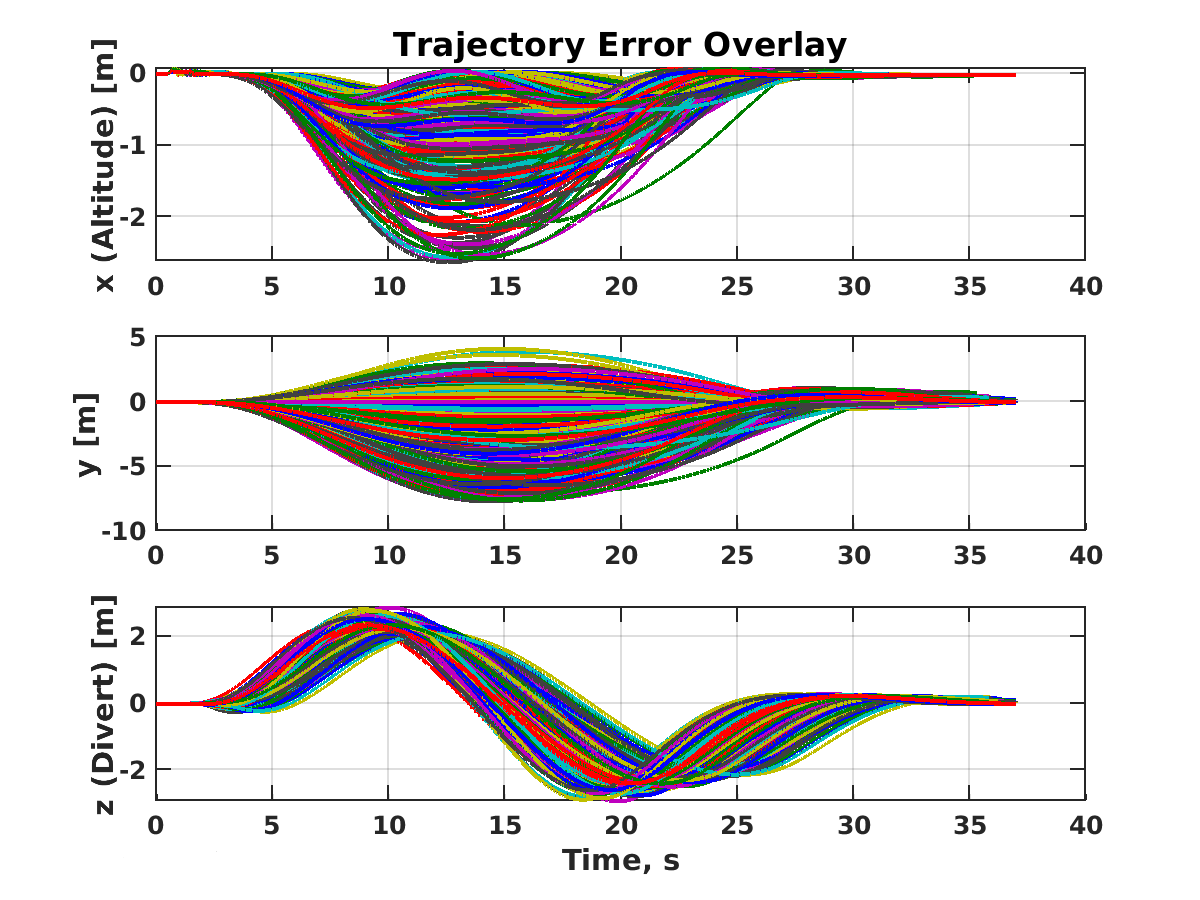}
    \caption{\bf Position Error Overlay for All Monte Carlo Simulations}
    \label{fig:perr}
\end{figure}

\begin{figure}[h]
    \centering
    \includegraphics[width=0.5\textwidth]{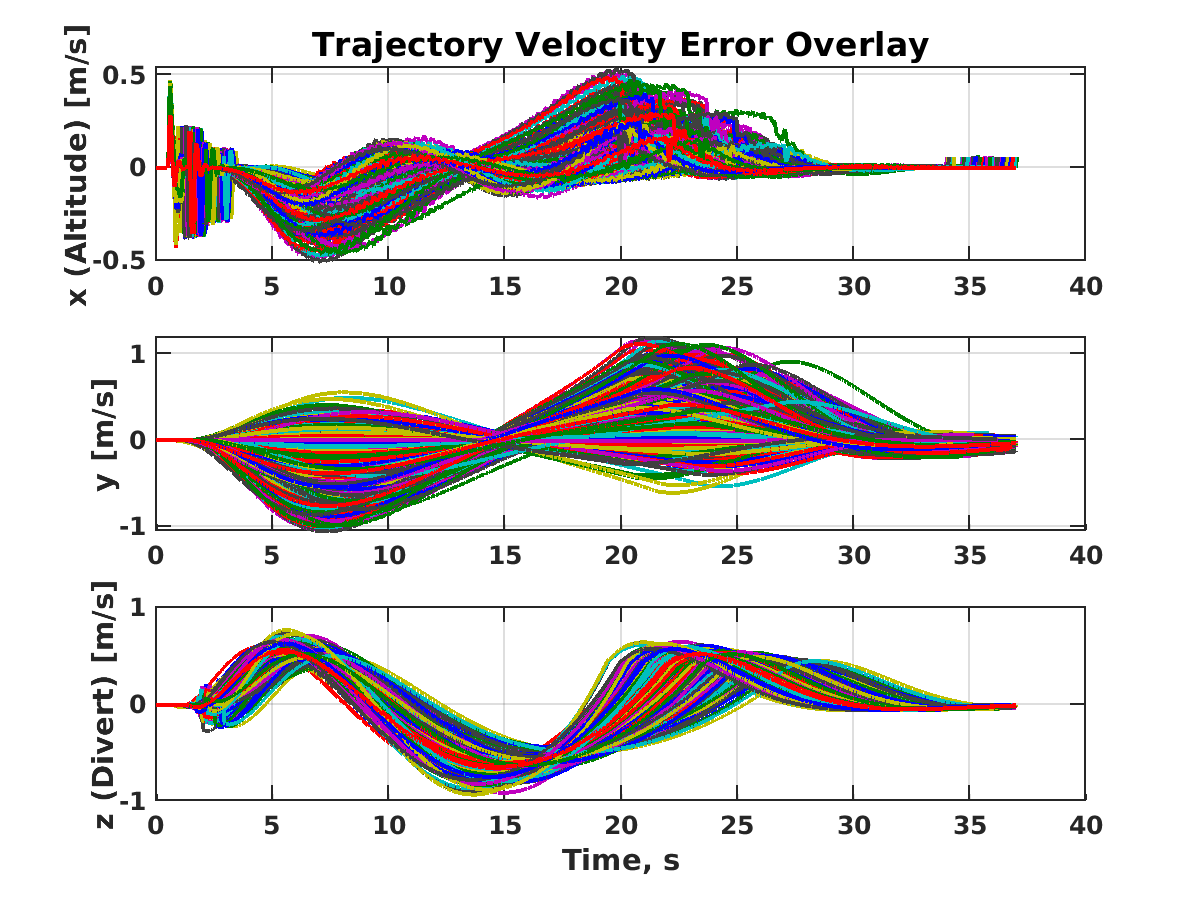}
    \caption{\bf Velocity Error Overlay for All Monte Carlo Simulations}
    \label{fig:verr}
\end{figure}

Figure \ref{fig:fuel_required} shows the resulting fuel required to complete the divert maneuver away from the backshell and achieve a hover platform during 12 seconds for the helicopter to spin up and lift off. The fuel use matches initial estimates, and the final altitude centers around the 200 m desired altitude. The full motion profile is shown in Figure \ref{fig:furball}. The top subplot shows the altitude, the bottom subplot shows the lateral displacement in the divert (crosstrack), and the middle subplot shows the lateral displacemnet along-range. The three subplots together show that the controller was able to closely match the prescribed polynomial guidance curve for height and divert distance (to the desired 42 m), while the crosstrack motion shows the effects of the controller working to compensate for the initial velocity distribution. Visible across these three plots are the separate phases of the jetpack's flight. First, the initial downward motion is arrested while diverting to reach a safe distance away from the backshell; this process takes about 20 seconds. Next, the jetpack maintains its altitude during the constant-altitude hover. During this time, the polynomial guidance law is still being used to create the controller reference. As a result, while the altitude is maintained, the lateral position still is allowed to remain at the jetpack's crosstrack direction. These phases would be followed by the jetpack cutting its thrusters as the helicopter flies away.  Before helicopter take-off, the residual angular rates of the closed-loop-controlled jetpack were under 0.2 deg/s (3$\sigma$). The position and velocity errors are shown for each axis in Figure \ref{fig:perr} and Figure \ref{fig:verr}, respectively.

The Monte Carlo capabilities of the initial simulation will continue to be refined as additional features are added to the initial simulation to help make controller architecture upgrades to the mission.

\begin{figure}[h]
    \centering
    \includegraphics[width=0.5\textwidth]{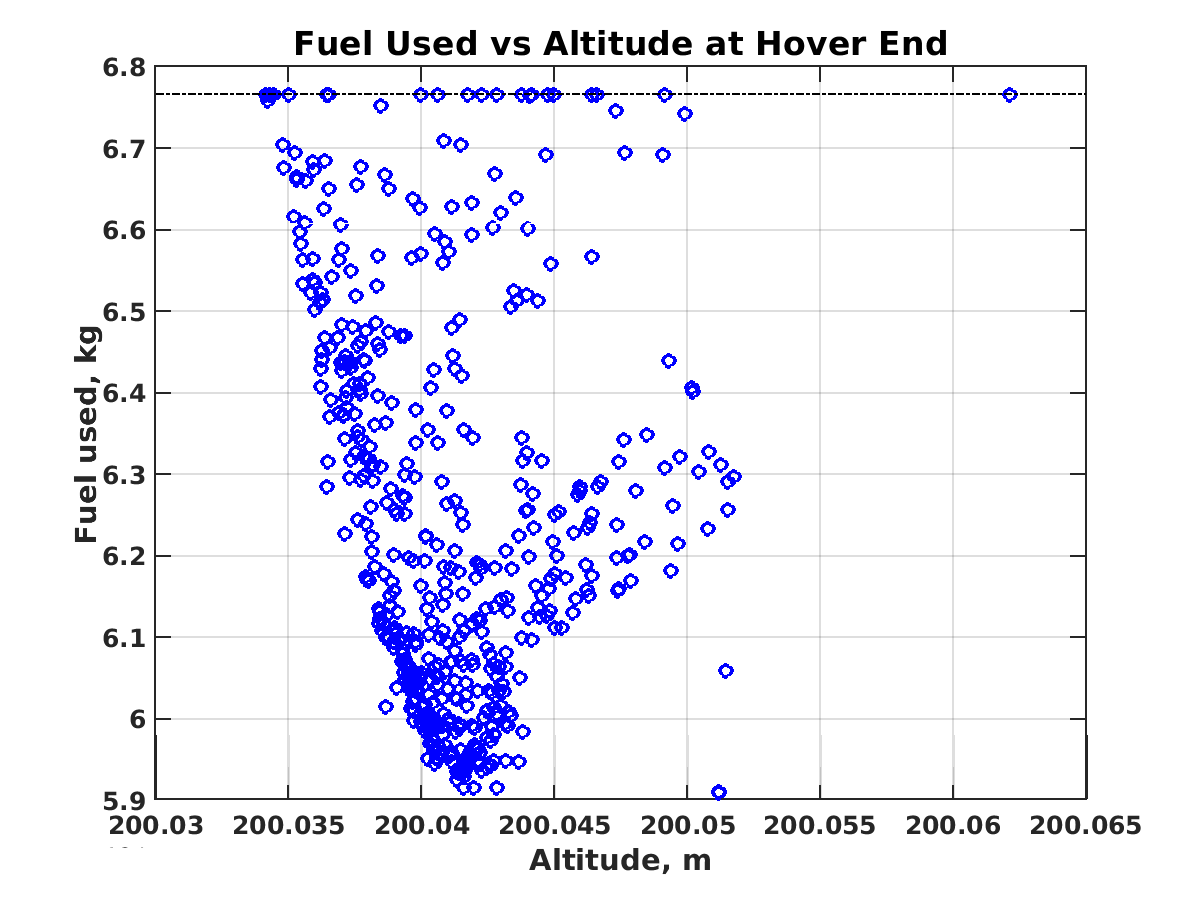}
    \caption{\bf Fuel Mass Required vs. Hover Ending Altitude Monte Carlo Results}
    \label{fig:fuel_required}
\end{figure}

\section{Helicopter Take-Off} \label{sec:takeoff}

The take-off maneuver starts when the propulsion system of the jetpack is shut down and the helicopter separates. These two events should ideally happen as close as possible to each other, at E+431s on Figure~\ref{fig:conops-detailed}. When the jetpack freefalls, the helicopter will encounter not only lateral winds, but also any remaining downwards flow entrainment left by the thrusters. The uncertainty around this flow condition might cause the helicopter to be uncontrollable, or recontact with the jetpack. These are significant risks compared to a nominal take-off from the ground. To address this, MAHD employs a force-torque meter to directly measure the total forces and moments on the helicopter before the instant of release.

\begin{figure}
\centering
\includegraphics[width=3.25in]{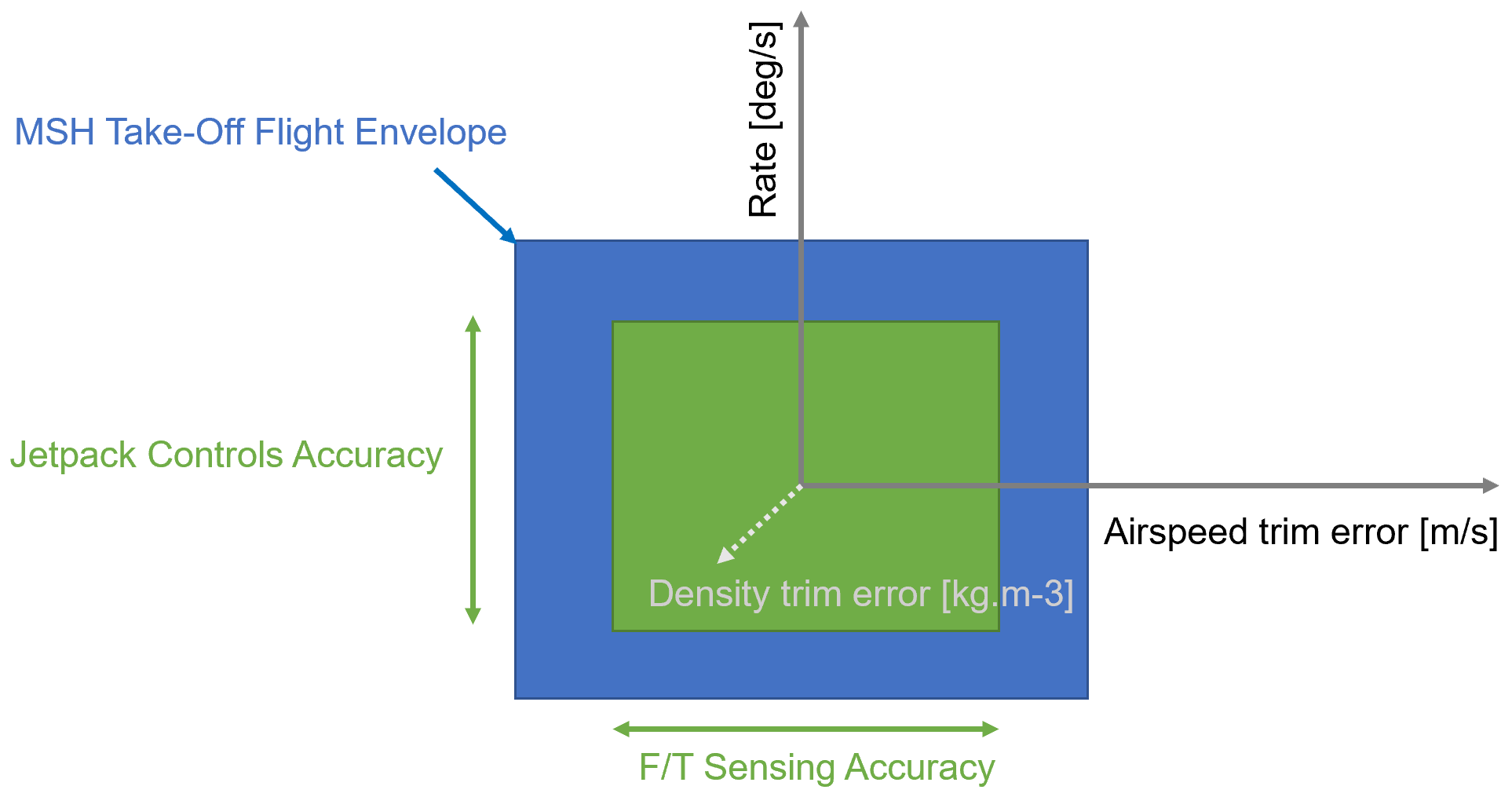}\\
\caption{\textbf{In MAHD, the MSH flight envelope at take-off (in blue) is driven by the airspeed trim error, the rotation rate, and to a lesser extent the density trim error. To guarantee safe take-off, MAHD's jetpack controls and force-torque sensing accuracy (in green) must lie within the MSH flight envelope.}}
\label{fig:takeoff-vnv}
\end{figure}

As illustrated in Figure~\ref{fig:takeoff-vnv}, MSH flight envelope at take-off is primarily driven by the airspeed trim error and the residual rotation rates on the jetpack. In Section~\ref{sec:GNC}, we demonstrated control rate error smaller than 0.2 deg/s (3$\sigma$) in simulation. Thus, to establish the preliminary feasibility of MAHD's take-off architecture, this section will provide an analytical analysis of the force-torque sensing dynamics, discuss candidate sensor accuracy, and simulation results of the helicopter control performance at take-off.

\subsection{Dynamic Analysis}

The primary advantage of the force-torque meter is the removal of uncertainty about the aerodynamic state of the rotor, which is what most directly affects the vehicle dynamics. Direct wind sensing methods were also considered, but ultimately could not be related to the rotor state: for example, wind sensing from a boom only provides a small snapshot of the highly non-uniform \& temporal flow field. A force-torque meter with six degrees of freedom, however, can measure the total aerodynamic wrench on the helicopter from a distal location, and this wrench can then be transposed to the vehicle center-of-mass using the known offset to the sensor. For example, the moment transforms as:

\begin{equation}
    \vec{M}^\text{H}_\text{FT} = \vec{M}_\text{FT} + \vec{r}^\text{ H}_\text{FT}  \times \vec{F}_\text{FT} \label{eq:mft}
\end{equation}

Where $\vec{M}^\text{H}_\text{FT}$ is the moment as displaced to the helicopter center of mass $H$, $\vec{F}_\text{FT}$ \& $\vec{M}_\text{FT}$ are the force and moments measured at the origin of the force-torque sensor, and $\vec{r}$ is the offset between. Force-torque sensing approaches are commonly used in wind-tunnel testing for system identification of aircraft, allowing the development of dynamics models, validating first-principle models, and carrying out control design about the relevant origin. The Ingenuity Mars Helicopter is no exception; force-torque measurements of the prototype vehicle in the low-density chamber at JPL allowed for the successful verification of the rotor control laws \cite{grip2020modeling} and are, accordingly, a high-heritage method for determining the aerodynamic state of Mars rotorcraft.  

\begin{figure}[h!]
    \centering
    \includegraphics[width=0.5\textwidth]{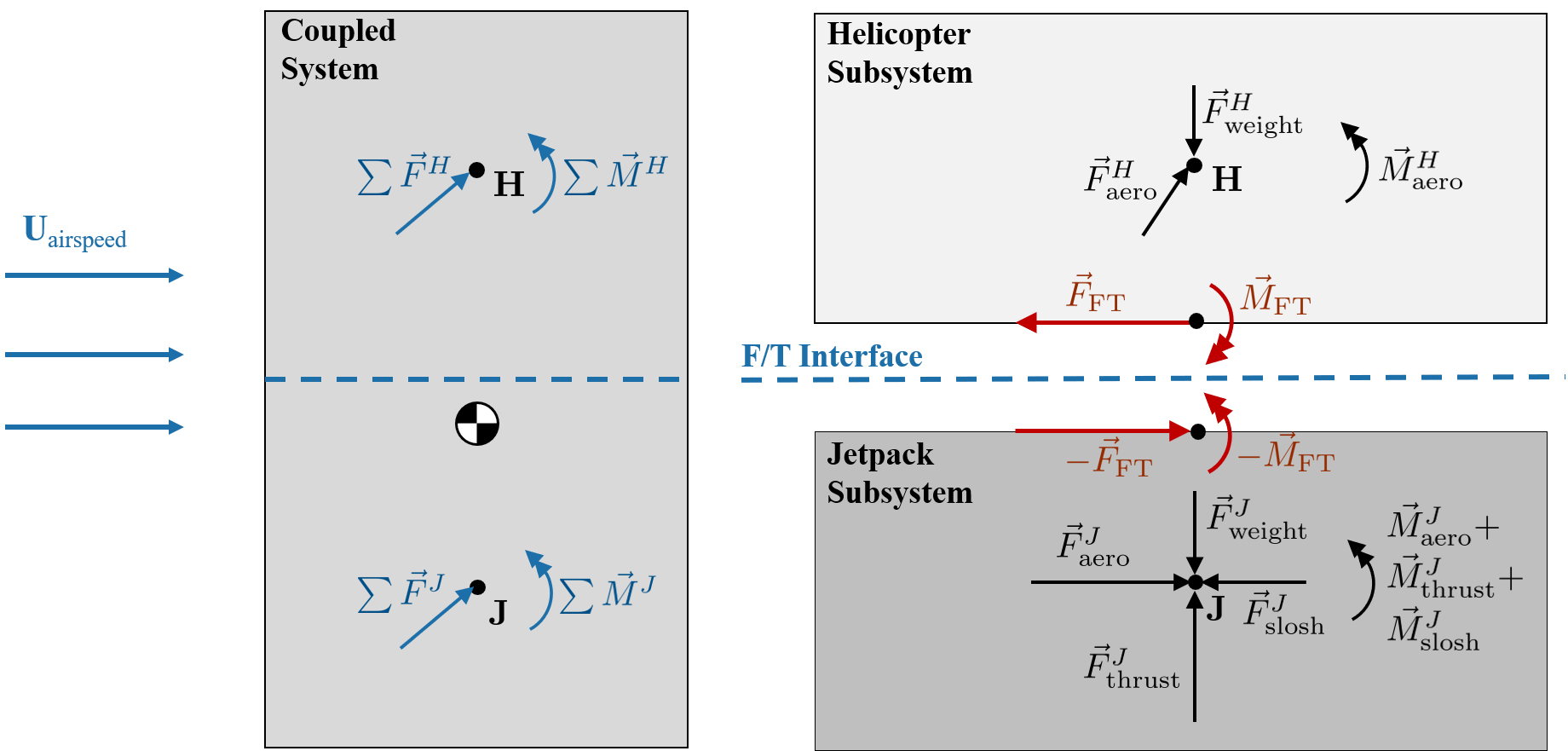}
    \caption{\bf Free body diagrams of helicopter (top-right), jetpack (bottom-right), and composite system (left).}
    \label{fig:fbd}
\end{figure}

\begin{figure}[h!]
    \centering
    \includegraphics[width=0.45\textwidth]{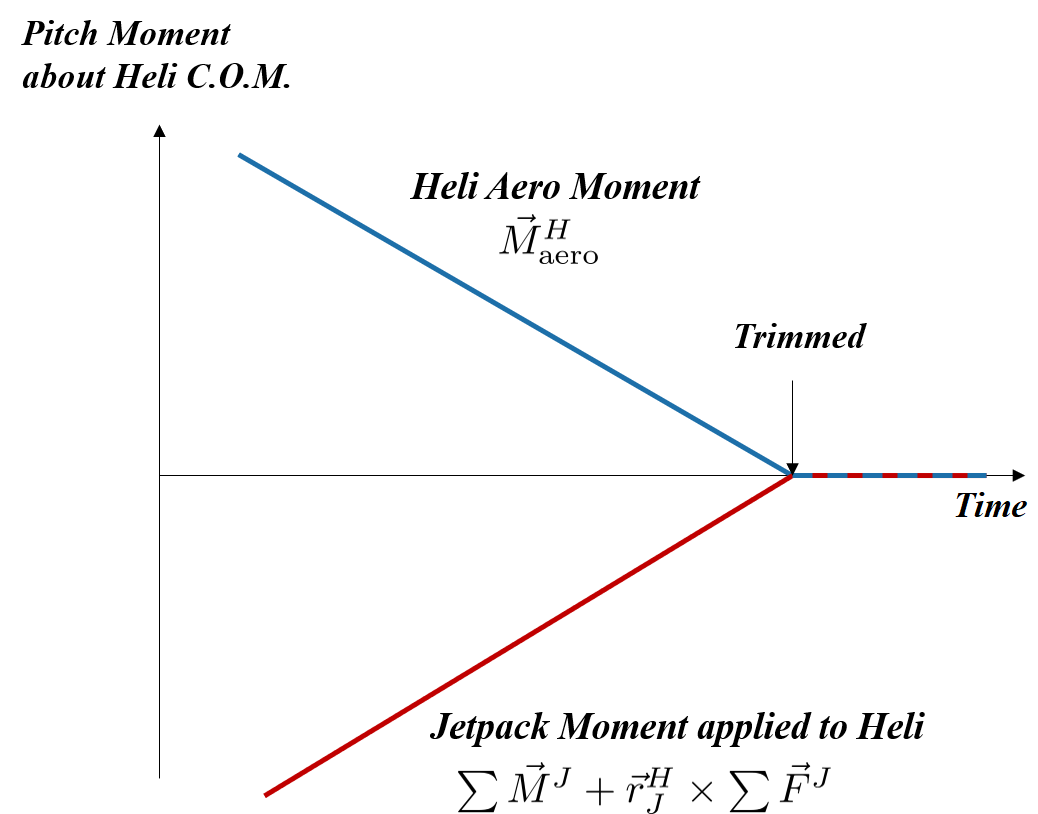}
    \caption{\bf Conceptual operation of trim maneuver: Jetpack thrusters maintain a total moment of zero to keep level attitude, and the trimming procedure slowly reduces the aerodynamic moment until the helicopter is ready for takeoff.}
    \label{fig:trimplot}
\end{figure}

Free body diagrams of the helicopter, the jetpack, and the composite system are illustrated in Figure \ref{fig:fbd}, and force/moment summations enumerated below. The jetpack controller acts on full system moments (Eqs. \ref{eq:fs} \& \ref{eq:ms}), as it closes the loop based on terrain-relative navigation for altitude and attitude control on the entire rigid body about its center of mass. 
\begin{equation}
    \sum \vec{F}^\text{\centerofmass} = \sum \vec{F}^\text{H} + \sum \vec{F}^\text{J} \label{eq:fs}
\end{equation}
\begin{multline}
    \sum \vec{M}^\text{\centerofmass} = \sum \vec{M}^\text{J} + \sum \vec{M}^\text{H} +  \vec{r}^\text{ \centerofmass}_{J}  \times \sum \vec{F}^\text{J} + \\  \vec{r}^\text{ \centerofmass}_{H}  \times \sum \vec{F}^\text{H} \label{eq:ms}
\end{multline}
Where forces and moments are summed over the helicopter, $H$, and jetpack, $J$, at their centers of masses, and transformed by their offsets to the total system center of mass~\centerofmass. Accordingly, as the rotors adjust the helicopter forces and moments ($\vec{F}^\text{H}$, $\vec{M}^\text{H}$) during the trim maneuver, the jetpack thrusters have to compensate for the changing aerodynamic forces to maintain altitude and level attitude (Figure \ref{fig:trimplot}).

On the helicopter, the total forces and moments about its center-of-mass are due to the total aerodynamic wrench (lift, drag, and moment) on the fuselage and rotors, the weight, and the reaction force from the force-torque meter:
\begin{equation}
    \sum \vec{F}^\text{H} = \vec{F}^{H}_\text{FT} + \vec{F}^{H}_\text{aero} + \vec{F}^{H}_\text{weight} \label{eq:fh}
\end{equation}
\begin{equation}
    \sum \vec{M}^\text{H} = \vec{M}^{H}_\text{FT} + \vec{M}^{H}_\text{aero}
    \label{eq:mh}
\end{equation}
Similarly, the jetpack has forces and moments about its center-of-mass due to its aerodynamics, weight, thrusters, and tank slosh.
\begin{equation}
    \sum \vec{F}^\text{J} = \vec{F}^{J}_\text{FT} + \vec{F}^{J}_\text{aero} + \vec{F}^{J}_\text{thrust} + \vec{F}^{J}_\text{slosh} + \vec{F}^{J}_\text{weight}
\end{equation}
\begin{equation}
    \sum \vec{M}^\text{J} = \vec{M}^{J}_\text{FT} + \vec{M}^{J}_\text{aero} + \vec{M}^{J}_\text{thrust} + \vec{M}^{J}_\text{slosh}
\end{equation}
As the force-torque sensor forces and moments are internal and accordingly cancel, they do not affect the system altitude and attitude dynamics (Eqs. \ref{eq:fs} \& \ref{eq:ms}) but instead solely hold each half of the system in place. Accordingly, given adequate acceleration knowledge, Eqs. \ref{eq:fh} and \ref{eq:mh} can be inverted to recover the helicopter aerodynamic wrench $(\vec{F}^{H}_\text{aero}, \vec{M}^{H}_\text{aero})$ from the force-torque meter $(\vec{F}^{H}_\text{FT}, \vec{M}^{H}_\text{FT})$.

In the following, ideal attitude and altitude closed-control loop are assumed for the jetpack. Prior to separation, (step 3 in Figure~\ref{fig:FTconops}), the external moments $\vec{M}^{J}_\text{slosh}$ and $\vec{M}^{J}_\text{aero}$ are canceled by $\vec{M}^{J}_\text{thrust}$. They do not affect the capability to trim the rotors. This is not the case for the internal moments, such as resonant vibrations stimulated by the thruster's PWM and rotor excitation. Thruster PWM excitation follows a 10 Hz duty cycle, while the rotor RPM is at $\sim$46 Hz. The first resonance frequency of the MSH rotor is 45 Hz~\cite{johnson2020}. While the flexible mode analysis of the jetpack and MSH fuselage is future work, a requirement can be set that their first resonance be above 10 Hz. \textbf{This will guarantee a decade separation between the frequencies at which the system vibrates and that at which the winds need to be measured ($\sim$1 Hz), and enable rotor trimming via force-torque sensing}.

Once the thrusters shut down for the brief instant before takeoff, $\vec{F}^{J}_\text{aero}$, $\vec{F}^{J}_\text{weight}$, $\vec{F}^{J}_\text{slosh}$, $\vec{M}^{J}_\text{slosh}$, and $\vec{M}^{J}_\text{aero}$ do contribute to the rigid-body system dynamics and, accordingly, contribute to angular/linear velocities at takeoff. Quantifying these takeoff parasitics is ongoing work, requiring coupled CFD simulations between the jetpack and helicopter (see Section \ref{sec:flow}), but initial first-principle estimates indicate they are negligible compared to the aerodynamic wrench of the rotor. 

\subsection{Force-Torque Sensor Design}
Preliminary discussions between JPL and ATI Industrial Automation, the supplier of the Mars 2020 sample handling force-torque meter, are ongoing on the feasibility of a force-torque meter customized to this context. To date, draft requirements have been formulated: sensing range on all six axes, temperature range, worst case loads, max allowable deflection, and stiffness of the mounting interface.

\begin{figure}[h!]
    \centering
    \includegraphics[width=0.4\textwidth]{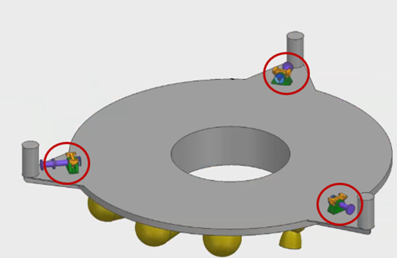}
    \caption{\bf Placement of three discrete force-torque meters below detach mechanisms}
    \label{fig:fts}
\end{figure}

The proposed location for this force-torque meter is illustrated in Figure \ref{fig:fts}, where a sensor would be placed below each of the three mountpoints of the helicopter. This placement ensures that all forces and moments between the two subsystems are captured through the interface, with the exception of necessary cabling that crosses the gap and must be calibrated out. Such alternate load paths always exist, but early architectural discussions on force-torque meter placement allow these to be minimized and mitigated sooner, reducing the complexity of later calibration work \cite{schaler2021two}.

Initial conceptual force-torque sensor designs indicate that the problem appears feasible, with sensing frequencies in the kilohertz and mass cost of a few kilograms. Such high-frequency data allows filtering out of the jetpack thruster pulse-width-modulated input disturbances (10 Hz and harmonics), and the structural vibration modes of the jetpack discussed earlier, while the meeting $>$1-Hz wind sensing requirement set in Section~\ref{sec:wind}. Forces and moments are expected to be resolved to 2 N vertical, 0.3 N lateral, and 0.2 Nm on all moment axes. Based on aerodynamic analysis of the MSH rotorcraft in the CAMRADII software, this translates into a wind sensing resolution of $<1$m/s.

Experiments with subscale models of this architecture are planned over the next year, with rotorcraft taking off from vibrating tables under lateral winds to validate the force-torque sensor concept-of-operations.

\subsection{Helicopter Control Analysis}


\begin{figure*}
\centering
  \begin{minipage}{7in}
    \includegraphics[width=2.2in]{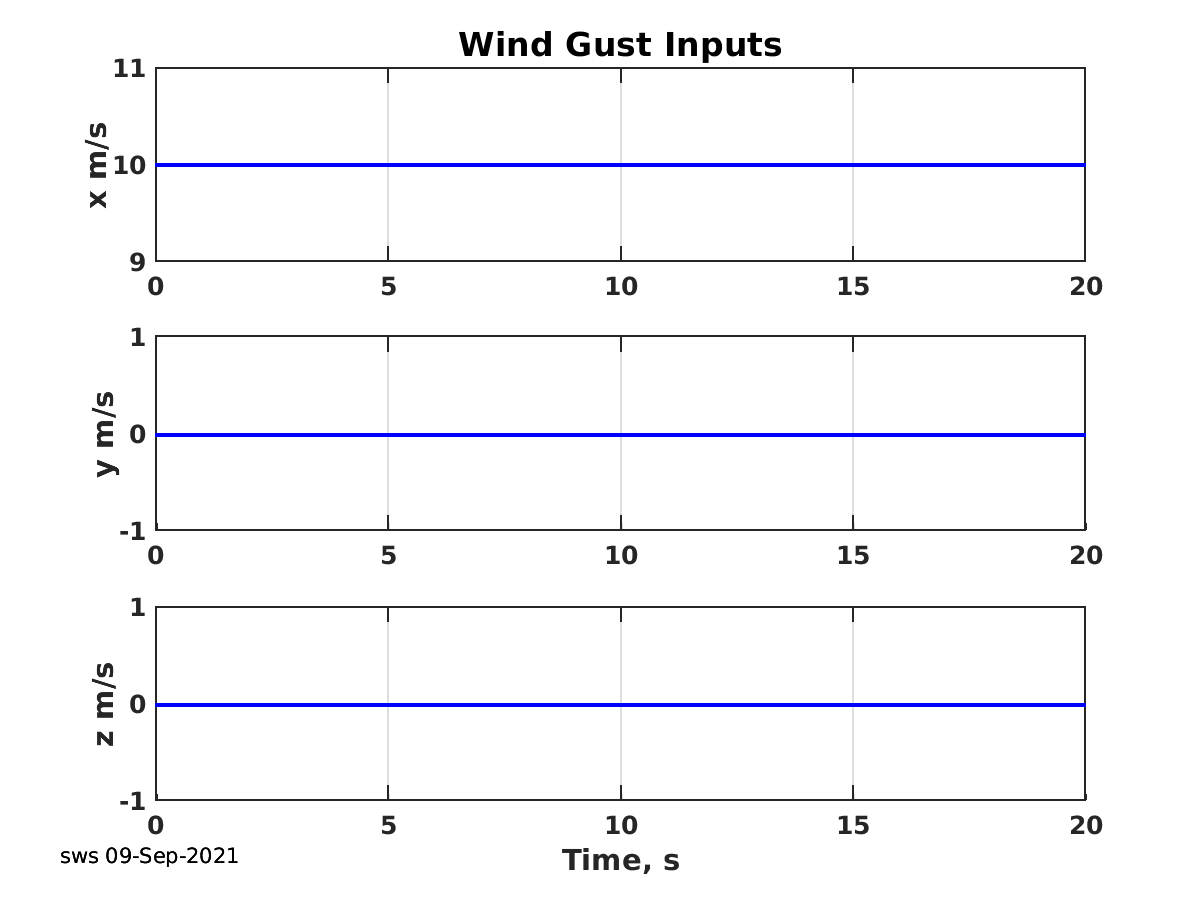}
    \includegraphics[width=2.2in]{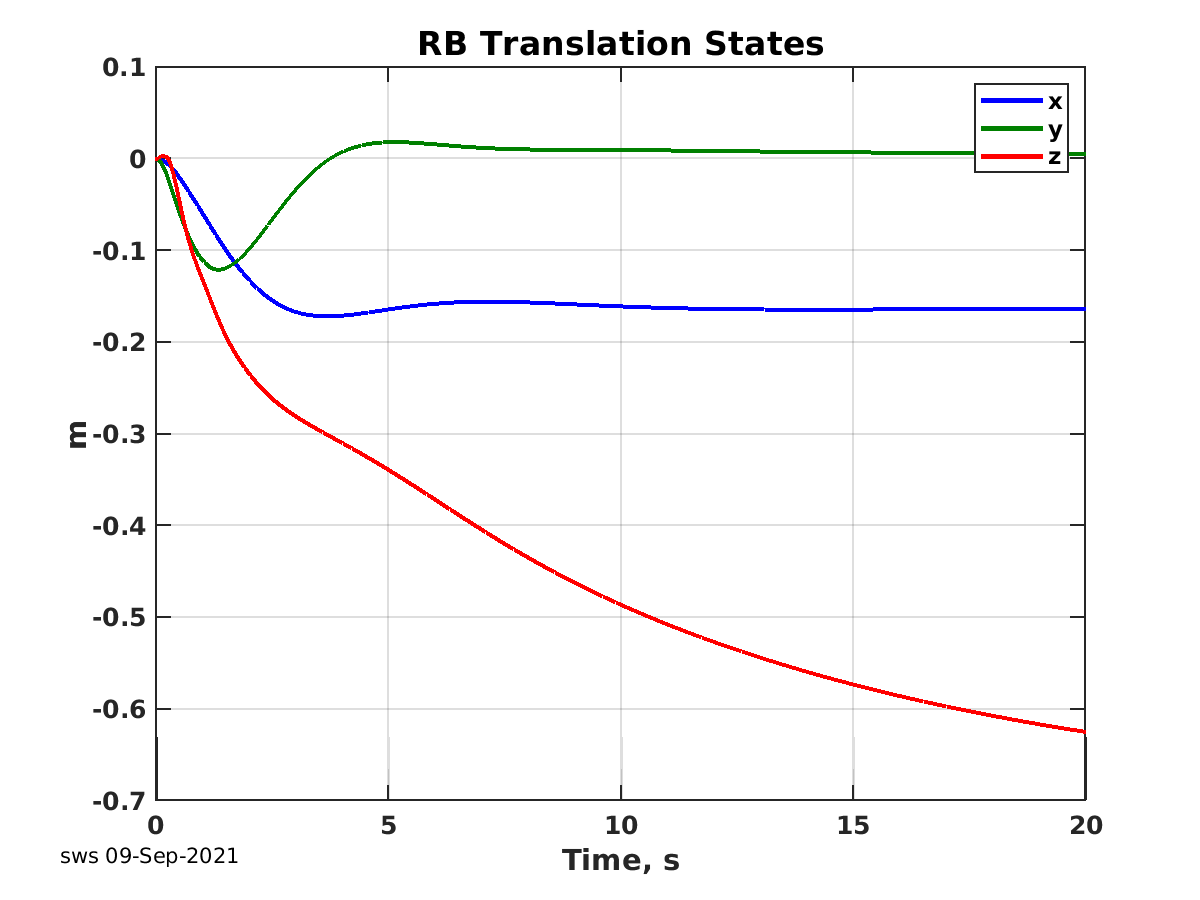}
    \includegraphics[width=2.2in]{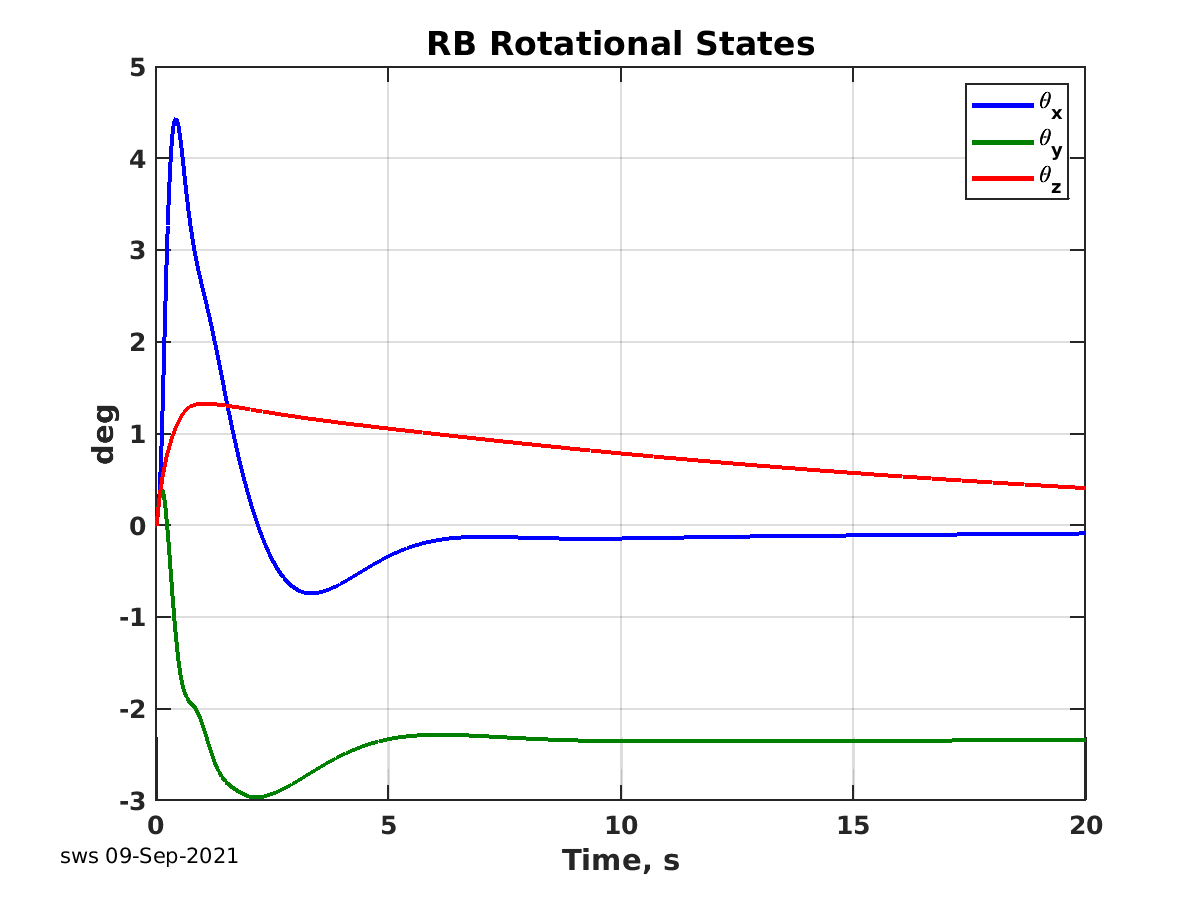}\\
    \includegraphics[width=2.2in]{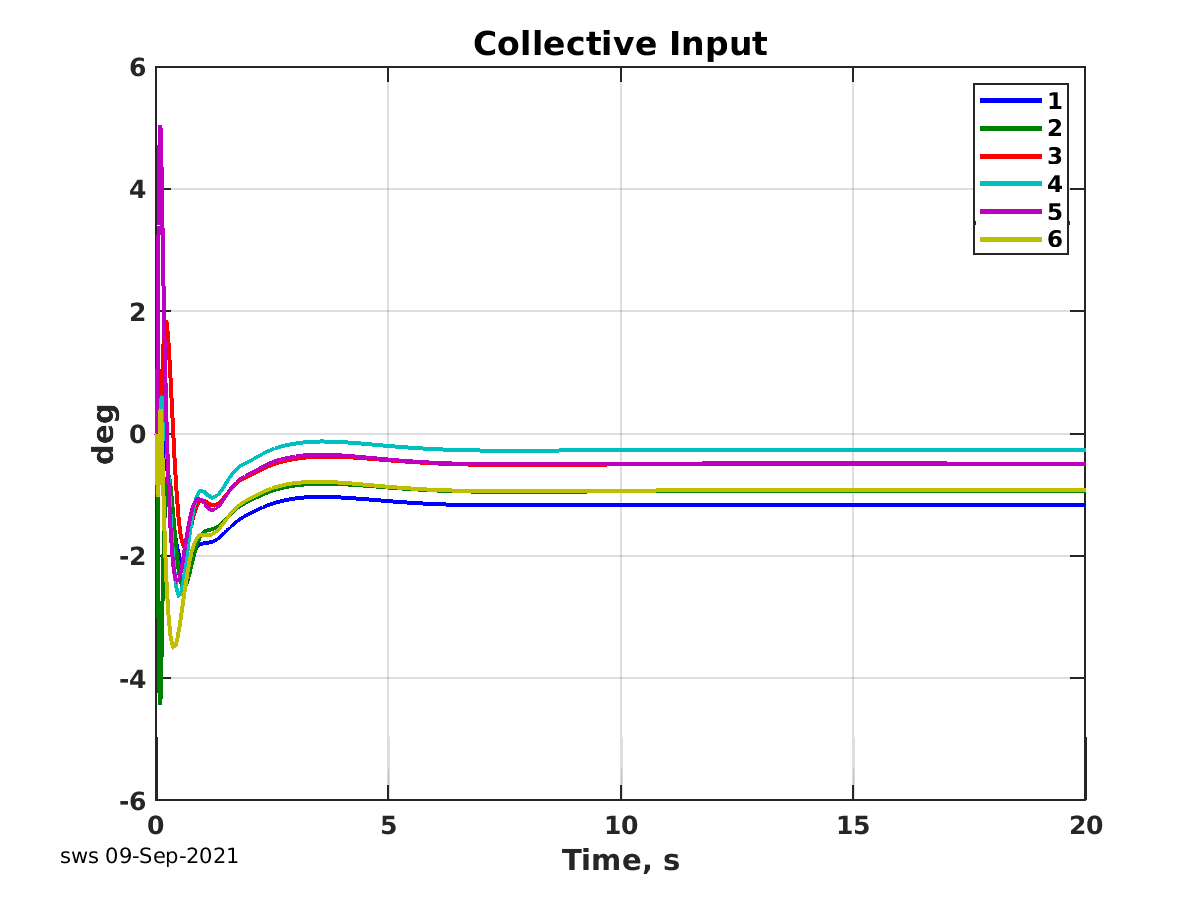}
    \includegraphics[width=2.2in]{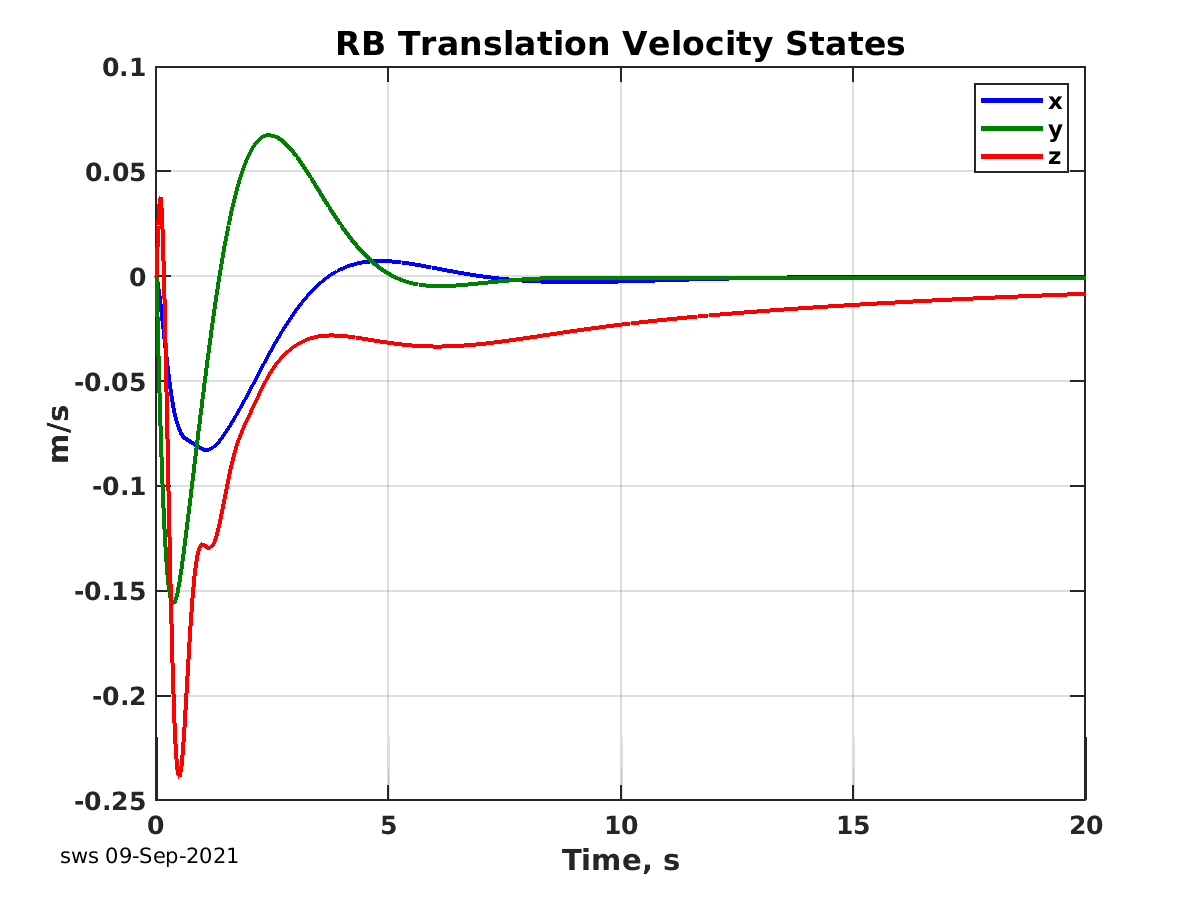}
    \includegraphics[width=2.2in]{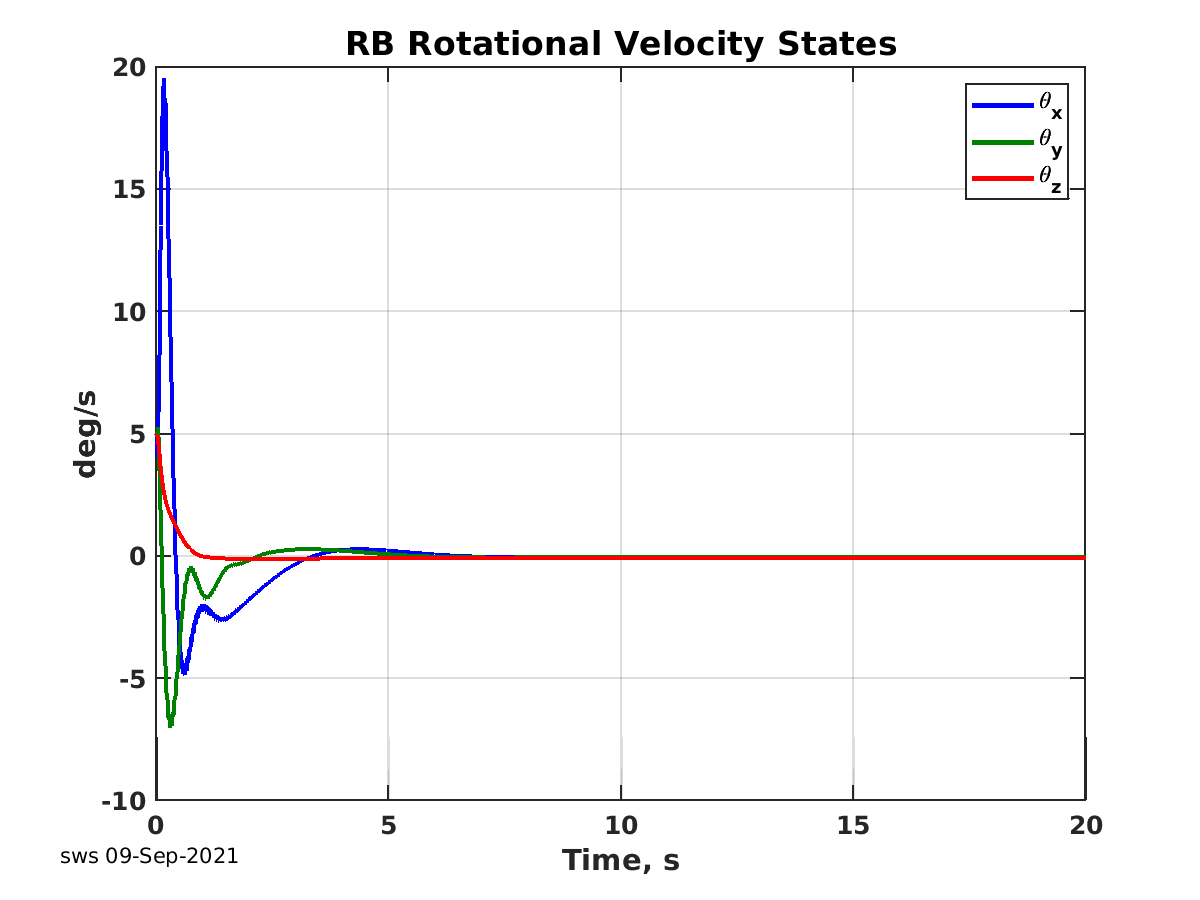}
  \end{minipage}
\caption{\bf{Mars Science Helicopter takeoff simulation example.}}
\label{fig:msh-simA}
\end{figure*}

A simple 6-Degrees-of-Freedom (DoF) controller was developed for MSH
in order to assess controllability in presence of rotor trimming errors. 
A simple simulation of helicopter takeoff is shown in
Figure~\ref{fig:msh-simA}. The simulation uses linearized dynamics for
MSH at hover, i.e., 0 m/s velocity. In this case, the simulation starts with a
forward velocity of 10 m/s, with an initial jump in wind speed of 10
m/s. Initial angular velocity is 5 \deg/s in roll, pitch, and yaw.
The controller quickly stabilizes the initial rates, with a steady 
additional pitch of +2\deg from the wind, and some translational
drift.  
Collective angles stay $< 11\deg$, a goal from staying in the
linear range of the dynamics. Later work will refine the maximum
allowable collective angles.

A simple Monte Carlo simulation was carried out in order to develop a
preliminary understanding of a reasonable range of initial conditions.
Note that here all 6 DoFs were controlled, though we would
expect to only control attitude initially. 
Initial condition (IC) limits were chosen  in order to keep
the resulting  collective angles $< 11\deg$\ over the course of the
simulation. Initial conditions were picked randomly from an
ellipsoidal region with specified limits. The initial wind gust was
allowed to be in a cube with components $\norm{w} < 10$ m/s.
The initial position command was set to the actual position (so, no
initial translational offsets).
2000 Monte Carlo trials were performed for the hover case. See Table \ref{tab:hovermc} for the state 
behavior. The resulting collective
angles stay $< 10.6 \deg$, and 95\% of the time they are $<
9.0\deg$. The state IC limits were chosen simply by trial and error to keep the
collective angles in bounds. An optimization procedure could be used
to refine the numbers and enlarge or shrink dimensions of the
ellipsoid. Here, quite large initial attitude rates were chosen. 
The IC maximum achieved by the random
choice during the trials is shown. The resulting maximum values
achieved over time for each of the trials is also shown.
There is some translation and altitude drift as a result, which would
be removed more slowly by an outer commander loop. Attitude
perturbations stay less than 3\deg. 
\begin{table}
  \caption{\bf Hover Case Monte Carlo Trial Results}
  \label{tab:hovermc}\tiny
  \begin{tabular}{lllllllll}
    \hline
    &\multicolumn{4}{c}{Position}&\multicolumn{4}{c}{Rate}\\
    &\multicolumn{2}{c}{Translation}&\multicolumn{2}{c}{Rotation}
    &\multicolumn{2}{c}{Translation}&\multicolumn{2}{c}{Rotation}\\
&xy&z&Rxy&Rz&Vxy&Vz&VRxy&VRz\\
&m&m&deg&deg&m/s&m/s&deg/s&deg/s\\\hline
IC\ limit&0&0&1.1&0.5&0.08&0.6&25&15\rule{0pt}{2.5ex} \\
IC\ max&0&0&0.855&0.38&0.0623&0.464&21.3&10.7\\
Max(t)&0.158&2.23&2.71&2.75&0.0821&0.464&27&10.7\\
\hline
\end{tabular}
\end{table}

In this section, we demonstrated MSH controls for take-off robust to 5-deg/s jetpack rate residual,
and 10 m/s airspeed trim residual at hover. We proposed a preliminary force-torque sensor
design based on first principles, which is accurate up to 1 m/s equivalent airspeed torque at hover. 
Given the jetpack control accuracy results presented in Section~\ref{sec:GNC} ($<$ 0.2 deg/s 3$\sigma$), these
preliminary results are promising indications of the feasibility of using force-torque sensing for the
transition of controls from the jetpack to MSH.


\section{Conclusion}

A new EDL system called MAHD was presented, to deliver a Mars Science Helicopter vehicle in mid-air using a jetpack. Compared to past Mars EDL approaches, we showed that MAHD results in improved rotorcraft performance (especially +150\% increased science payload mass), a simpler architecture likely to reduce cost, access to more hazardous and higher-elevation terrains. We proposed a preliminary design for the full system, as well as a detailed concept of operations. We established preliminary evidence of feasibility through a combination of analytical and numerical analyses. Highlights include:
\begin{itemize}
    \item a mechanical interface design which can fit both MSH and the jetpack within the heritage 2.65-m aeroshell;
    \item an analysis of the wind dynamics at a representative landing site, which shows winds are mostly horizontal and up to $\sim$30 m/s;
    \item preliminary CFD simulation models of the interactions between the rotors, jets and side, showing promising results regarding the controllability of MSH in the jet-entrained flow;
    \item a GNC architecture for the jetpack, and a Monte Carlo analysis of the closed-loop performance in simulation;
    \item an architecture for the transition of controls from the jetpack to MSH, using a force-torque meter to handle the winds at take-off.
\end{itemize}

On-going research is focused on the experimental characterization of the critical events for MAHD. Specifically, aerodynamically-scaled experiments will be conducted in a 1-atm wind tunnel to model the interactions between the jet-rotor wake and the side winds, as well as the force-torque sensing and take-off dynamics. These updated models will be incorporated in end-to-end high-fidelity simulations to bring further evidence of the feasibility of MAHD on Mars.

\acknowledgments

The authors would like to thank the technical reviewers who provided valuable feedback at different stages of the design: Jeffery Hall, Miguel San Martin, Joe Melko, Kim Aaron, Erik Bailey, Chris Porter, Kim Aaron, Mike Paul Hughes, Art Casillas, Carl Guernsey, Kalind Carpenter, Matthew Heverly, Joshua Ravich and Ian Clark.

Part of the research described in this paper was carried out at the Jet Propulsion Laboratory, California Institute of Technology, under a contract with the National Aeronautics and Space Administration (80NM0018D0004).

\textcopyright~2021. All rights reserved.

\bibliographystyle{IEEEtran}
\bibliography{IEEEabrv,root}

\thebiography
\begin{biographywithpic}
{Jeff Delaune}{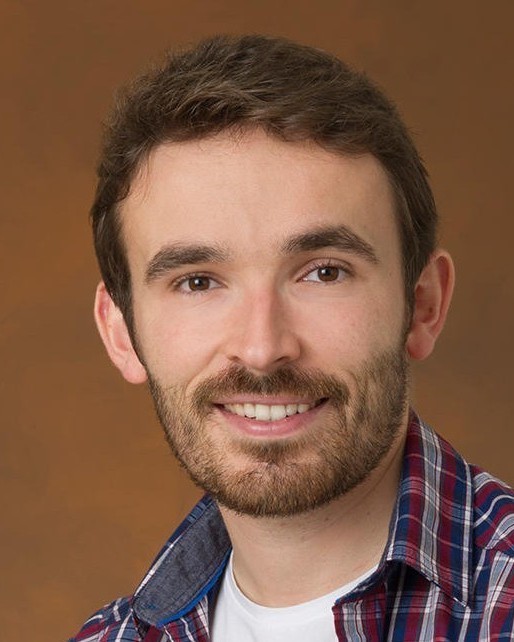}
is a robotics technologist in the Aerial Mobility Group at the Jet Propulsion Laboratory. His interests include analysis, design, implementation, and testing of new technology for planetary exploration, with a particular focus on autonomous aerial systems and vision-based navigation for now. Jeff is part of the Guidance, Navigation and Control team for NASA’s Ingenuity Mars Helicopter. He received his Ph.D. in Robotics from Institut Sup\'erieur de l’A\'eronautique et de l’Espace (ISAE, France) in 2013, after a M.S. in Astronautics and Space Engineering from Cranfield University (United Kingdom), and a B.S./M.S. in Engineering from \'Ecole Centrale de Nantes (France).
\end{biographywithpic}

\begin{biographywithpic}
{Jacob Izraelevitz}{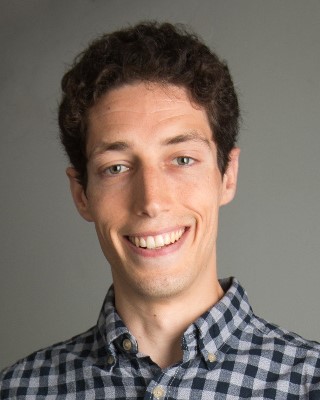} is a robotics technologist in the Extreme Environments Robotics Group at the Jet Propulsion Laboratory. Jacob received a B.S. in Mechanical Engineering from Olin College of Engineering in 2011, followed by a M.S. and Ph.D. in Mechanical Engineering from the Massachusetts Institute of Technology (2013 and 2017 respectively) in the aerodynamics of flapping wings. His career at JPL has primarily sat at the interface of controls and fluid mechanics, covering both powered and buoyant aerial platforms. Jacob was the project manager and principal aerodynamicist for JPL's contribution to the DARPA Mobile Force Protection Program, which demonstrated the aerial deployment of a rotorcraft from a moving vehicle.
\end{biographywithpic}

\begin{biographywithpic}
{Sam Sirlin}{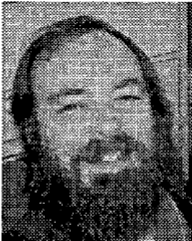}
has a Ph.D. in Mechanical Engineering from Columbia University (1983). He is a Principal Engineer in Guidance and Control at JPL. His background and expertise cover the areas of dynamics, simulation, controls and estimation.   During his time at JPL he has worked on a number of research and flight projects including Galileo, Mars Pathfinder, Deep Space 1, Space Interferometer Mission, Mars Exploration Rover, Spitzer Telescope, Mars Science Laboratory, and Europa Clipper.
\end{biographywithpic}

\begin{biographywithpic}
{David Sternberg}{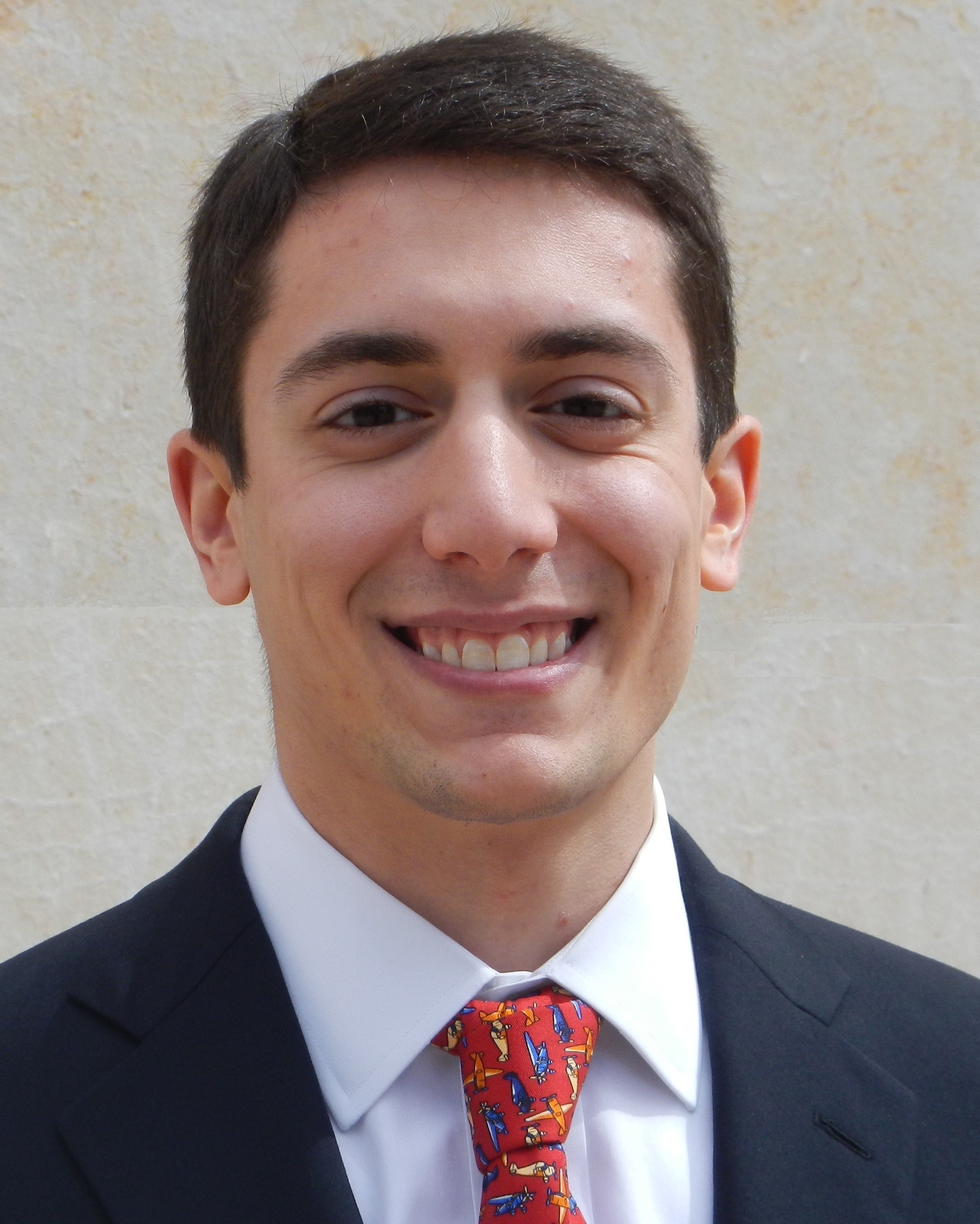}
is a guidance and control systems engineer at the NASA Jet Propulsion Laboratory, having earned his S.B., S.M., and Sc.D. degrees in the MIT Department of Aeronautics and Astronautics. He is currently working on the development, testing, and operation of satellite attitude determination and control hardware for small satellites, having served as the lead attitude control system operator for the MarCO satellites, and is a guidance and control analyst for the Psyche mission. 
\end{biographywithpic}

\begin{biographywithpic}
{Dr. Lou Giersch}{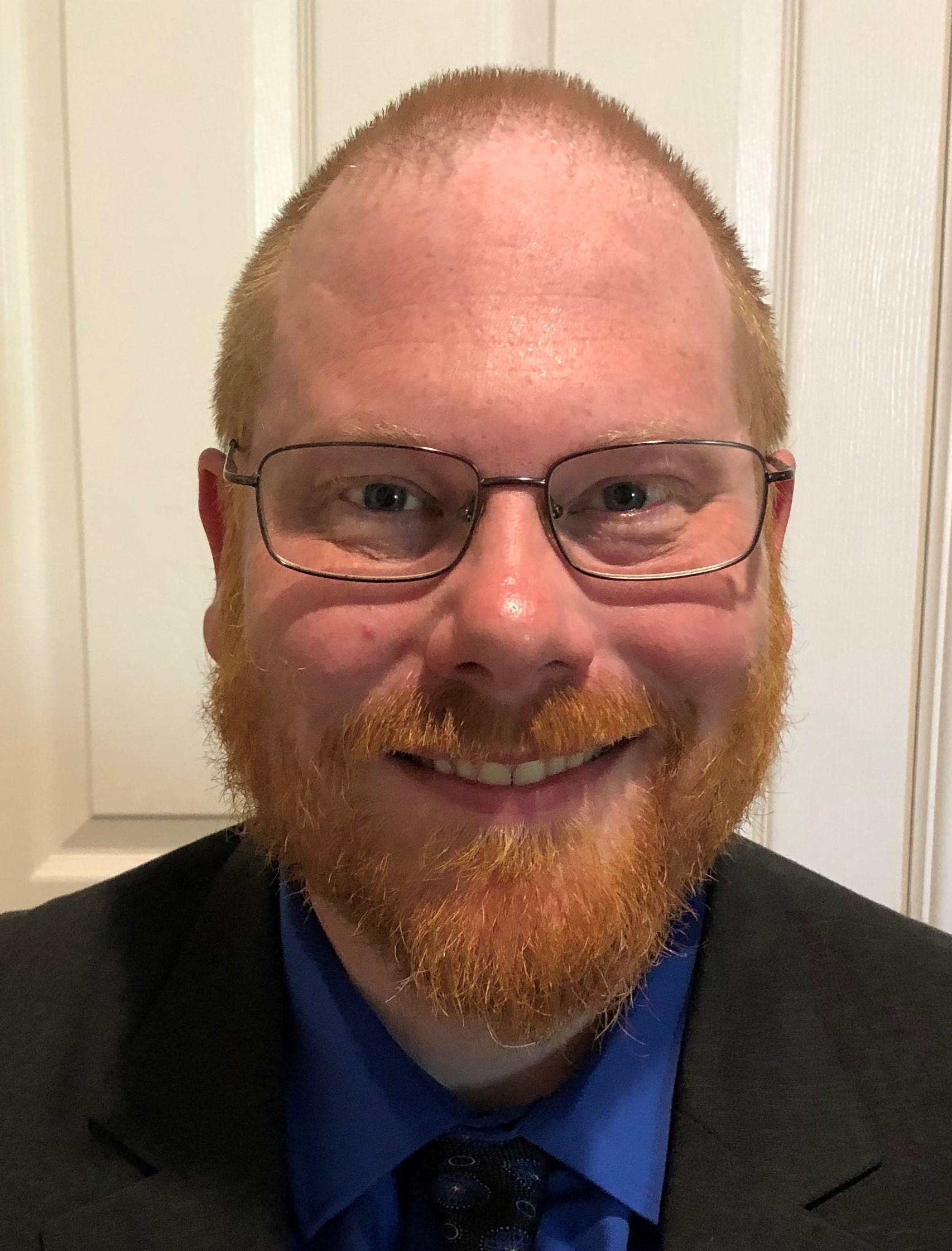}
received his PhD in Aeronautics and Astronautics from the University of Washington in Seattle, Washington, USA, in 2005, and has been working at the NASA Jet Propulsion Lab since 2007. He is currently the primary investigator for the SHIELD Mars lander concept, and lead mechanical engineer for MAHD. 
\end{biographywithpic}

\begin{biographywithpic}
{L. Phillipe Tosi}{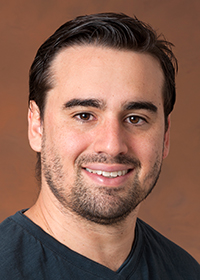} received his Ph.D. in mechanical engineering from the California Institute of Technology with a focus on  fluid-structure interaction, numerical methods, reduced-order models, and experimental methods. He has a B. S. and M. Eng. in mechanical and aerospace engineering from Cornell University, and a M. S. in mechanical engineering also from Caltech. He works as a robotics technologist in the extreme environment robotics group at NASA’s Jet Propulsion Laboratory, where he manages and contributes to robotics research projects including a long lived Venus lander, a Mars deep drill, the Europa lander, and Mars flight technology.  
\end{biographywithpic}

\begin{biographywithpic}
{Evgeniy Sklyanskiy}{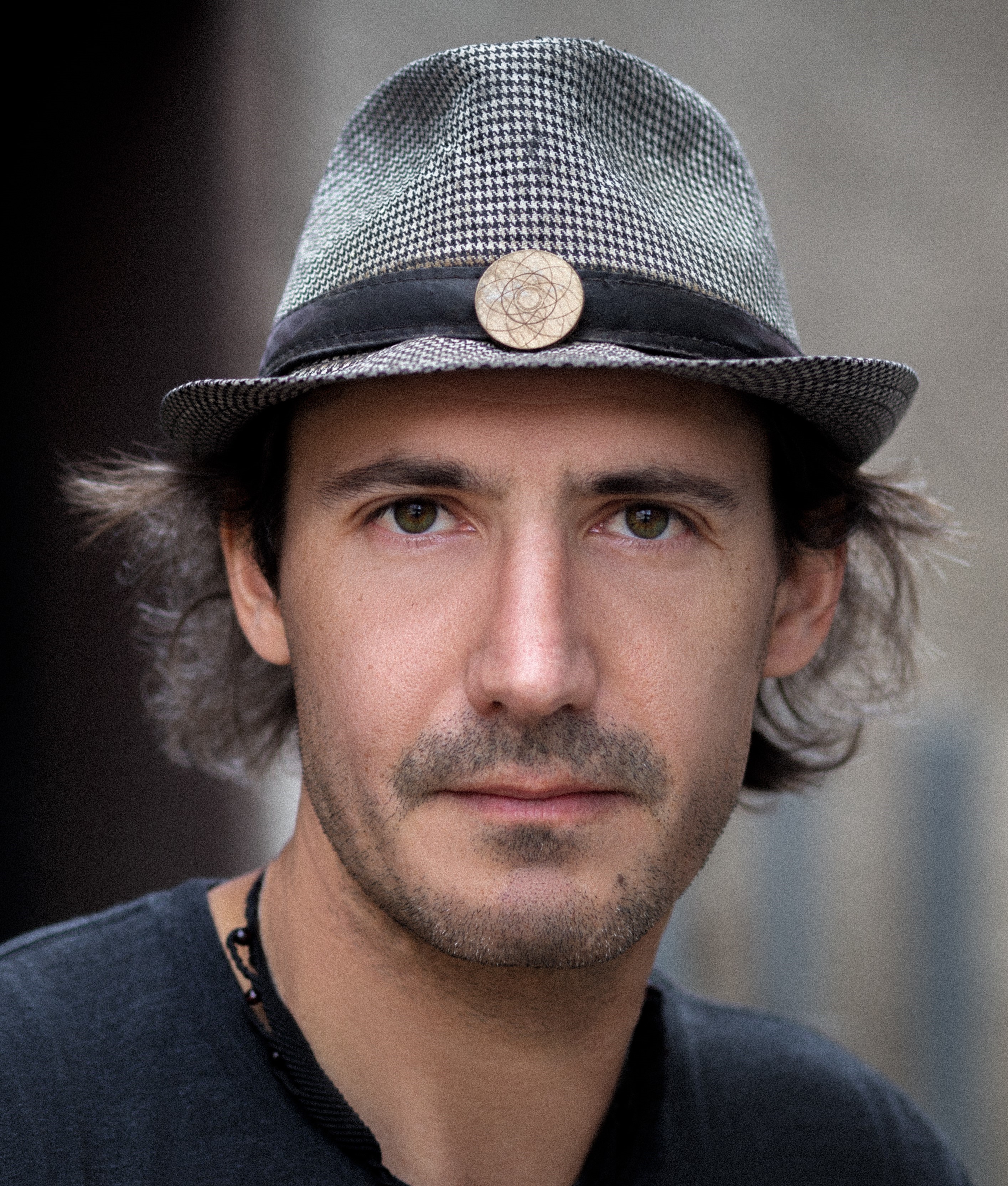} is a guidance and control engineer in Entry Descent and Landing (EDL) Group at the Jet Propulsion Laboratory. Evgeniy received a B.S. in Aerospace Engineering from University of Illinois at Urbana-Champaign (UIUC) in 2001, followed by a dual M.S. degree in Mechanical Engineering and Applied Mathematics from UIUC in 2004. The same year Evgeniy joined JPL in outerplanetary trajectory analysis and design group as part of the Mission Design and Navigation (MDNAV) section. As an EDL/MDNAV trajectory analyst, he supported many flight missions including InSight, Hayabusa I and II.
\end{biographywithpic}

\begin{biographywithpic}
{Larry Young}{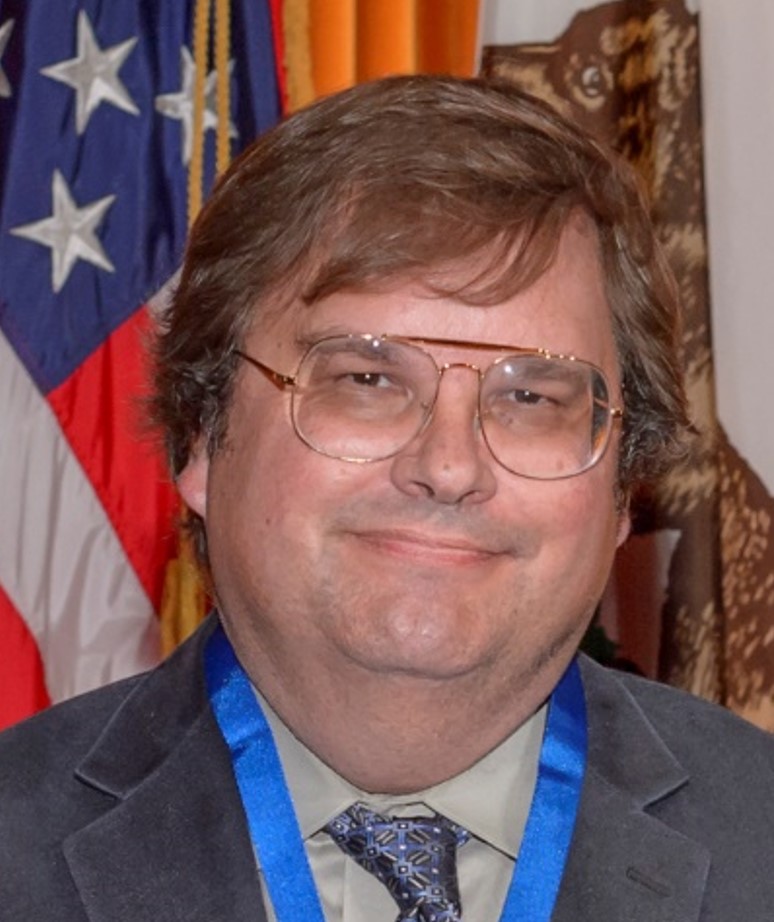} received his B.S. and M.S. in Mechanical Engineering from Washington State University and now works in the Ames Aeromechanics Office. Mr. Young performs research on advanced aerial vehicle and aerospace system conceptual design. Among his current and past projects are studies into fundamental vortical flow physics, planetary aerial vehicles, rotary-wing vehicles for disaster relief and emergency response missions, and advanced tilt-rotor aircraft design.  Mr. Young was an early researcher into Mars rotorcraft and other vertical takeoff vehicles for planetary exploration.
\end{biographywithpic}

\begin{biographywithpic}
{Michael Mischna}{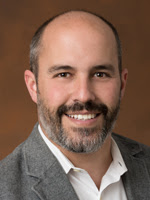} is an atmospheric scientist studying Mars climate and climate evolution.  He received his B.S. in Atmospheric Science from Cornell University, and Ph.D in Geophysics and Space Physics from UCLA. He has worked at the Jet Propulsion Laboratory for 17 years as a research scientist and in support of spacecraft entry, descent and landing for numerous Mars surface missions.  He is presently supervisor of the Planetary and Exoplanetary Atmospheres group at JPL and Principal Scientist of NASA’s Mars Exploration Program Office.
\end{biographywithpic}

\begin{biographywithpic}
{Shannah Withrow-Maser}{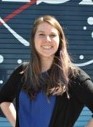} is an aerospace engineer in the Aeromechanics Branch at NASA Ames Research Center with a M.S. in Aerospace Engineering and an emphasis in Engineering Management from Missouri University of Science and Technology. Shannah started in the Aeromechanics Branch as a student researcher in 2014 and joined the branch full-time in 2018. Shannah is member of the vehicle design and flight dynamics groups and is also the Mars Science Helicopter Vehicle Systems Lead. 
\end{biographywithpic}

\begin{biographywithpic}
{Dr. Juergen Mueller}{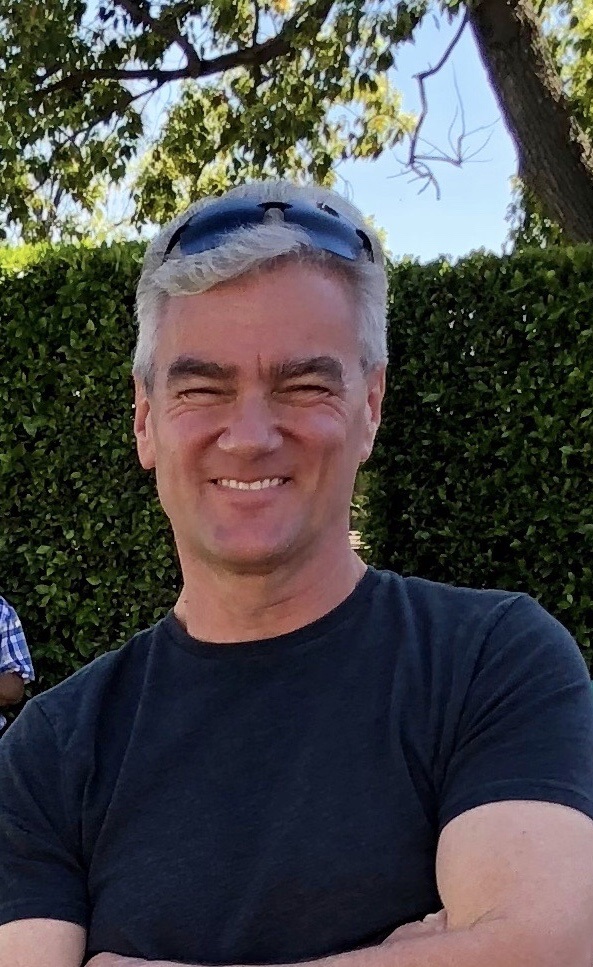} has 35 years of experience in space propulsion. He received a Diploma in Physics at the University of Giessen, Germany in 1987 and a Ph.D. in Aerospace Engineering from Penn State University in 1993 under a Fulbright scholarship. Since joining JPL in 1991, Dr. Mueller has worked in a large variety of fields, including microwave thrusters, ion engines, Hall thrusters, electrospray thrusters, fluid systems components, chemical propulsion, and propulsion support on JPL’s Team X mission study team. Beginning in 1995, he established JPL in the emerging field of micropropulsion, and its application to microspacecraft. He led JPL’s micro-inspector spacecraft effort as principal investigator for the NASA Engineering Systems Mission Directorate (ESMD) beginning in 2004. More recently, he has supported various lunar \& outer planetary lander mission studies as the propulsion lead, studies of ice probes to reach oceans of outer planet icy worlds, and advanced mission studies, including a feasibility study of missions into near interstellar space. He is currently supporting the propulsion system design for the Mars Sample Return mission/Sample Return Lander. Dr. Mueller has published in excess of 40 technical, journal papers, and book chapters in these areas. He was awarded the 2001 JPL Lew Allen Award of Excellence for his work in micropropulsion, has received JPL Mariner \& Discovery awards, and several JPL team awards.
\end{biographywithpic}

\begin{biographywithpic}
{Josh Bowman}{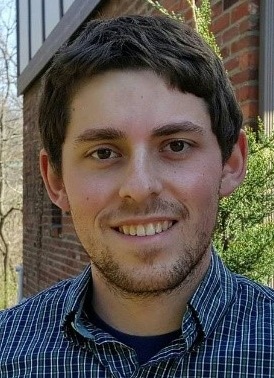} is a mechanical engineer in the Aeromechanics Office at NASA Ames Research Center. He received his M.S. in Nuclear Engineering specializing in Materials at the Pennsylvania State University in 2020 after a B.S. in Mechanical Engineering and a B.S. in Nuclear Engineering in 2016, also from Penn State. He primarily works in mechanical design and analysis, focusing on rotorcraft test stands and vehicle designs. His current work includes numerical solvers for rotor blade structural strength, ROAMX blade optimization, and rotor system identification testing.
\end{biographywithpic}

\begin{biographywithpic}
{Mark S Wallace}{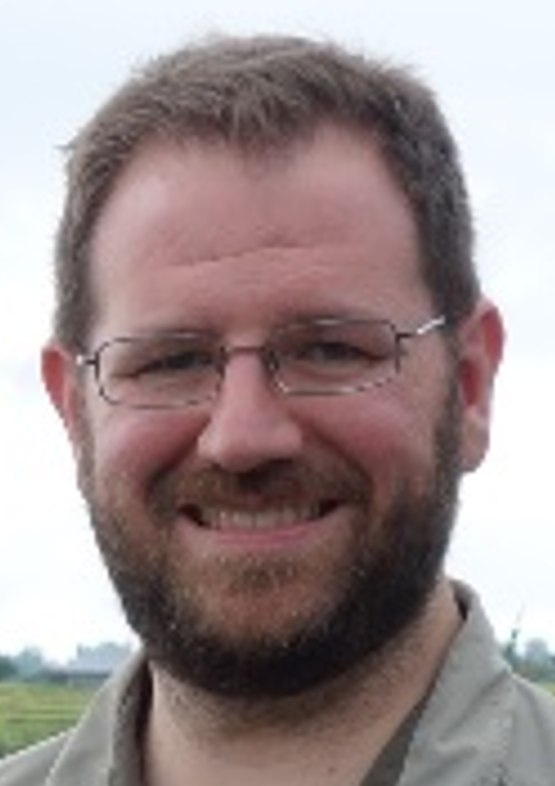} is a Mission Design Engineer at the NASA/CalTech Jet Propulsion Laboratory in Pasadena, California. He received a BS in Aeronautical \& Astronautical Engineering from the University of Illinois at Urbana-Champaign in 2003 and an MS in Aerospace Engineering from the University of Texas at Austin in 2005. A JPLer since 2000, he has contributed to many missions and concepts for robotic and human exploration of the solar system and beyond. He was the Mission Design Chair Lead for the Advanced Projects Development Team, or Team X between 2007 and 2013. He was the main mission designer for the GRAIL extended mission and the lead trajectory analyst for the InSight mission to Mars and is presently the VERITAS Mission Design \& Navigation lead. His research interests include mission design system engineering, multidisciplinary optimization, and concept development.
\end{biographywithpic}

\begin{biographywithpic}
{H\r{a}vard Fj{\ae}r Grip}{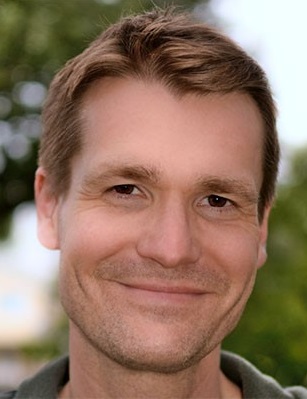} received his MSc and PhD in Engineering Cybernetics from the Norwegian University of Science and Technology in 2006 and 2010, respectively. Prior to joining JPL in 2013, he performed research and development work at the SINTEF Research Group in Trondheim, Norway; Daimler AG in Stuttgart, Germany; and Washington State University in Pullman, Washington. He led the development of Ingenuity's Flight Control system and is currently the helicopter's Chief Pilot.
\end{biographywithpic}

\begin{biographywithpic}
{Larry Matthies}{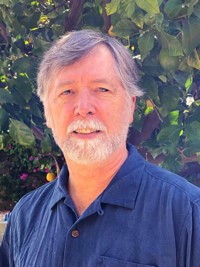} received BS, MMath, and PhD degrees in Computer Science from the University of Regina (1979), University of Waterloo (1981), and Carnegie Mellon University (1989). He has been with JPL for more than 32 years. He supervised the JPL Computer Vision group for 21 years and led development of computer vision algorithms for Mars rovers, landers, and helicopters. He is a Fellow of the IEEE and a member of the editorial boards of Autonomous Robots and the Journal of Field Robotics.
\end{biographywithpic}

\begin{biographywithpic}
{Wayne Johnson}{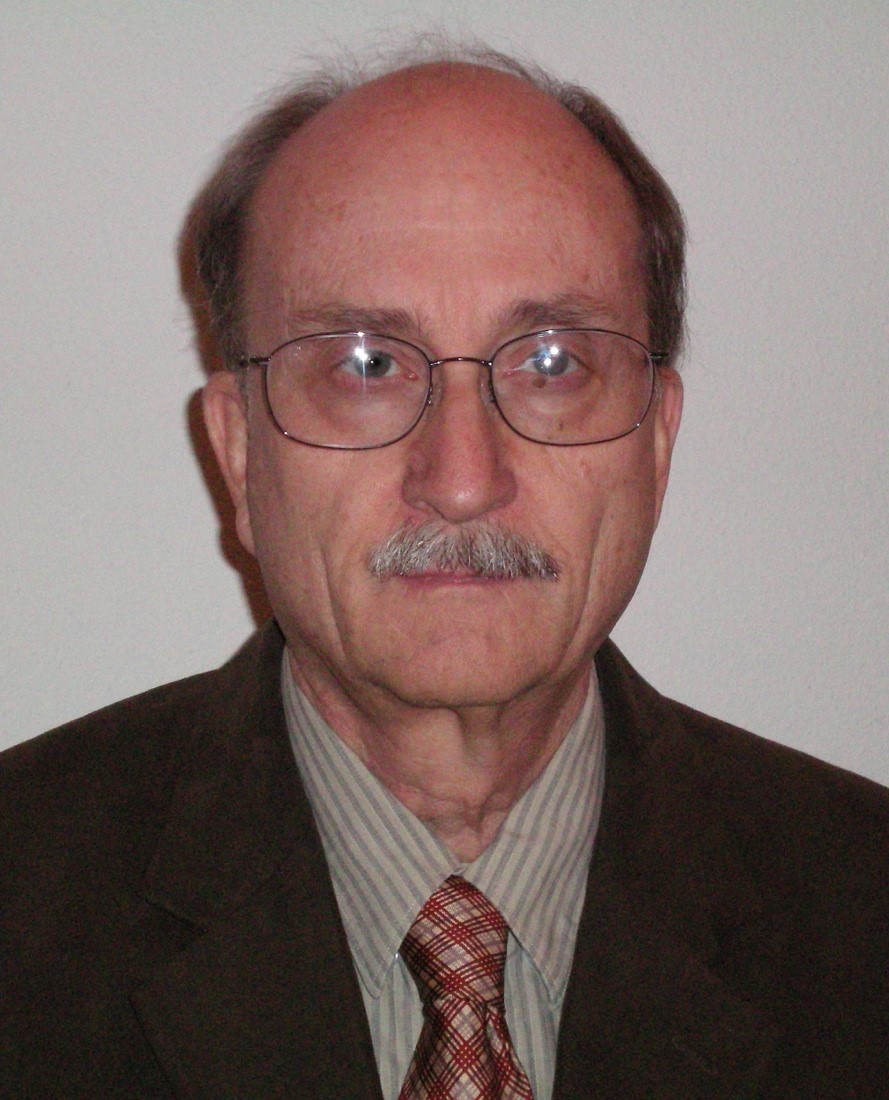} obtained SB, SM, and ScD degrees in aeronautical engineering from the Massachusetts Institute of Technology. He has worked at the U.S. Army Aeromechanics Laboratory and NASA at the 40- by 80-Foot Wind Tunnel of Ames Research Center, then founded Johnson Aeronautics to develop rotorcraft software. He currently works at the Aeromechanics Office of NASA Ames Research Center. Dr. Johnson is author of the comprehensive analysis CAMRADII and the rotorcraft design code NDARC; and the books "Helicopter Theory" and "Rotorcraft Aeromechanics". He is a Fellow of AIAA and AHS, and an Ames Fellow.

\end{biographywithpic}

\begin{biographywithpic}
{Matthew Keennon}{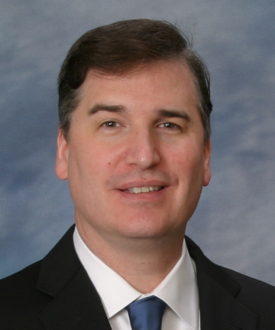} received a B.S. in Physics from University of California, Los Angeles in 1988. He has been with AeroVironment Inc. for more than 25 years.  He was the technical lead for the rotor system development for the JPL Mars Ingenuity helicopter, with particular contributions to the propulsion motor fabrication and electrical wiring.  He maintains the ‘Nano Lab’ in the MacCready Works division of AeroVironment in Simi Valley, providing technical support for many cutting edge UAV projects that require extreme small size or extreme light weight electrical and mechanical systems.  He has been the principal investigator or technical lead on many of the most challenging UAV developments at AeroVironment including the Black Widow squad level micro air vehicle system and the Nano Hummingbird miniature flapping wing, robotic hummingbird.  He specializes in the conceptual design and rapid prototyping of vehicles, propulsion systems, and avionics for novel or unusual UAV systems.  
\end{biographywithpic}

\begin{biographywithpic}
{Benjamin T. Pipenberg}{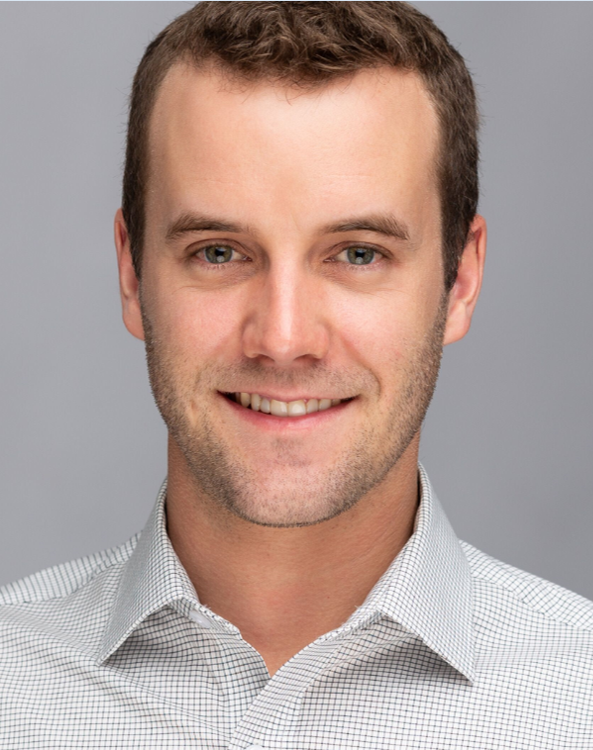} is an aeromechanical engineer at AeroVironment, Inc. in Simi Valley, CA. He received his B.S. in Aerospace Engineering from The Pennsylvania State University in 2011. At AeroVironment, he is a member of the MacCreadyworks division where he has worked on AV’s Nano Air Vehicle programs, the Ingenuity Mars Helicopter, and several Group 3 UAS development programs. His experience includes aircraft design, mechanism design, and composite structure design and fabrication.  
\end{biographywithpic}

\begin{biographywithpic}
{Christopher Lim}{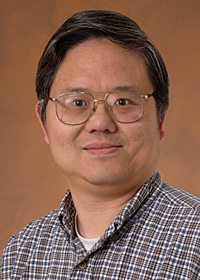} is a Robotics Technologist at JPL. He has been working in the JPL DARTS Lab developing the Darts/Dshell, ROAMS, and EDL/Dsends dynamics simulation platforms for MSL, Phoenix Mars Lander, Mars 2020, Mars Sample Return, Europa Sampling Autonomy and Ocean Worlds projects.  
\end{biographywithpic}

\begin{biographywithpic}
{Marcel Veismann}{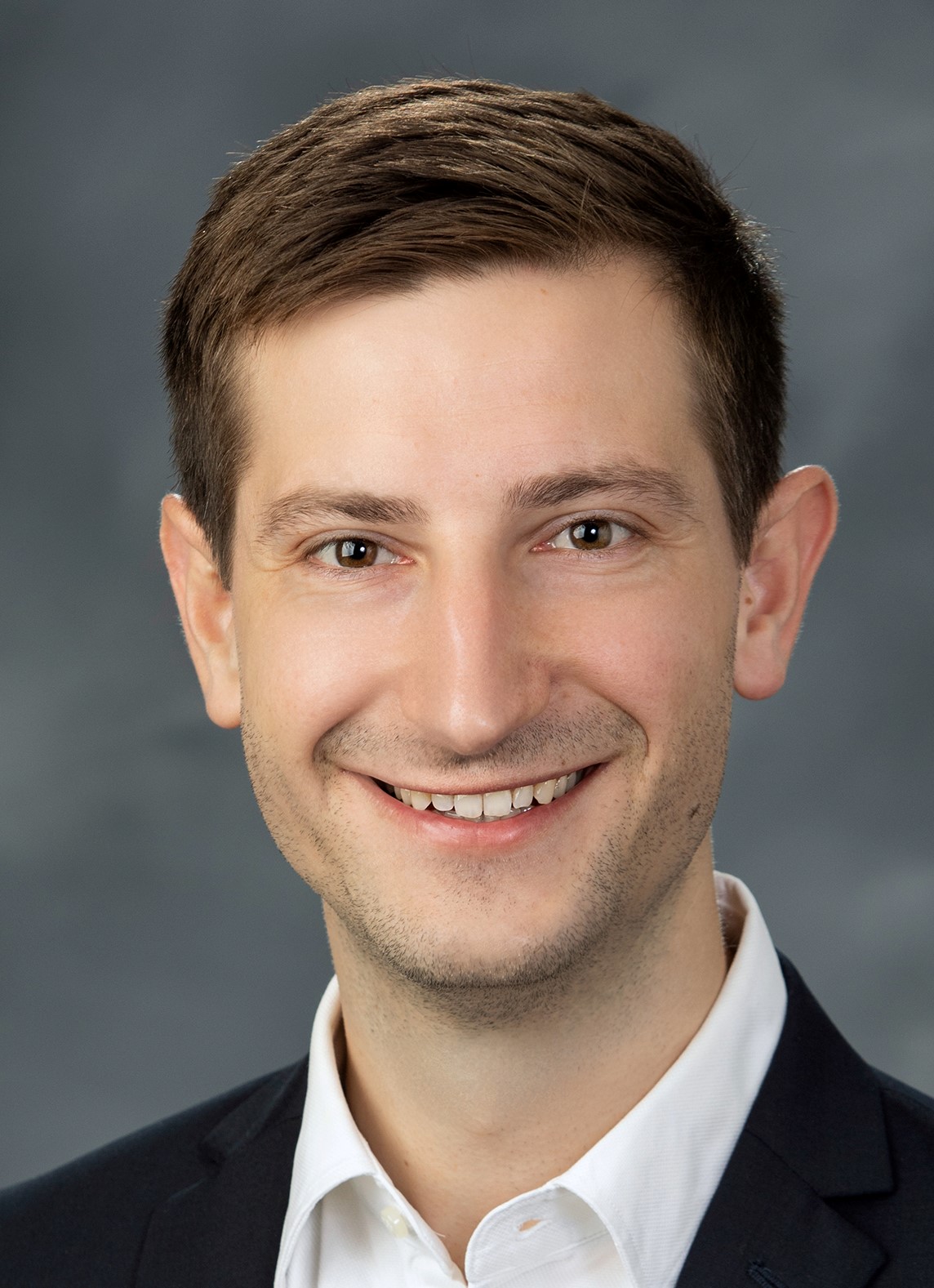} is a Ph.D. candidate in aeronautics at the California Institute of Technology. He earned his MS in aeronautics from Caltech in 2016 and his BS in mechanical engineering from the TU Braunschweig in 2014. Marcel works in the field of experimental fluid mechanics with a focus on low Reynolds number multirotor aerodynamics. His Ph.D. thesis project is on the experimental investigation of multirotor configurations in axial descent. Marcel’s other research interests include novel multi-fan wind tunnel design concepts and active flow control technologies for advanced aircraft designs.  
\end{biographywithpic}

\begin{biographywithpic}
{Haley Cummings}{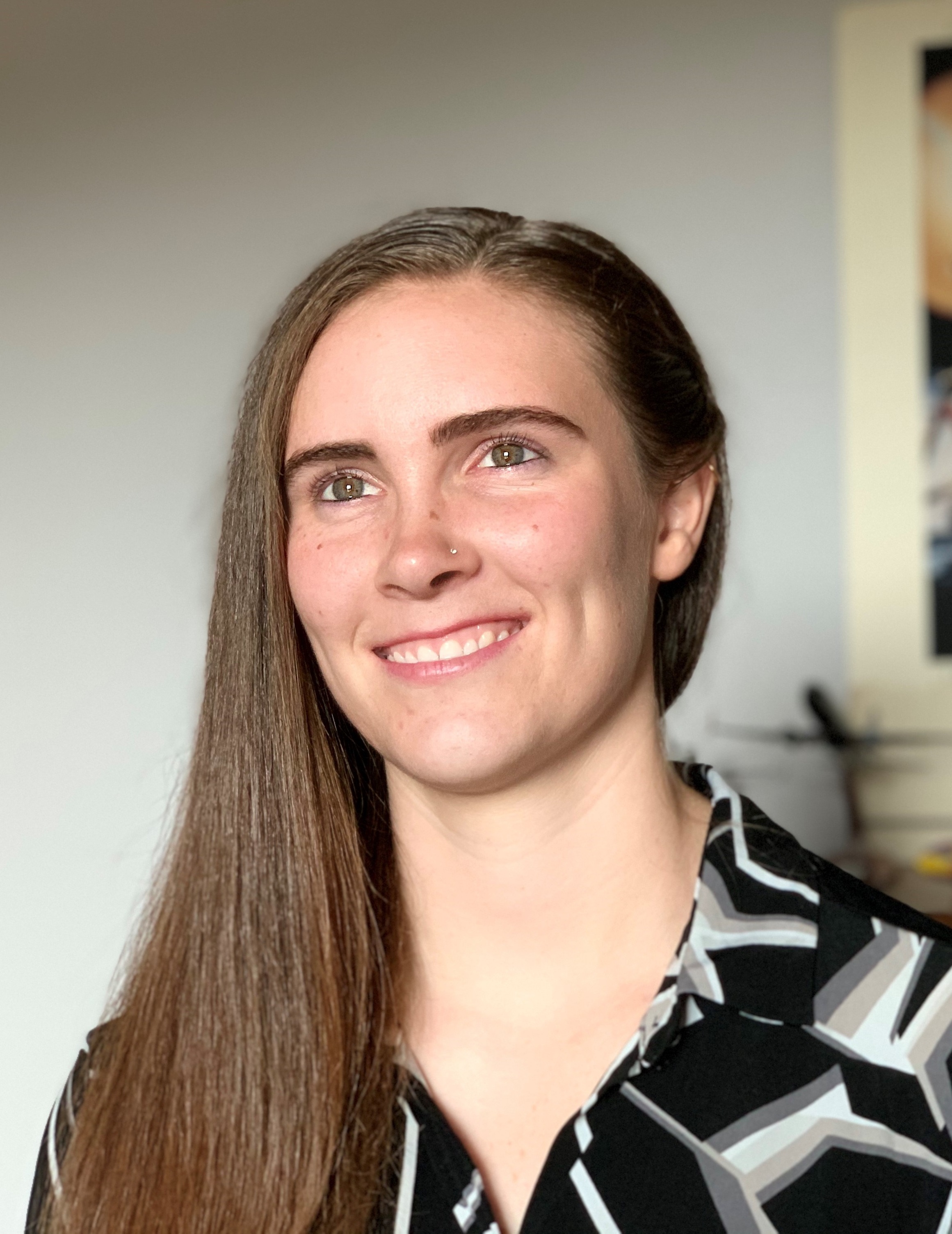}
is the Principal Investigator for the Early Career Initiative project entitled Rotor Optimization for the Advancement of Mars eXploration (ROAMX) at NASA Ames Research Center in California, which is seeking to computationally optimize and experimentally validate helicopter rotors for flight on Mars. She started working at Ames in 2014 as an intern, came back again in 2016 as an intern, was hired as a Pathways intern in 2017, and has worked full time at Ames since then. She earned her Bachelor’s and Master’s in mechanical engineering from Northern Illinois University in 2016 and 2018.
\end{biographywithpic}

\begin{biographywithpic}
{Jonathan Bapst}{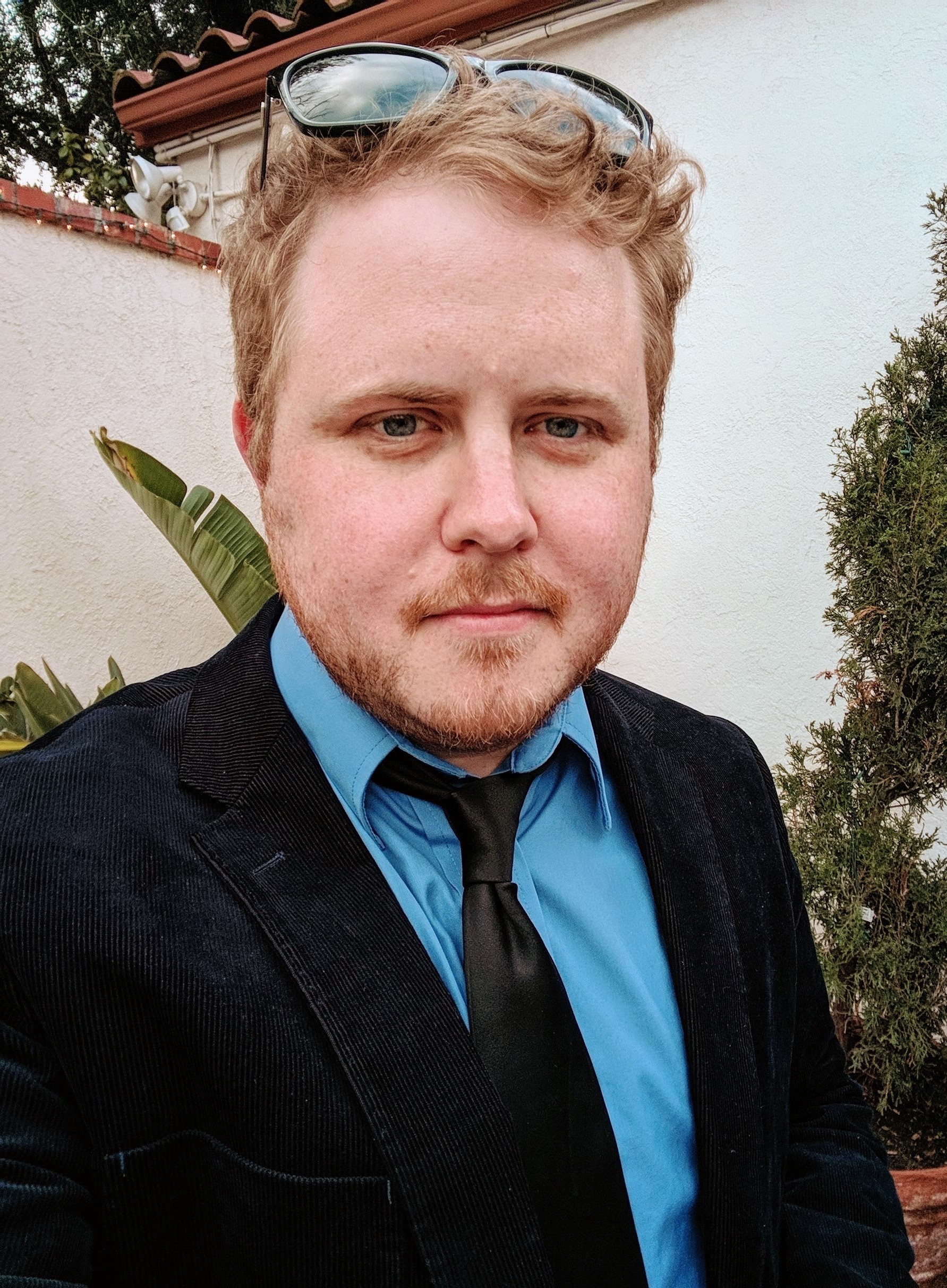}
completed his bachelor’s degree in mechanical and aerospace engineering at SUNY Buffalo and received his PhD from University of Washington in Earth and Space Sciences and Astrobiology. He is interested in using spacecraft data and theoretical models to characterize the surface properties of other worlds. At JPL, Jonathan leads the science efforts for Mars Science Helicopter SRTD, where he is defining mission concepts for a next-generation Mars rotorcraft.
\end{biographywithpic}

\begin{biographywithpic}
{Theodore Tzanetos}{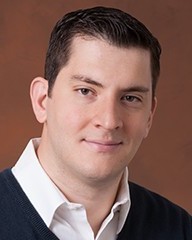}
received a B.S. in Computer Science and Electrical Engineering from the Massachusetts Institute of Technology (MIT) in 2012, and a M.Eng. in the same field also from MIT in 2013. He has served as Ingenuity’s Team Lead, Operations Lead, Deputy Operations Lead, Helicopter Assembly-Test-and-Launch-Operations Lead, Flight Test Conductor, and Electrical Ground Support Equipment Lead. He is also a Principal Investigator for NASA JPL’s Mars Science Helicopter System research initiative, working to develop the future of Martian aerial exploration. He is part of JPL’s Aerial Mobility Group, having worked at JPL for 5 years.
\end{biographywithpic}

\begin{biographywithpic}
{Roland Brockers}{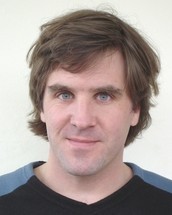}
is a Research Technologist at the Jet Propulsion Laboratory. He received his Ph.D. in Electrical Engineering from the University of Paderborn (Germany) in 2005, and has been conducting research in autonomous navigation of unmanned robotic systems for more than 20~years with a focus on aerial vehicles since 2010 where he worked on autonomous landing and ingress, visual pose estimation, and autonomous obstacle avoidance for micro air vehicles. Roland was part of the NASA Mars Helicopter Ingenuity GNC team and Ingenuity operations team and wrote the image-processing software that Ingenuity uses to observe its motion. 
His current research interests include 3D perception systems for small mobile robotic platforms, and autonomous robotic systems with applications in earth science and planetary exploration.
\end{biographywithpic}

\begin{biographywithpic}
{Dr. Abhinandan Jain}{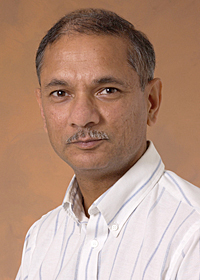}
 is a Senior Research Scientist at the Jet Propulsion Laboratory. He leads the Dynamics and Real-Time Simulation (DARTS) Laboratory which is responsible for the development of high- performance simulations for NASA's deep space missions. He has authored the development of the DARTS multibody dynamics software for which he received the NASA Software of the Year award. Dr. Jain is one of the inventors of the Spatial Operator Algebra methodology for dynamics modeling and is the author of a book on the topic and several research publications. Under his leadership, the DARTS Lab has developed advanced physics-based simulators such as DSENDS and ROAMS for NASA missions involving entry, descent \& landing, surface rovers, robotics and near-surface airships. These tools are widely used by NASA/JPL projects including Mars Sample Return, Mars 2020, ARTEMIS, Mars Helicopter, Venus Balloon etc. as well as by R\&D projects for technology development.
\end{biographywithpic}

\begin{biographywithpic}
{David S. Bayard}{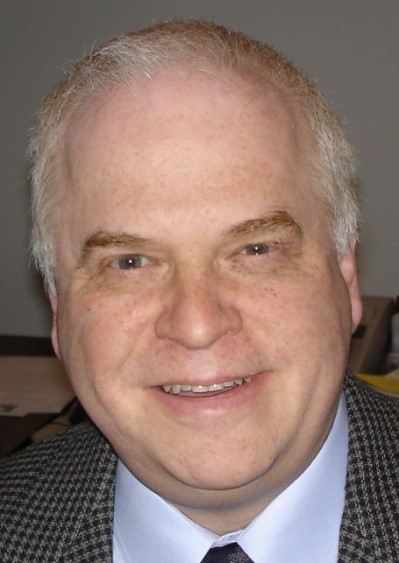}
is a JPL Technical Fellow with 35 plus years’ experience in the aerospace industry. He has a Ph.D in Electrical Engineering from the State University of New York at Stony Brook (1984). At JPL, he has been involved in the application of modern estimation and control techniques to a wide range of emerging spacecraft and planetary missions. Recently, Dr. Bayard led the JPL Mars Helicopter Navigation Team responsible for developing the navigation system for the Ingenuity helicopter. Dr. Bayard received numerous NASA Awards and Medals during his career. He is an Associate Fellow of AIAA.	
\end{biographywithpic}

\begin{biographywithpic}
{Olivier Toupet}{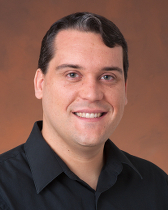} received his M.S. degree in Aeronautics and Astronautics from MIT in 2006. He is currently the supervisor of the Robotic Aerial Mobility Group at the Jet Propulsion Laboratory, which develops innovative technologies for UAVs with a focus on guidance, navigation, and control. Additionally Mr. Toupet has several roles on the Mars 2020 project, including deputy lead of the Rover Planner team, lead of the Strategic Route Planning team, deputy lead of the Helicopter Integration Engineer team, and lead of the Enhanced Autonomous Navigation Flight Software team.  
\end{biographywithpic}

\begin{biographywithpic}
{Joel Burdick}{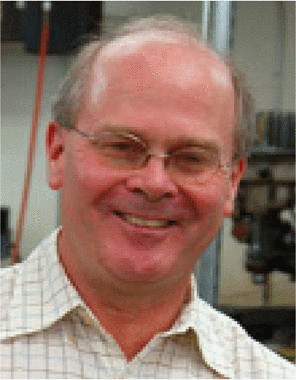} is a Professor of mechanical engineering and bioengineering at the California Institute of Technology. He is also a laboratory research scientist and the Jet Propulsion Laboratory. Professor Burdick focuses on robotics, kinematics, mechanical systems and control. Active research areas include: robotic locomotion, sensor-based motion planning algorithms, multi-fingered robotic manipulation, applied nonlinear control theory, neural prosthetics, and medical applications of robotics.  
\end{biographywithpic}

\begin{biographywithpic}
{Mory Gharib}{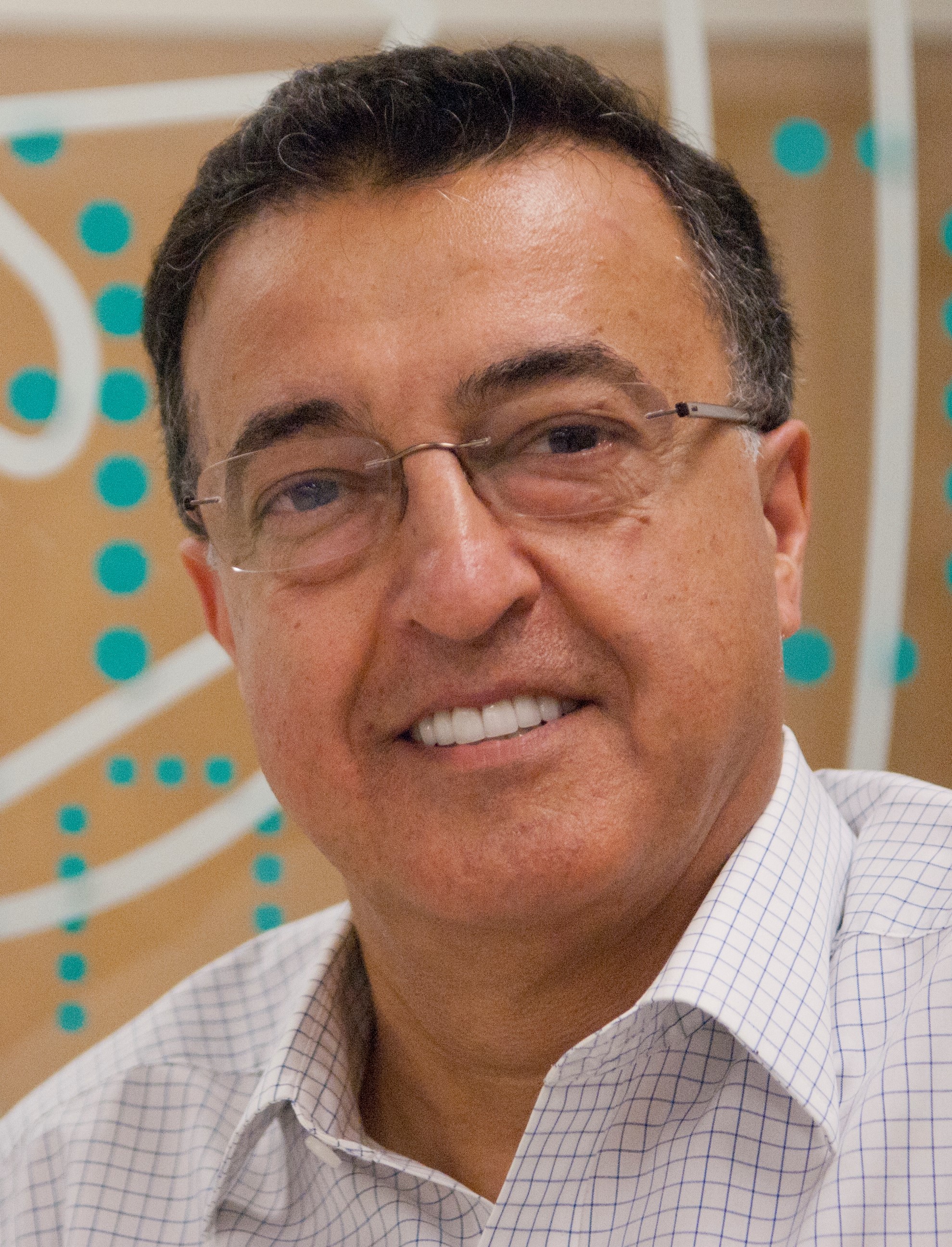} is the Hans W. Liepmann Professor of Aeronautics and Bioinspired Engineering at Caltech and the Booth-Kresa Leadership Chair for the Center for Autonomous Systems and Technologies. Currently he serves as Chair of the Graduate Aerospace Department (GALCIT) and as the Director of the Center for Autonomous Systems and Technologies. Professor Gharib's research interests are in conventional fluid dynamics and aeronautics, including vortex dynamics, active and passive flow control, autonomous flight, and underwater systems. His research in the life sciences include cardiovascular and eye physiology, and medical device development. Dr. Gharib has published more than 250 papers in refereed journals and has been issued more than 130 U.S. Patents.  
\end{biographywithpic}

\begin{biographywithpic}
{J. (Bob) Balaram}{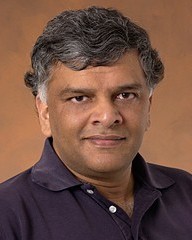}
is Principal Member of Technical Staff at the NASA Jet Propulsion Laboratory where he is with the Mobility \& Robotic Systems Section. He received his Ph. D. in Computer and Systems Engineering from Rensselaer Polytechnic Institute. At JPL he has been active in the area of telerobotics technology development for Mars Rovers, planetary balloon aerobot systems, and multi-mission, high-fidelity Spacecraft simulators for Entry, Descent and Landing and Surface Mobility. Bob was the chief Engineer for the Mars Helicopter project at the Jet Propulsion Laboratory.
\end{biographywithpic}

\end{document}